\documentclass[journal]{IEEEtran}

\usepackage{lineno}
\usepackage{hyperref}
\hypersetup{
    colorlinks=true,
    linkcolor=blue,
    filecolor=magenta,      
    urlcolor=blue,
}

\usepackage{amsmath, amssymb}
\usepackage{multirow}
\usepackage{subfig}
\usepackage{graphicx}
\usepackage{adjustbox}
\usepackage{caption}
\usepackage{xcolor}

\begin{document}

\title{Deep Learning for Medical Anomaly Detection - A Survey}

\author{ Tharindu Fernando, Harshala Gammulle, Simon Denman, Sridha Sridharan, and Clinton Fookes  
\thanks{T. Fernando, H. Gammulle, S. Denman, S. Sridharan and C. Fookes are with SAIVT, Queensland University of Technology, Australia (e-mail: t.warnakulasuriya@qut.edu.au.)}
}
\maketitle

\begin{abstract}
 Machine learning-based medical anomaly detection is an important problem that has been extensively studied. Numerous approaches have been proposed across various medical application domains and we observe several similarities across these distinct applications. Despite this comparability, we observe a lack of structured organisation of these diverse research applications such that their advantages and limitations can be studied. The principal aim of this survey is to provide a thorough theoretical analysis of popular deep learning techniques in medical anomaly detection. In particular, we contribute a coherent and systematic review of state-of-the-art techniques, comparing and contrasting their architectural differences as well as training algorithms. Furthermore, we provide a comprehensive overview of deep model interpretation strategies that can be used to interpret model decisions. In addition, we outline the key limitations of existing deep medical anomaly detection techniques and propose key research directions for further investigation.

\end{abstract}

\begin{IEEEkeywords}
Deep Learning, Anomaly detection, Machine Learning, Temporal analysis
\end{IEEEkeywords}


\IEEEpeerreviewmaketitle

\section{Introduction}

Identifying data samples that do not fit the overall data distribution is the principle task in anomaly detection. Anomalies can arise due to various reasons such as noise in the data capture process, changes in underlying phenomenon, or due to new or previously unseen conditions in the captured environment. Therefore, anomaly detection is a crucial task in medical signal analysis. 

The dawn of deep learning has revolutionised the machine learning field and it's success has seeped into the domain of medical anomaly detection, which has resulted in a myriad of research articles leveraging deep machine learning architectures for medical anomaly detection. 

The principal aim of this survey is to present a structured and comprehensive review of this existing literature, systematically comparing and contrasting methodologies. Furthermore, we provide an extensive investigation in to deep model interpretation strategies, which is critical when applying `black-box' deep models for medical diagnosis and to understand why a decision is reached. In addition, we summarise the challenges and limitations of existing research, and identify key future research directions, paving the way for the prevalent and effective application of deep learning in medical anomaly detection.   

\subsection{What are Anomalies?}
\label{sec:intro_1}
Anomaly detection is the task of \textit{identifying out of distribution examples}. Simply put, it seeks to detect examples that do not follow the general pattern present in the dataset. This is a crucial task as anomalous observations correlate with types of problem or fault, such as structural defects, system or malware intrusions, production errors, financial frauds or health problems.
Despite the straightforward definition, identifying anomalies is a challenging task in machine learning. One of the main challenges arises from the inconsistent behaviour of different anomalies, and the lack of constant definition of what constitutes an anomaly \cite{fernando2020neural, thudumu2020comprehensive, chalapathy2019deep}. For example, in a particular context a certain heart rate can be normal, while in a different context it could indicate a health concern. Furthermore, noisy data capture settings and/or dynamic changes in monitoring environments can lead normal examples to appear as out of distribution samples (i.e. abnormal), yielding higher false positive rates \cite{zhao2019robust}. Hence, intelligent learning strategies with high modelling capacity are required to better segregate the anomalous samples from normal data. 

\subsection{Why are Medical Anomalies Different?}

\begin{figure}[htbp]
    \centering
    \includegraphics[width=.45\linewidth]{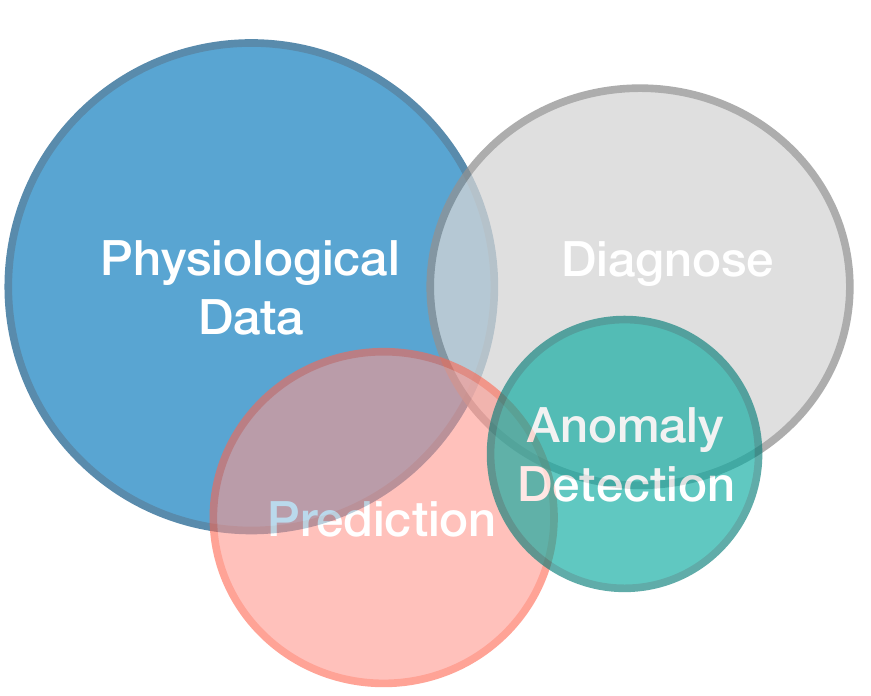}
    \caption{Illustration of the main stages in medical data processing and how anomaly detection relates to other stages.}
    \label{fig:figure_1}
\end{figure}

The diagram in Fig. \ref{fig:figure_1} illustrates the main stages with respect to medical data processing with machine learning, and how each stage relates to anomaly detection. Collected physiological data is analysed and typically utilised for i) prediction and/or ii) diagnosis. Prediction tasks include predicting future states of physiological signals such as blood pressure, or other characteristics such as recovery rates. For diagnosis tasks a portion of the data is analysed to recognise pathological signs of specific medical conditions. Anomaly detection relates to both prediction and diagnosis tasks, as it captures unique characteristics of the physiological data that could offer information about the data or patient. 

\begin{figure*}[htbp]
    \centering
    \subfloat[][]{\includegraphics[width=.55\linewidth]{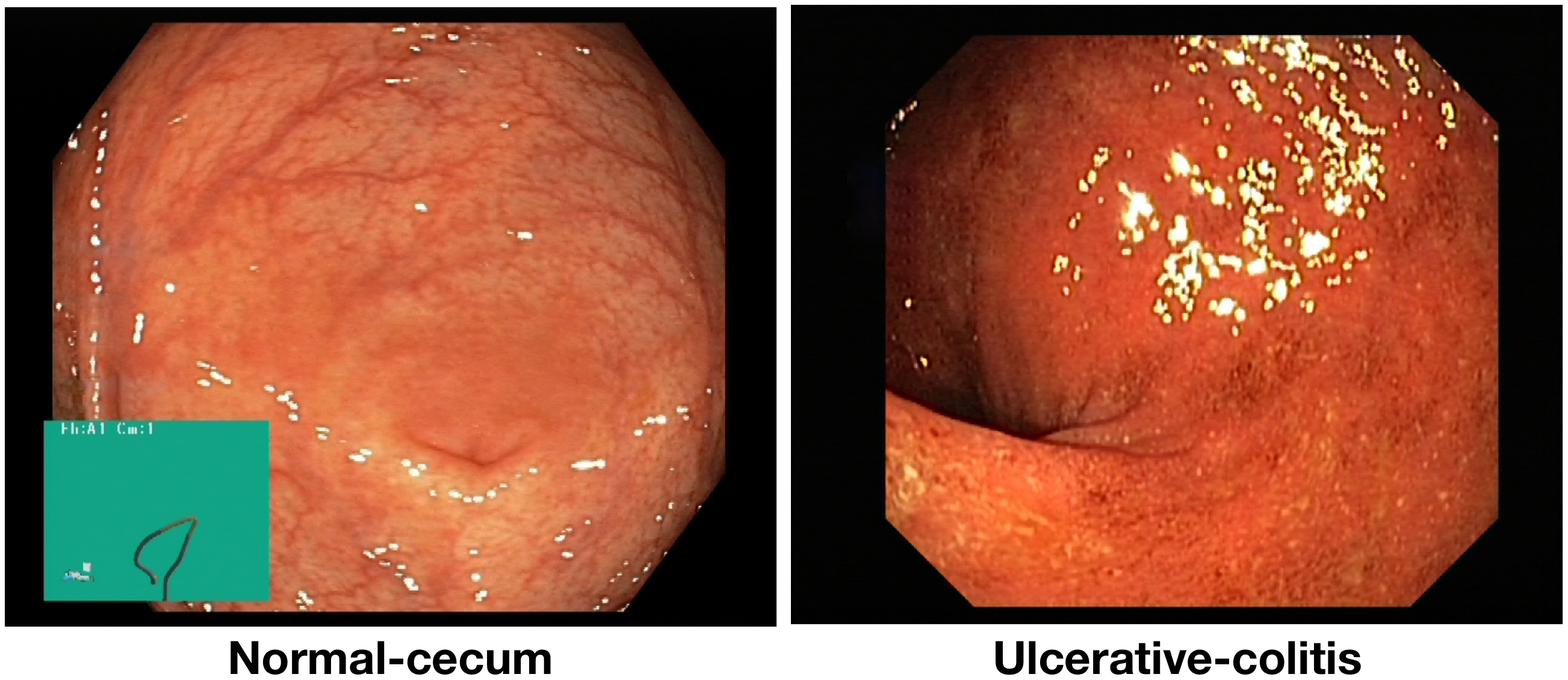}}
    \subfloat[][]{\includegraphics[width=.40\linewidth]{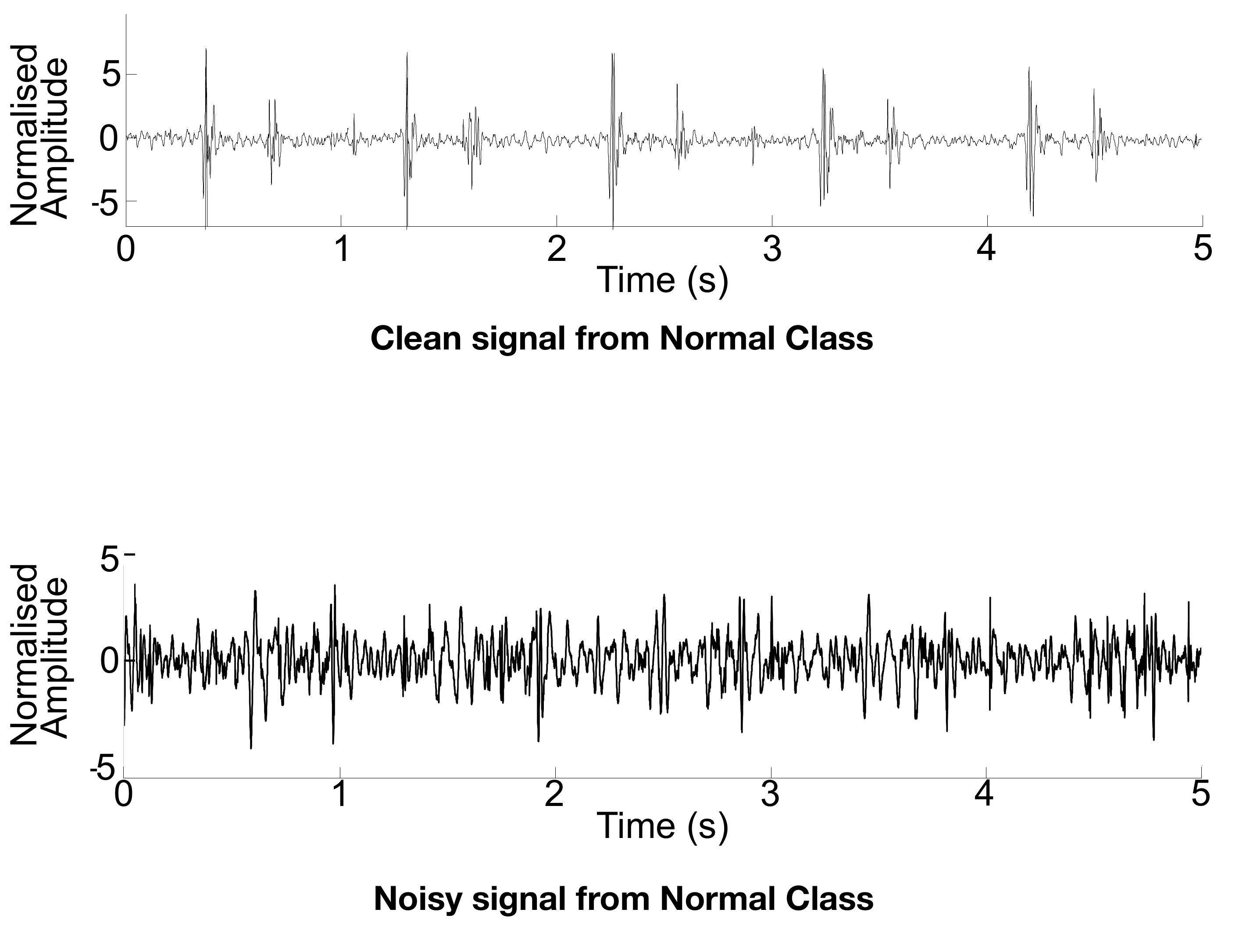}}
     \caption{Challenges associated with medical anomaly detection. (a) Two examples (normal and abnormal) from the Kvasir endoscopy image dataset \cite{pogorelov2017kvasir} with strong visual similarities. (b) Two normal examples from the PhysioNet CinC 2016 heart sound dataset \cite{clifford2016classification}, where the signal in the bottom row is corrupted by noise.}
    \label{fig:challenges}
\end{figure*}

Similar to other application domains, medical anomaly detection also inherits the challenges described in Sec. \ref{sec:intro_1}. For instance, Fig. \ref{fig:challenges} (a) illustrates two examples from the Kvasir endoscopy image dataset \cite{pogorelov2017kvasir}. Despite the strong visual similarities, the left figure is an example from the normal-cecum class while the right figure is an example from the ulcerative-colitis disease category. Another example is given in Fig. \ref{fig:challenges} (b) which illustrates the diverse nature of the normal data in a typical medical dataset. These examples are heart sound recordings from the PhysioNet Computing in Cardiology Challenge 2016 \cite{clifford2016classification}. The top figure shows a clean normal heart sound recording. While the figure in the 2nd row represents a recording of the normal category that has been corrupted by noise during data capture. Therefore, when modelling normal examples, the model should have the capacity to represent the diverse nature of the normal data distribution.
Apart from these inherent challenges, medical anomaly detection has additional hindrances which are application specific. Firstly as the end application is primarily medical diagnosis, the test sensitivity (the ability to correctly identify the anomalous samples) is a decisive and crucial factor, and the abnormality detection model is required to be highly accurate. Secondly, there are numerous patient specific characteristics that contribute to dissimilarities among different data samples. For instance, in \cite{alahmadi2016different} the authors have identified substantial differences among children from different demographics with respect to their resting state in EEG data. There are also substantial differences between different age groups, genders, etc. Therefore, when designing an accurate medical anomaly detection framework measures should be taken to mitigate such hindrances. Considering these challenges, medical anomaly detection is often posed as a supervised learning task \cite{gornitz2013toward, fernando2020neural}, where a supervision signal is presented for the model to learn to discriminate normal from abnormal examples. This is in contrast to other domains such as production defect detection or financial frauds detection, where anomalies are detected in an unsupervised manner.

\subsection{Why use Deep Learning for Medical Anomaly Detection?}

Deep learning is becoming increasingly popular among researchers in biomedical engineering as it offers a way to address the above stated challenges. One prominent characteristics of deep learning is it's ability to model non-linearity. Increasing non-linearity in the model can better segregate normal and anomalous samples, and better model the inconsistencies in the data. An additional merit that deep learning brings is its automatic feature learning capability. The availability of big-data \cite{beam2018big} and increased computational resources has empowered deep learning's hierarchical feature learning process, avoiding the need to explicitly hand-craft and define what constitutes an anomaly. Another interesting trait of deep learning is its ability to uncover long-term relationships within the data seamlessly through the neural network architecture \cite{fernando2020neural}, without explicitly defining them during feature design. For instance, recurrent architectures such as Long Short-Term Memory (LSTM) \cite{hochreiter1997long} and Gated Recurrent Units (GRU) \cite{chung2014empirical} can efficiently model temporal relationships in time series data using what is termed `memory'.

\subsection{Our Contributions}
Although several recent survey articles \cite{chalapathy2019deep, thudumu2020comprehensive} on anomaly detection have briefly touched upon the medical anomaly detection domain, and despite numerous survey papers published on specific medical application domains \cite{rasheed2020machine, li2020review, du2019review, lundervold2019overview, soffer2020deep}, there is no systematic review of deep learning based medical anomaly detection techniques which would allow readers to compare and contrast the strengths and weakness of different deep learning techniques, and leverage those findings for different medical application domains. {Tab. \ref{tab:comparison_of_this_surve} summarises these limitations.} This paper directly addresses this need and contributes a thorough theoretical analysis of popular deep learning model architectures, including convolutional neural networks, recurrent neural networks, generative adversarial networks, auto encoders, and neural memory networks; and their application to medial anomaly detection. Furthermore, we extensively analyse different model training strategies, including unsupervised learning, supervised learning and multi-task learning.

\begin{table}[htbp]
\caption{Comparison of Our Survey to Other Related Studies}
{\resizebox{\linewidth}{!}{
\begin{tabular}{cc|c|c|c|c|c|c|c|c|}
\cline{3-10}
\multicolumn{1}{l}{}                                        & \multicolumn{1}{l|}{}           & \multicolumn{8}{c|}{Reference}               \\ \cline{3-10}
\multicolumn{1}{l}{}                                        & \multicolumn{1}{l|}{}           & \cite{chalapathy2019deep} & \cite{thudumu2020comprehensive} & \cite{du2019review} & \cite{li2020review} & \cite{lundervold2019overview} & \cite{rasheed2020machine} & \cite{soffer2020deep} & Proposed\\ \hline
\multicolumn{1}{|c|}{\multirow{4}{*}{Algorithmic Approach}} & Unsupervised                    & \checkmark  & \checkmark  & \checkmark  & \checkmark  & \checkmark  & \checkmark  &     &   \checkmark       \\ \cline{2-10} 
\multicolumn{1}{|c|}{}                                      & Supervised                      & \checkmark  & \checkmark  & \checkmark  & \checkmark  & \checkmark  & \checkmark  & \checkmark   & \checkmark          \\ \cline{2-10} 
\multicolumn{1}{|c|}{}                                      & Recurrent                       & \checkmark  &    & \checkmark  & \checkmark  & \checkmark  & \checkmark  &     &  \checkmark        \\ \cline{2-10} 
\multicolumn{1}{|c|}{}                                      & Multi-Task                      &    &    &    &    &    &    &     & \checkmark         \\ \hline
\multicolumn{1}{|c|}{\multirow{5}{*}{Network Architecture}} & Auto-Encoders                   & \checkmark  &    & \checkmark  &    &    &    &     & \checkmark         \\ \cline{2-10} 
\multicolumn{1}{|c|}{}                                      & Generative Adversarial Networks & \checkmark  &    & \checkmark  &    & \checkmark  & \checkmark  &     &  \checkmark        \\ \cline{2-10} 
\multicolumn{1}{|c|}{}                                      & Neural Memory Networks          &    &    &    &    &    &    &     &  \checkmark        \\ \cline{2-10} 
\multicolumn{1}{|c|}{}                                      & Long Short-Term Memory Networks & \checkmark  &    & \checkmark  & \checkmark  & \checkmark  &    &     & \checkmark         \\ \cline{2-10} 
\multicolumn{1}{|c|}{}                                      & Gated Recurrent Units           & \checkmark  &    & \checkmark & \checkmark  & \checkmark  &    &     &  \checkmark        \\ \hline
\multicolumn{1}{|c|}{\multirow{4}{*}{Applications}}         & MRI-based Anomaly Detection     & \checkmark  &    &    &    & \checkmark  &    &     & \checkmark         \\ \cline{2-10} 
\multicolumn{1}{|c|}{}                                      & Anomalies in Endoscopy Images   &    &    & \checkmark  &    &    &    & \checkmark   &    \checkmark      \\ \cline{2-10} 
\multicolumn{1}{|c|}{}                                      & Heart Sound Anomalies           &    &    &    & \checkmark  &    &    &     &  \checkmark        \\ \cline{2-10} 
\multicolumn{1}{|c|}{}                                      & Epileptic Seizures              & \checkmark  &    &    &    &    & \checkmark  &     & \checkmark         \\ \hline
\multicolumn{1}{|c|}{Model Interpretation Methods}          &                                 &    &    &    &    &    &    &     &         \checkmark \\ \hline
\multicolumn{1}{|c|}{Medical Data Capturing Processes}      &                                 &    &    &    &    &    &    &     & \checkmark         \\ \hline
\end{tabular}}}
\label{tab:comparison_of_this_surve}
\end{table}

Moreover, this paper provides a comprehensive overview of deep model interpretation strategies that can be used to interpret model decisions. This analysis systematically illustrates how these methods generates model agnostic interpretations, and the limitations of these methods when applied to medical data. 

Finally, this review details the limitations of existing deep medical anomaly detection approaches and lists key research directions, inspiring readers to direct their future investigations towards generalisable and interpretable deep medical anomaly detection frameworks, as well as probabilistic and causal approaches which may reveal cause and effect relationships within the data. 

\subsection{Organisation}
In Sec. \ref{sec:aspects_of_DL} we illustrate different aspects of deep anomaly detection algorithms, illustrating the motivation for these architectures, and highlighting the complexities associated with medical anomaly detection. Specifically, Sec. \ref{sec:data} illustrates the types of data available in the medical anomaly detection domain, and how different deep learning architectures are designed to capture information from different modalities. Sec. \ref{sec:algorithms} categorises deep anomaly detection architectures based on their training objectives, discussing the theories behind these algorithms and the merits and deficiencies of them. Sec. \ref{sec:applications} provides an overview of key application domains to which deep medical anomaly detection has been applied. In Sec. \ref{sec:interpretation} we theoretically outline deep model interpretation strategies which are a key consideration when deploying deep models in medical applications. Sec. \ref{sec:challenges_and_future} illustrates some of the challenges and limitations of existing deep anomaly detection frameworks, and provides future directions to pursue. Sec. \ref{sec:conclusion} contains concluding remarks. 

\section{Detecting Medical Anomalies with Deep Learning}
\label{sec:aspects_of_DL}

In this section we identify different aspects of deep medical anomaly detection algorithms, including the types of data used, different algorithmic architectures, and different application domains that are considered. The following subsections discuss existing deep medical anomaly detection algorithms within each of these categories. 

\subsection{Types of Data}
\label{sec:data}
Biomedical signals can be broadly categorised into biomedical images, electrical biomedical signals, and other biomedical data such as data from laboratory results, audio recordings and wearable medical devices. The following subsections provide a brief discussion of popular applications scenarios. We also refer the readers to supplementary material where we provide a more comprehensive discussion regarding each of these categories.

\subsubsection{Biomedical Imagining}~\\
\noindent\textbf{X-ray radiography}: X-rays have shorter wave lengths than visible light and can pass through most tissue types in the human body. However, the calcium contained in bones is denser and scatters the x-rays. The film that sits on the opposite side of the x-ray source is a negative image such that areas that are exposed to more light appear darker. Therefore, as more x-rays penetrate tissues such as lungs and mussels, these areas are darkened on the film and the bones appear as brighter regions. X-ray imaging is typically used for various diagnostic purposes, including detecting bone fractures, dental problems, pneumonia, and certain types of tumor.


\noindent\textbf{Computed Tomography scan (CT):}
In CT imaging, cross sectional images of the body are generated using a narrow beam of x-rays that are emitted while the patient is quickly rotated. CT imaging collects a number of cross sectional slices which are stacked together to generate a 3 dimensional representation, which is more informative than a conventional X-ray image. 
CT scans are a popular diagnostic tool when identifying disease or injury within various regions of the body. Applications include detecting tumors or lesions in the abdomen, and localising head injuries, tumors, and clots. They are also used for diagnosing complex bone fractures and bone tumors. 



\noindent\textbf{Magnetic Resonance Imaging (MRI):}
As the name implies MRI employs a magnetic field for imagining by forcing protons in the body to align with the applied field. 

Specifically, the protons in the human body spin and create a small magnetic field. When a strong magnetic field such as from the MRI machine is introduced, the protons align with that field. Then a radio frequency pulse is introduced which disrupts the alignment. When the radio frequency pulse is turned off the protons discharge energy and try to re-align with the magnetic field. The energy released varies for different tissue types, allowing the MRI scan to segregate different regions. Therefore, MRIs are typically used to image non-bony or soft tissue regions of the human body. Comparison studies have shown that the brain, spinal cord, nerves and muscles are better captured by MRIs than CT scans. Therefore, MRI is the modality of choice for tasks such as brain tumor detection and identifying tissue damage.

In addition to these popular biomedical imaging sensor categories there exists other common data sources such as Positron Emission Tomography (PET), Ultrasound and Medical Optical Imaging. An illustration of different medical imaging signal types is provided in Fig. \ref{fig:medical_imagin}. In the Sec. 1.1 of supplementary material we provide a comprehensive discussion of these different data sources, including a discussion regarding their recent applications in deep learning as well as a list of publicly available datasets.

\begin{figure}[htbp]
  \centering
  \subfloat[][]{\includegraphics[width=.2\linewidth]{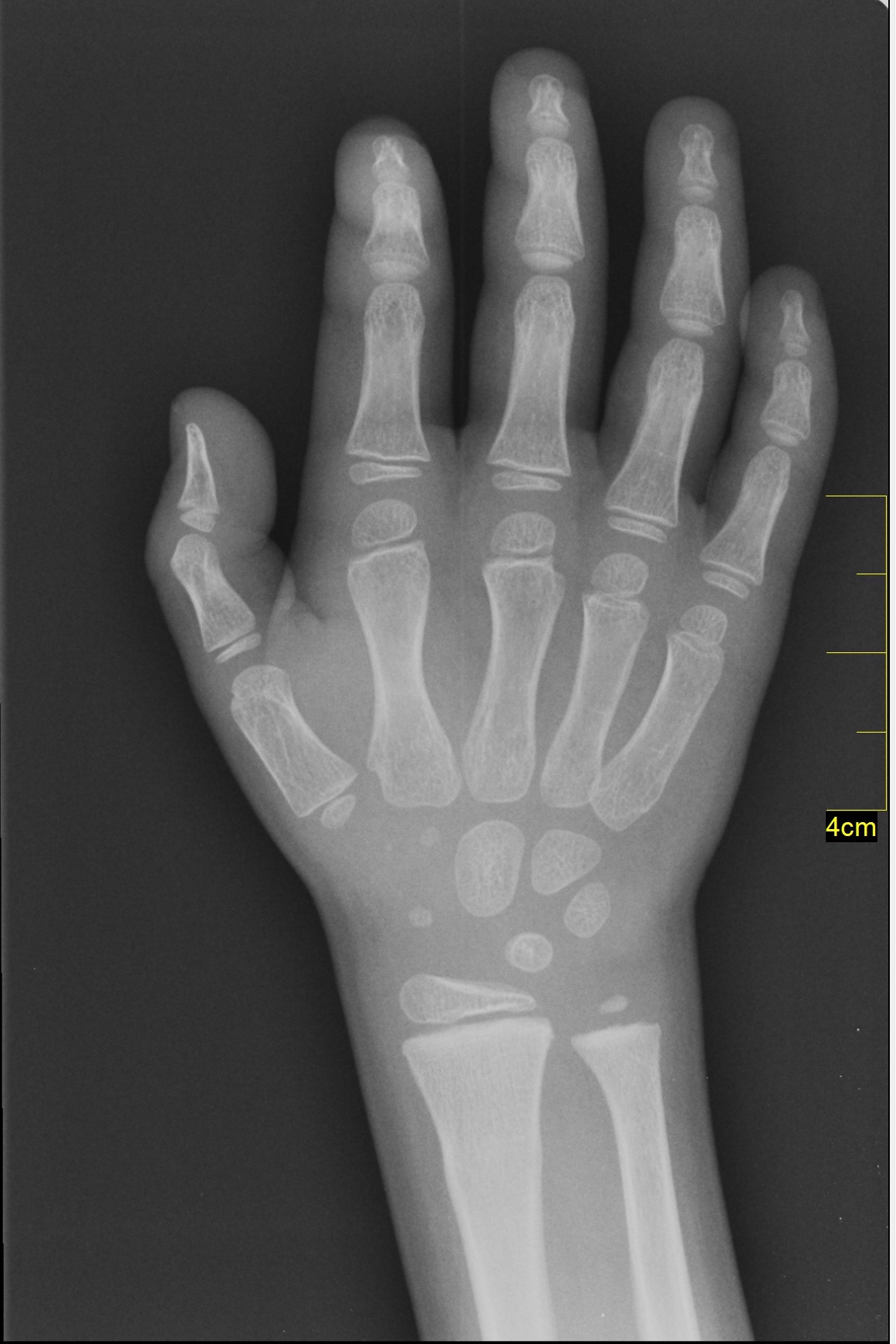}}
  \subfloat[][]{\includegraphics[width=.395\linewidth]{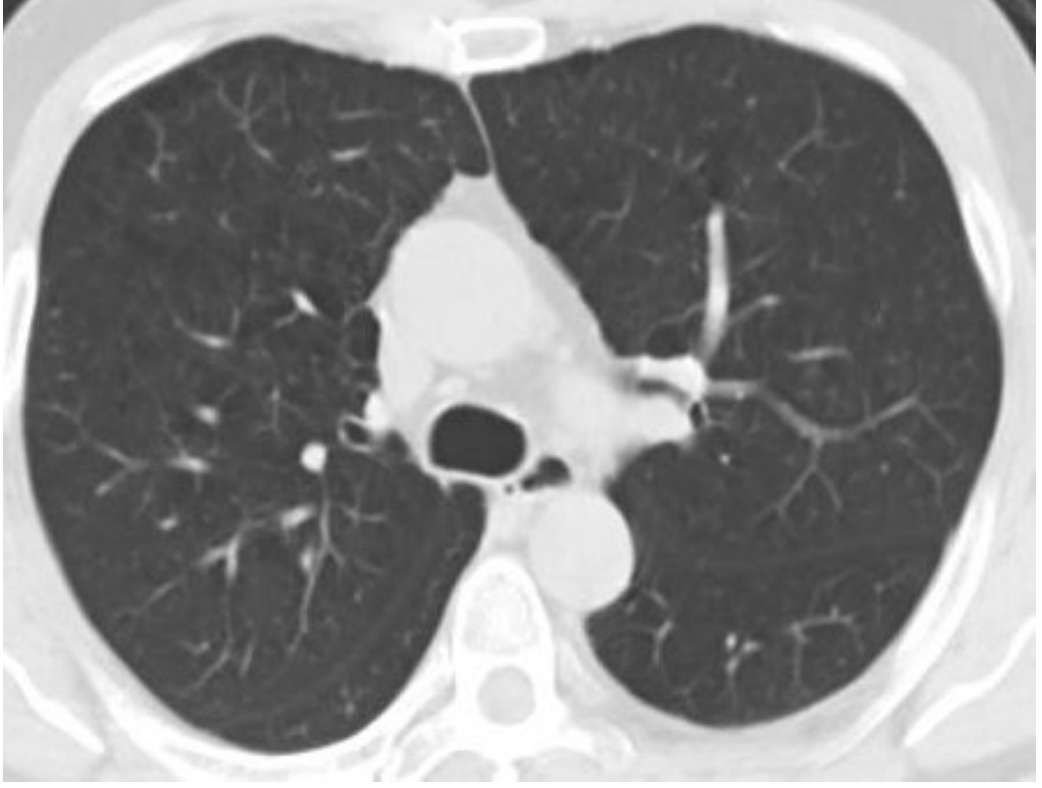}}
  \subfloat[][]{\includegraphics[width=.3\linewidth]{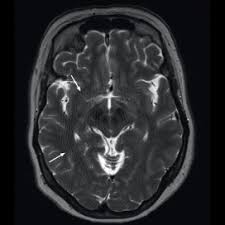}} \\
  \subfloat[][]{\includegraphics[width=.230\linewidth]{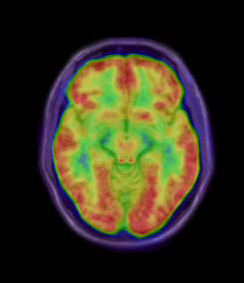}}
  \subfloat[][]{\includegraphics[width=.370\linewidth]{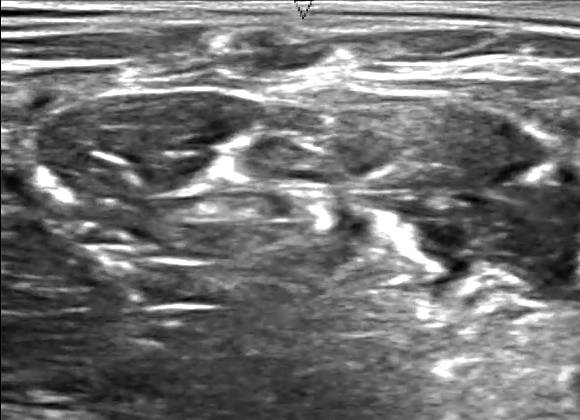}}
  \subfloat[][]{\includegraphics[width=.275\linewidth]{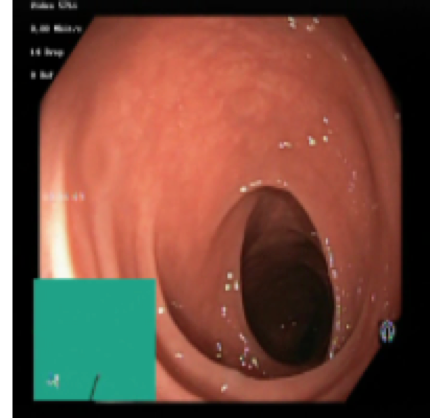}}
    \caption{Illustration of different medical imaging signals. (a)  X-ray image (\href{https://commons.wikimedia.org/}{Image Source}), (b) Lung CT scans of healthy and diseased subjects taken from the  \href{https://www.kaggle.com/plameneduardo/sarscov2-ctscan-dataset}{SARS-CoV-2 CT scan dataset}, (c) An MRI image with a brain tumor taken from \href{https://www.kaggle.com/navoneel/brain-mri-images-for-brain-tumor-detection}{Kaggle Brain MRI Images for Brain Tumor Detection dataset}, (d) An example PET scan. Image taken from \href{https://www.kaggle.com/c/pet-radiomics-challenges}{PET radiomics challenges}, (e) An ultrasound image of the neck which is taken from \href{https://www.kaggle.com/navoneel/brain-mri-images-for-brain-tumor-detection}{Kaggle Ultrasound Nerve Segmentation dataset}, (f) An endoscopy image of the gastrointestinal tract which is taken from \href{https://datasets.simula.no/nerthus/}{The Nerthus endoscopy dataset.}}
    \label{fig:medical_imagin}
\end{figure}

\subsubsection{Electrical Biomedical Signals}~\\
\noindent\textbf{Electrocardiogram (ECG)}: ECG is a tool to visualise electricity flowing through the heart which creates the heart beat, and starts at the top of the heart and travels to the bottom. At rest, heart cells are negatively charged compared to the outside environment and when they become depolarized they become positively charged. The difference in polarization is captured by the ECG. There are two types of information which can be extracted by analysing the ECG \cite{ebrahimi2020review}. First, by measuring time intervals on an ECG one can screen for irregular electrical activities. Second, the strength of the electrical activity provides an indication of the regions of the heart that are over worked or stressed. 



\noindent\textbf{Electroencephalogram (EEG)}: The EEG detects electrical activity in the brain, which uses electrical impulses to communicate. To capture the electrical activity, small metal discs (electrodes) are placed on the scalp. The electrical signals captured by these electrodes are amplified to better visualise brain activity.



EEGs are a prominent tool for observing the cognitive process of a subject. They are often used to study sleep patterns, psychological disorders, brain damage from head injury, and epilepsy. 

{\noindent\textbf{Magnetoencephalography (MEG)}: As described above, an EEG captures the electrical fields generated by extracellular currents of the human brain while MEG primarily detects the magnetic fields induced by these extracellular currents \cite{singh2014magnetoencephalography}. We acknowledge that MEG is not an electrical biomedical signal, however, we list MEG alongside with other electrical signals since it’s capturing a by-product of the brain’s electrical activity. Several studies \cite{lardone2018mindfulness, jacini2018amnestic, rucco2019mutations, hasasneh2018deep, lopez2020detection, aoe2019automatic} have investigated the utility of MEG signals for the detection of anomalous brain activities and conditions.}

In addition to ECGs, EEGs and MEGs which are the most commonly utilsed electrical biomedical signals we would like to acknowledge Electromyography (EMG) sensors where electric potential generated by muscle cells is monitored to diagnose the health of muscles and motor neurons. We refer the readers to Sec. 1.2 of supplementary material for a more comprehensive discussion related to ECGs, EEGs, MEGs, EMGs, and discussion regarding their recent applications in deep learning research and a list of publicly available datasets.

\subsubsection{Miscellaneous data types}
In addition to the primary data types discussed above we would like to acknowledge other miscellaneous data sources such as Phonocardiography (PCG) and wearable medical devices which also provide useful information to medical diagnosis. We refer the reader to Sec. 1.3 of the supplementary material where we discuss these data sources in detail, providing discussion related to their recent applications in deep learning research.  

\subsection{Algorithmic Approaches for Medical Anomaly Detection}
In this subsection, we summarise existing deep learning algorithms based on their training objectives, and whether labels for normal/abnormal are provided during training. In addition, Sec. \ref{sec:recurrent_nn} summarises popular recurrent deep neural network architectures used in the medical domain. Finally, a discussion of dimensionality differences between different data types, and how this is managed by existing deep learning research is presented. 

\label{sec:algorithms}
\subsubsection{Unsupervised Anomaly Detection}
\label{sec:unsupervised}
In unsupervised anomaly detection, no supervision signal (indicating if a sample is normal or abnormal) is provided during training. Therefore, unsupervised algorithms do not require labelled datasets, making them appealing to the machine learning community. Auto Encoders (AEs) and Generative Adversarial Networks (GANs) are two common unsupervised deep learning architectures, and are presented in the following subsections.

\begin{figure}[htbp]
    \centering
    \includegraphics[width=.95\linewidth]{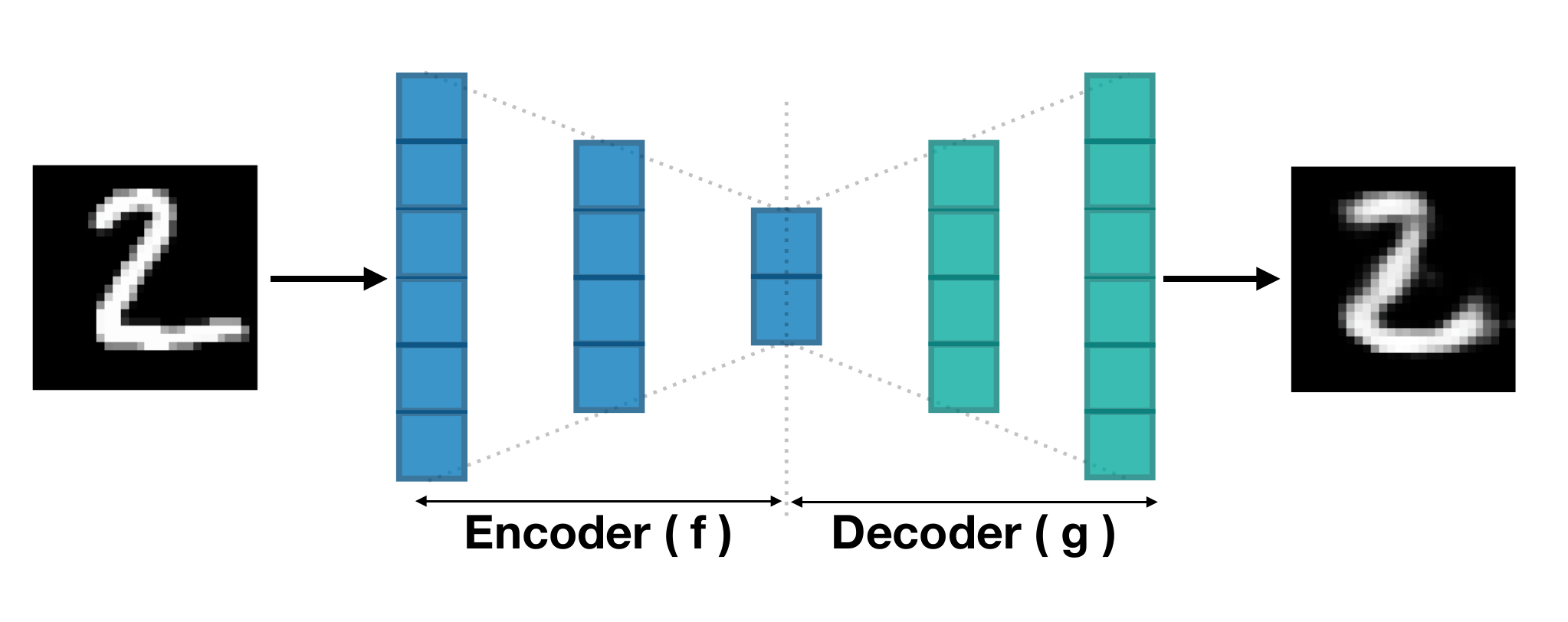}
    \caption{Illustration of the main components of an Auto Encoder.}
    \label{fig:figure_AE}
\end{figure}

\textbf{Auto Encoders (AEs)}
Since their introduction in \cite{ballard1987modular} as a method for pre-training deep neural networks, AEs have been widely used for automatic feature learning \cite{charte2018practical}. Fig. \ref{fig:figure_AE} illustrates the structure of an AE. They are symmetric and the model is trained to re-construct the input from a learned compressed representation, captured at the center of the architecture.

Formally, let there be $N$ samples in the dataset, the current input be $x$, and $f$ and $g$ denote the encoder and decoder networks respectively. Then, the compressed representation, $z$, is given by,
\begin{equation}
    z = f(x),
\end{equation}
and reconstructed using,
\begin{equation}
    y = g(z).
\end{equation}
This model is trained to minimise the reconstruction loss,
\begin{equation}
     \sum_{x \in N}{L(x, g(f(x)))},
\label{eq:AE}
\end{equation}
where $L$ is a distance function, which is commonly the Mean Squared Error (MSE).

There exist several variants of AEs. One of which is the \textit{sparse Auto Encoder (S-AE)}. An added sparsity constraint limits the number of non-zeros elements in the encoded representation, $z$. This is enforced by a penalty term added to the loss (i.e. Eq. \ref{eq:AE}),
\begin{equation}
    L_{S-AE} = \sum_{x \in N}{L(x, g(f(x)))} + \lambda \frac{1}{|N|}\sum_{x \in N}|f(x)|,
\end{equation}
where $\lambda$ is a hyper-parameter which controls the strength of the sparsity constraint.

\textit{De-Noising AEs} \cite{vincent2008extracting} learn to reconstruct a clean signal from a noisy (corrupted) input; with the aim being to leverage the de-noising capability to learn a robust and general feature encoding. \textit{Contractive AEs} try to mitigate the sensitivity of AEs to perturbations of the input samples. A regularisation term is added to the loss defined (see Eq. \ref{eq:AE}) which measures the sensitivity of the learned embedding to small changes in the input. This sensitivity is measured using Frobenius Norm of the Jacobian matrix of the encoder \cite{charte2018practical}. 

Finally, the \textit{Variational AE (VAE)} assumes that the observations, $x$, are sampled from a probability distribution and seeks to estimate the parameters of this distribution. Formally, given observations, $x$, the VAE tries to approximate the latent distribution, $P_{\phi}$. Let ${\phi}$ represent the parameters of the distribution approximating the true latent distribution and ${\theta}$ represent the parameters of the sampled distribution, then the objective of the VAE is,
\begin{equation}
L_{VAE}(\theta, \phi; x ) = \mathrm{KL} (P_\phi (z | x) || P_\theta (z)) - \mathbb{E}_{P_\phi (z | x) } (log (P_\theta (x | z))),
\end{equation}
where $\mathrm{KL}$ is the Kullback-Leibler divergence.

There have been a number of applications of auto-encoders for medical anomaly detection. In \cite{cowton2018combined} the authors proposed an AE based method for early detection of respiratory diseases in pigs. The AE is composed of GRUs to learn from the recordings temporally. An EEG based anomaly detection method is proposed in \cite{wang2016research} where the authors employ a Convolutional Neural Network (CNN) based  AE. In \cite{sato2018primitive}, a 3D-CNN based AE is used  to learn from volumetric CT scans. 

In \cite{lu2018anomaly} a VAE based framework is proposed to detect anomalies in skin images. In \cite{zimmerer2018context} the authors introduce perturbations to evaluate the effect of input representation variations on the modeled representation; and propose a two branch structure where `context-dependent' variations are also added to a VAE branch of the model. This method is validated on an MRI anomaly detection task. Another conditional model is proposed in \cite{uzunova2019unsupervised} where the authors condition the VAE output on prior knowledge. The method has been validated on both 2D and 3D anomaly detection tasks. 

Despite their interesting characteristics, AEs have limited capabilities when modelling high-dimensional data distributions, often leading to erroneous re-constructions and inaccurate approximations of the modelled data distribution \cite{saxena2020generative}. Hence, another class of generative models, \textit{Generative Adversarial Networks}, have received increasing attention. 

\textbf{Generative Adversarial Networks (GANs):} Another class of AEs are adversarial AEs, more widely known as GANs \cite{goodfellow2014generative}. They train two networks, a `Generator' (G) and a `Discriminator' (D), which play a min-max game. G tries to fool D, while D tries avoid being fooled.

\begin{figure}[htbp]
    \centering
    \includegraphics[width=\linewidth]{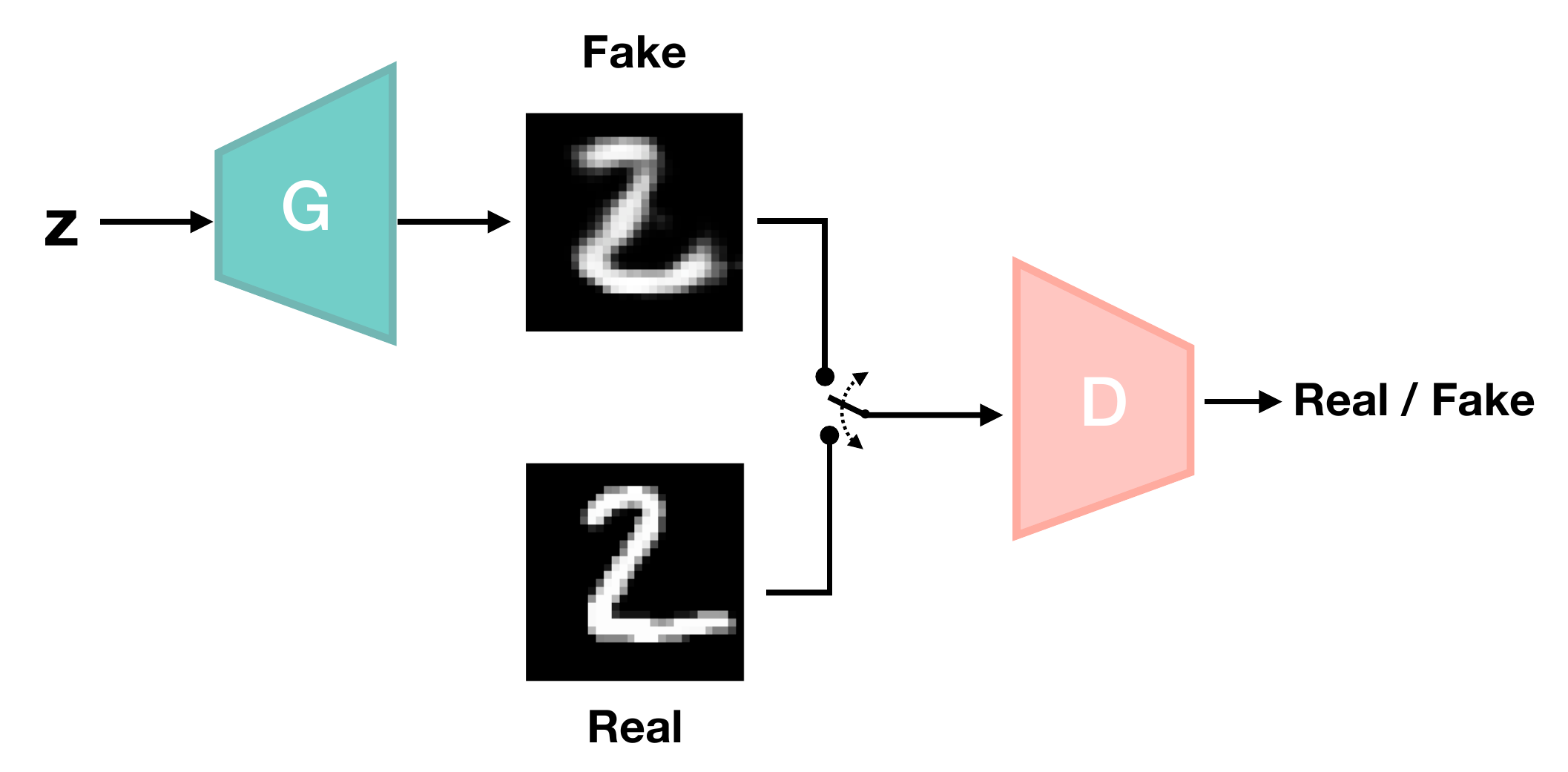}
    \caption{Main components in Generative Adversarial Networks}
    \label{fig:figure_GAN}
\end{figure}

Fig. \ref{fig:figure_GAN} illustrates the basic structure of GAN training. The generator receives noise sampled from $P_z(z)$ and seeks to learn a distribution of the true data, $P_{data}(x)$, modelling the mapping from noise space to the data space. The discriminator, $D$, outputs a scalar variable when given a synthesised (fake) or true (real) example. The discriminator is trained to output the correct label for the real/fake classification, and this objective can be written as,
\begin{equation}
\min_{G}\max_{D}V(D, G) = \mathbb{E}_{x \sim P_{data}(x)}[log D(x)] + \mathbb{E}_{z \sim P_{z}(z)}[log 1 - D(G(z))]. 
\end{equation}

While a supervised signal is provided to the discriminator for training the real/fake classification task, the real/fake classification is not the primary task and the model is not shown any anomalous examples. Hence, no supervision is given to the GAN framework regarding how to identify abnormalities, making it unsupervised.  

A popular sub-class of GANs are \textit{conditional-GANs (cGANs)}, in which both generator and discriminator outputs are conditioned on additional data, $c$. The cGAN objective is given by,
\begin{equation}
\min_{G}\max_{D}V(D, G) = \mathbb{E}_{x \sim P_{data}(x)}[log D(x | c)] + \mathbb{E}_{z \sim P_{z}(z)}[log 1 - D(G(z | c) | c)]. 
\end{equation}
cGANs are popular for tasks where the synthesised output should relate to a stimulus \cite{fernando2018task, fernando2018gd}. 

\textit{Cycle Consistency GANs (Cycle-GANs)} are popular for image-to-image translation tasks. Cycle-GANs provide an additional constraint to the framework: that the original input can be synthesised from the generated output. 

An example application of GANs for medical anomaly detection is in Schlegl et. al \cite{schlegl2019f}, where the authors used a GAN framework for anomaly detection in Optical Coherence Tomography (OCT). They trained a GAN to generate normal OCT scans using the latent distribution $z$. Then an encoder is used to map normal OCT scans to $z$. Hence, it should be possible to recover an identical image when mapping from the image to $z$ using the encoder, and from $z$ to image using the generator. When there are anomalies the authors show that there exist discrepancies in this translation, and identify anomalies using this process. 


\subsubsection{Supervised Anomaly Detection}
\label{sec:supervised}
Considering the requirement for a high degree of sensitivity and robustness, particularly due to the diagnostic applications, supervised learning has been widely applied for medical anomaly detection and has demonstrated superior performance to unsupervised methods. In contrast to unsupervised learning methods (see Sec. \ref{sec:unsupervised}), in supervised anomaly detection a supervised signal is provided indicating which examples are from the normal category and which are anomalous. Hence, this is actually a binary classification task and models are typically trained using binary cross entropy loss \cite{zhang2018generalized},
\begin{equation}
    L = -y log (f(x)) - (1-y)log(1 - f(x)),
\end{equation}
where $y$ is the ground truth label, $f$ is the classifier and $x$ is the input to the model. Example architectures include the CNN structures in \cite{schmidt2018artificial} and \cite{esteva2017dermatologist} where they employ supervised learning to identify anomalies in retina images and for automated classification of skin lesions, respectively. In \cite{turner2017deep} a deep belief network is trained to detect seizures in EEG data. 

\textit{Multi-task Learning (MtL)} is a sub category of supervised learning, and seeks to share relevant information among several related tasks rather than learning them individually \cite{jawanpuria2015efficient}. For instance, to overcome challenges which arise due to subject specific variations, a secondary subject identification task can be coupled with the primary abnormality detection task. Hence, the model learns to identify similarities and differences among subjects while learning to classify abnormalities. Several studies have leveraged MtL in the medical domain. In \cite{jawanpuria2015efficient} an efficient kernel learning structure for multiple tasks is proposed and applied to regress Parkinson's disease symptom scores. In \cite{kisilev2016medical} a multi-task learning strategy is formulated through detection and localisation of lesions in medical images, which jointly learns to detect suspicious images and segment regions of interest in those images. 

The deep learning architectures discussed so far are feed-forward architectures, i.e. the data travels in one direction, from input to output. This limits their ability to model temporal signals. To address this limitation, Recurrent Neural Networks are introduced. 

\subsubsection{Recurrent Neural Networks (RNNs)}
\label{sec:recurrent_nn}
Recurrence is a critical characteristic for tasks such as time-series modelling, and means the output of the current time step is also fed as an input to the next time step.
In medical data processing, this is important for modelling sequential data such as EEG and Phonocardiographic data, where we are concerned with the temporal evolution of the signal. 

\begin{figure}[htbp]
    \centering
    \includegraphics[width=.85\linewidth]{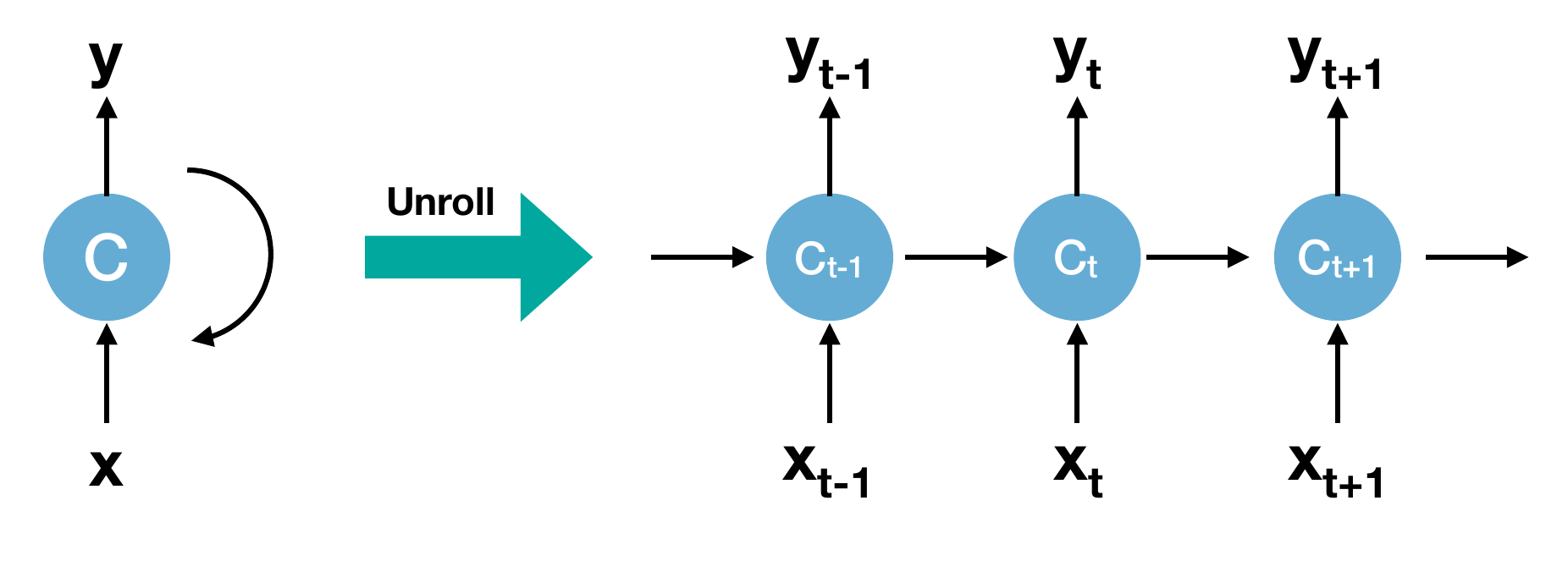}
    \caption{Illustration of Recurrent Neural Network structure and how it temporally unrolls.}
    \label{fig:figure_RNN}
\end{figure}

Fig. \ref{fig:figure_RNN} illustrates the basic structure of an RNN, where we show how it temporally unrolls. This structure requires the use of Backpropagation Through Time (BPTT) \cite{williams1995gradient}, as the gradient of the error at a given time step depends upon the prediction at the previous time step, and errors accumulate over time. Despite their interesting characteristics, vanishing gradients \cite{hochreiter1998vanishing, bengio1994learning} are a major drawback of simple RNN structures due to BPTT, and this makes them ineffective for modelling long-term dependencies (relationships between distant time-steps). 

Several RNN variants have been introduced to address this limitations. In the following sections we illustrate three of such popular variants: Long Short-Term Memory (LSTM) Networks, Gated Recurrent Units (GRUs) and Neural Memory Networks (NMNs).

\textit{Long Short-Term Memory (LSTM) \cite{hochreiter1997long}} networks are specifically designed to model long-term dependencies which are ill represented by RNNs due to vanishing gradients. They introduce a `memory cell' (or cell state) to capture long-term dependencies, and a series of gated operations to manipulate the information stored in the memory and update it over time. The core idea behind LSTMs is that long-term dependencies can be stored in the cell state \cite{hochreiter1997long}.
\begin{figure}[htbp]
    \centering
    \includegraphics[width=\linewidth]{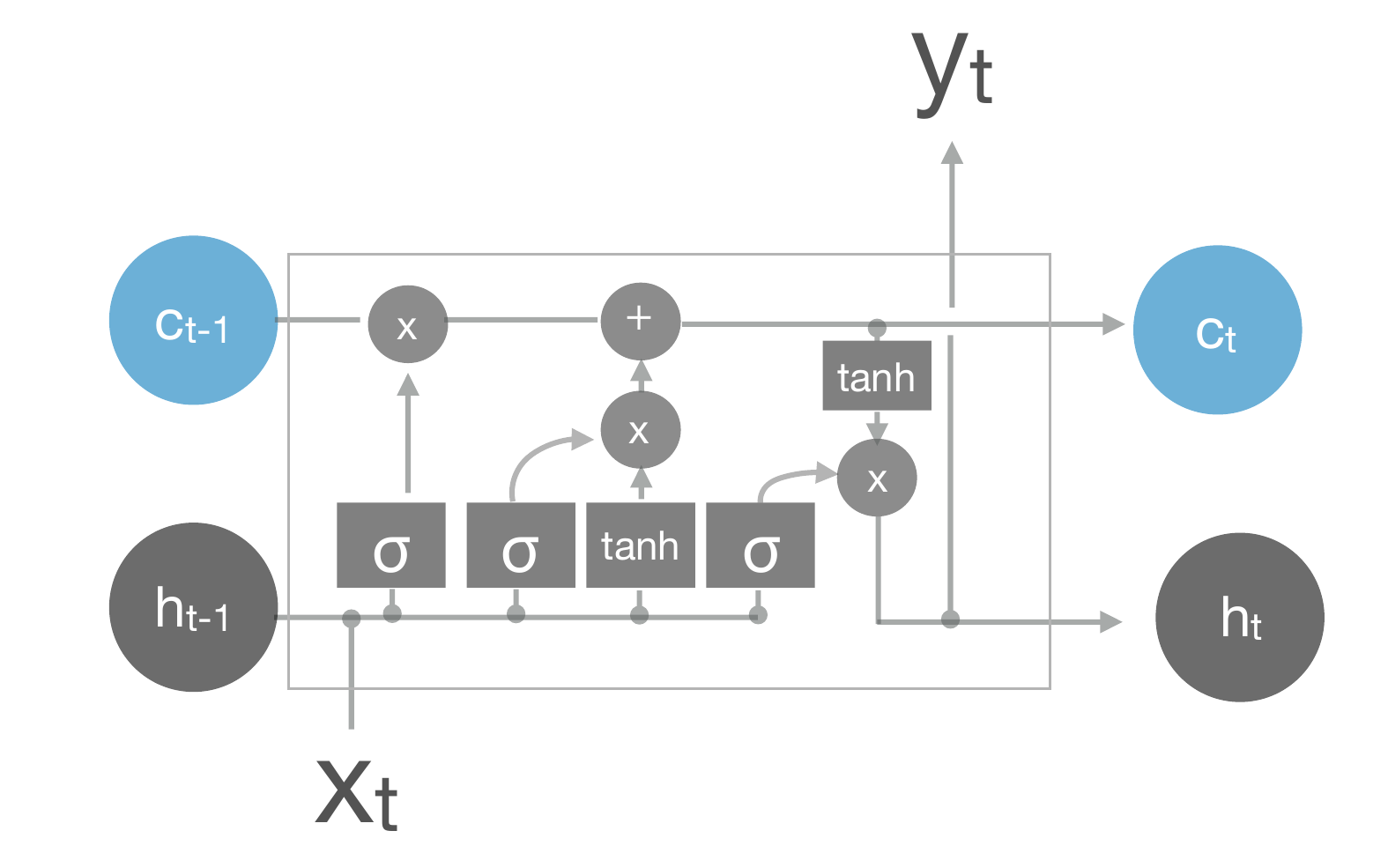}
    \caption{Visual illustration of Long Short-Term Memory cell.}
    \label{fig:figure_LSTM}
\end{figure}

There are three gates which control what is retrieved from and written to the cell state. The `forget gate' determines what portion of information from the previous time step is kept at the current time step. It is controlled by the output at the previous time step and the current input, and has a value between 0 and 1 to control information flow (See Fig. \ref{fig:figure_LSTM}). This can be written as, 
\begin{equation}
    f_t = \sigma(w^f [h_{t-1}, x_t] + b^f),
\end{equation}
where $w^f$ and $b^f$ are the gate's weights and bias, $h_{t-1}$ is the previous time step's output, $x_t$ is the current input and $\sigma$ is a sigmoid function. 

The `input gate' decides what information is written to the cell. Similar to the previous gate we have a function deciding what portion of information to write, 
\begin{equation}
    g_t = \sigma(w^i [h_{t-1}, x_t] + b^i),
\end{equation}
and we use a $\mathrm{tanh}$ function to generate the information to write,
\begin{equation}
    \tilde{c}_t = \mathrm{tanh}(w^c[h_{t-1}, x_t] + b^c).
\end{equation}
Then the cell state can be updated using,
\begin{equation}
c_t = f_t \times c_{t-1} + i_t \times \tilde{c}_t. 
\label{eq:LSTM}
\end{equation}

The information from the previous cell state and the current time step are controlled via the forget and input gate values. The final step is to decide what information to output from the cell at the current time step. This is done through the output gate, 
\begin{equation}
    o_t = \sigma(w^o [h_{t-1}, x_t] + b^o),
\end{equation}
and, $h_t$, the current time step's output is given by,
\begin{equation}
    h_t = o_t \times \mathrm{tanh}(c_t).
\end{equation}

\textit{Gated Recurrent Units (GRUs)} are a popular variant of the LSTM which were introduced by Cho et. al in 2014 \cite{cho2014learning}. They combine the forget gate and input gate into a single `update gate'. Specifically, Eq. \ref{eq:LSTM} becomes, 
\begin{equation}
c_t = f_t \times c_{t-1} + (1- f_t) \times \tilde{c}_t.
\end{equation}

\textit{Neural Memory Networks (NMNs)} are another variant of RNNs, where an external memory stack is used to store information. A limitation of LSTMs and GRUs is that content is erased when a new sequence is presented \cite{fernando2018tree, fernando2018learning}, as these architectures are designed to map temporal relationships within a sequence, not between sequences \cite{fernando2018tree, fernando2018learning, fernando2019memory, gammulle2019forecasting}. Hence, the limited capacity of the internal cell state is insufficient to model relationships across a large corpus \cite{ma2019taxonomy, fernando2018tree}. 

\begin{figure}[htbp]
    \centering
    \includegraphics[width=\linewidth]{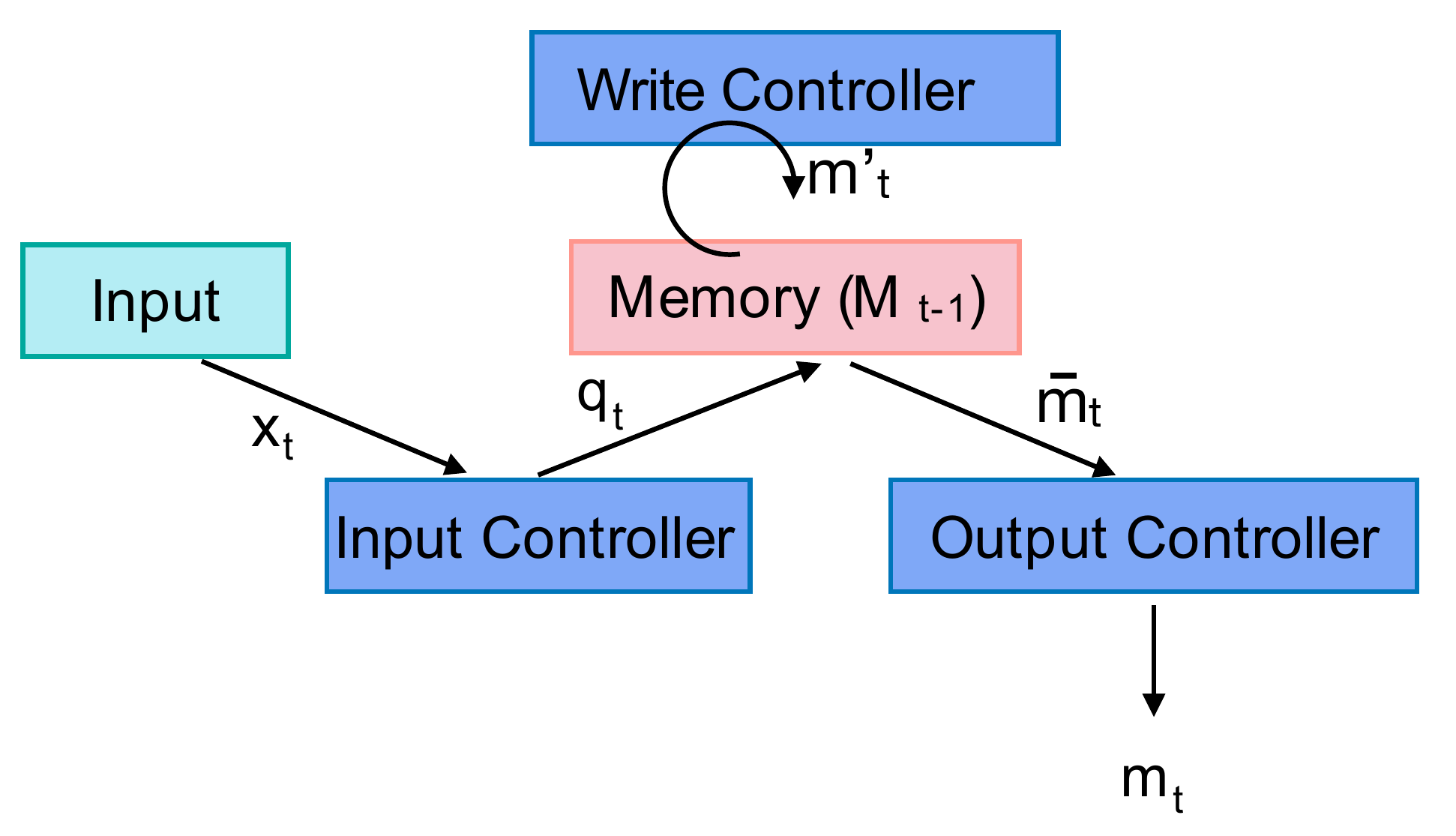}
    \caption{Illustration of the main components of a Neural Memory Network.}
    \label{fig:figure_NMN}
\end{figure}

Fig. \ref{fig:figure_NMN} illustrates a typical NMN architecture, which is composed of a memory stack to store information, and a set of controllers (read, output and write) to manipulate the memory. There are similarities between the LSTM gated operations and these controllers which manipulate the external memory. Specifically, let the state of the external memory, $M \in \mathbb{R}^{k \times l}$, at time instance $t-1$ be given by $M_{t-1}$, where $l$ is the number of memory slots and $k$ is the size of each slot. The current input is denoted $x_t$. Then the read controller, $f^r$, generates a vector, $q_t$, to query the memory,
\begin{equation}
q_t = f^r(x_t).
\end{equation}
Using a softmax function we measure the similarity between the content of each memory slot and the query vector such that, 
\begin{equation}
z_t = \textrm{softmax}({q_t}^\top M_{t-1}).
\end{equation}
The score vector, $z_t$, capture the relevance of the memory content to the current input. This parallels the input gate functionality of an LSTM, which determines what information to extract from the history. However, in contrast to the LSTM where there is only one vector storing historical information, the NMN has multiple memory blocks to consider. Hence, attention is employed to extract the most relevant information.

The output controller, $f_o$, generates the output such that,
\begin{equation}
 \Bar{m}_t=z_{t} [M_{t-1}]^\top,
\end{equation}
and, 
\begin{equation}
 m_t=f^o(\Bar{m}_t).
\end{equation}
This aligns with the output gate functionality of the LSTM, using the input at the current time-step and information retrieved from the memory to compose the output.  
Finally, the write controller, $f^w$, generates a vector to update the memory,
\begin{equation}
m'_t = f^w (m_t),
\end{equation}
and updates the memory using,
\begin{equation}
M_t = M_{t-1} (I - z_{t} \otimes e_k)^\top + (m'_t \otimes e_l) (z_{t} \otimes e_k)^\top,
\end{equation}
where $I$ is a matrix of ones, $e_l \in \mathbb{R}^l$ and $e_k \in \mathbb{R}^k$ are vectors of ones and $\otimes$ denotes the outer product which duplicates its left vector $l$ or $k$ times to form a matrix \cite{munkhdalai2017neural, fernando2020neural}. As NMNs are a relatively new concept we refer interested readers to \cite{ma2019taxonomy} for further details. 

While the exact memory update mechanisms for LSTMs and NMNs are dissimilar, we would like to highlight the parallels between the LSTM forget gate and the write controller. The LSTM forget gate considers the current time-step's input and the previous cell state (i.e memory) and determines what to pass through from the history. Similarly, the write controller, utilising the NMN output and the retrieved historical information, determines what memory slots to update. 

There are numerous works that have utilised RNNs for medical anomaly detection. For instance, RNNs have been used in \cite{jagannatha2016bidirectional} and \cite{yang2018toward} for text based abnormality detection in electronic health records; and in \cite{latif2018phonocardiographic} to detect abnormal heart beats in Phonocardiographic recordings. 

More recently, NMNs have been applied in medical anomaly detection, where works have illustrated the value of external memory storage to memorise similarities and differences between normal and anomalous examples. Specifically, in \cite{fernando2020neural} the authors couple an NMN with a neural plasticity framework to identify tumors in MRI scans and abnormalities in EEGs. Furthermore, in \cite{ahmedt2019neural} the same architecture is used to identify different seizure types in EEGs.  

\color{black}
\subsection{Background and Related Applications}
\label{sec:applications}

This subsection provides a detailed discussion of popular application domains within deep medical anomaly detection, illustrating how previously discussed architectural variants are leveraged in these domains. 

\subsubsection{MRI based Anomaly Detection}
\label{sec:MRIapplications}

In this section we summarize some of key recent literature in deep learning based anomaly detection with MRI data. Fusion of modalities has been a popular research direction which has recently emerged for MRI analysis. In \cite{yoo2018deep} the authors investigate the fusion of T1-weighted (T1w) MRIs and myelin water imaging of MRIs (which is a quantitative MRI technique that specifically measures myelin content) for diagnosis of Multiple sclerosis (MS). In their proposed architecture, they utilise two modality specific Deep Belief Networks (DBN) \cite{hinton2009deep} to extract features from the individual T1w and myelin maps. This is followed by a multi-modal DBN which jointly learns complementary information from both modes. They retrieve a multi-modal feature vector by concatenating the top-level hidden unit activations of the multimodal DBN. As the final step a Random Forest \cite{breiman2001random} is trained to detect abnormalities. The proposed algorithm is validated using an in-house dataset, which consists of 55 relapse-remitting MS patients and 44 healthy controls. The classification accuracy was $70.1 \pm 13.6 $\% and $83.8 \pm 11.0$ \% for T1w and myelin map modalities, respectively, while the fused representation achieved $87.9 \pm 8.4$ \%.

A strategy to fuse MRI images with fluorodeoxyglucose positron emission tomography (FDG-PET) samples has been proposed in \cite{lu2018multimodal}. In their approach, the authors first segment the MRI images into gray and white matter regions, followed by subdivision of the gray matter into 87 anatomical Regions Of Interest (ROI). Then they extract patches of sizes 1488, 705 and 343 from these regions. Similar size patches are also extracted from FDG-PET images. Then 6 independent Deep Neural Networks (DNNs) with dense layers are used to embed patch information and another DNN is used to fuse the encoded embedding. A softmax layer is used generate the final abnormality classification. The authors utilise this architecture to detect pathologies related to Alzheimer’s Disease and the framework is evaluated using publicly available Alzheimer’s Disease Neuroimaging Initiative (ADNI) database \cite{jack2008alzheimer}, which contains 1242 subjects. The proposed method achieves 82.93 \% accuracy and an approximately 1.5\% improvement over utilising FDG-PET alone. 

A method to fuse Apparent Diffusion Coefficients (ADCs) of MRIs together with T2-weighted MRI images (T2w) is proposed in \cite{le2017automated}. In contrast to predicting a single score level classification, they proposed a method which outputs a segmentation map for each modality, indicating the likelihood of each pixel belonging to the class of interest. They propose to utilise a novel similarity loss function such that the ADC and T2WI streams produce consistent predictions, allowing complementary information to be shared between the streams. The initial segmentation maps are combined with hand-crafted features and passed through an SVM to generate the final predictions. Evaluations were carried out using a dataset with 364 subjects, with a total of 463 prostate cancer lesions and 450 identified noncancerous image patches in which the framework achieves a sensitivity of 89.85\% and a specificity of 95.83\% for distinguishing cancerous from non-cancerous tissues.

In contrast to the above approach which employs feature level fusion, an architecture using decision level fusion in proposed in \cite{islam2018brain}. The proposed approach has an ensemble of classifiers composed of 3 convolutional neural networks which are trained separately. Each network provides a softmax classification denoting the likelihood of four Alzheimer’s disease classes: non-demented, very mild, mild and moderate. The fusion of the individual classifications is performed using majority voting. The evaluation is conducted on the public OASIS dataset \cite{marcusoasis} which consists of 416 subjects, and the proposed ensemble method achieves 94 \% precision and 93 \% recall. 

Despite the architectural differences, the above discussed methods are all supervised Deep CNN (DCNN) models and these dominate the MRI based anomaly detection literature. This is clearly motivated by the fact that supervised CNN models are highly effective when extracting task specific spatial information from two dimensional inputs. 

Despite the prevalence of supervised DCNN models, a number of approaches have also used Auto-Encoders (AE) \cite{shehata2018computer, zeng2018multi}. In \cite{zeng2018multi} an AE network with a sparsity constraint has been proposed for the diagnosis of individuals with schizophrenia. First the AE is trained in an unsupervised manner for feature extraction and in the second stage the authors use validation set of the dataset to fine tune the network after adding a softmax layer to the AE. As the final stage a linear support vector machine is used to classify samples. The system is validated using a large scale MRI dataset which is collected from 7 image sources and consists of 474 patients with schizophrenia and 607 healthy controls. This model achieves approximately 85 \% accuracy in a k-fold cross validation setting. Similarly, in \cite{shehata2018computer} an AE is trained for early detection of acute renal transplant rejection. As the first stage the AE is trained in an unsupervised manner. To classify inputs, the decoder of the AE is removed and a softmax layer is trained using supervised learning. This method achieves 97\% classification accuracy in a leave-one-subject-out experimental setting on 100 subjects. 

Critically, unlike the unsupervised AE models discussed in Sec. \ref{sec:unsupervised} these models are not purely unsupervised architectures. Rather after the preliminary training of the AE, a classification layer is added and trained using a supervised signal to detect anomalies. 

\begin{figure}[htbp]
    \centering
    \includegraphics[width=\linewidth]{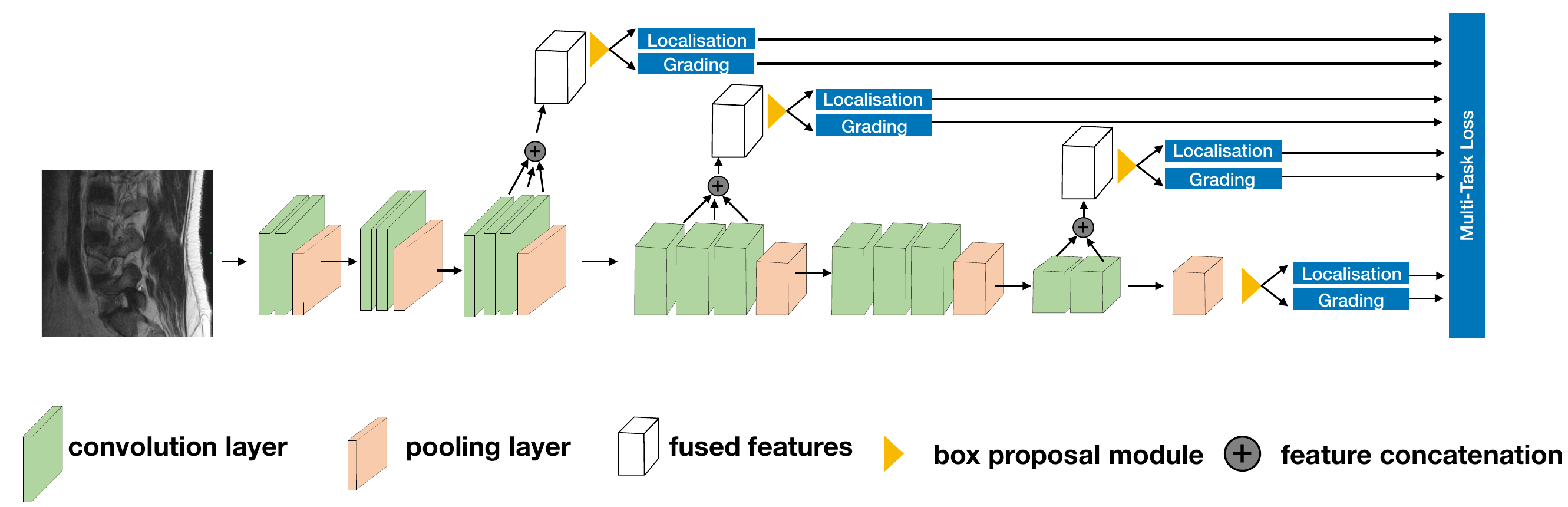}
    \caption{The architecture proposed in \cite{han2018automated} for diagnosis of Lumbar Neural Foraminal Stenosis. Recreated from \cite{han2018automated}}
    \label{fig:mri_application_model}
\end{figure}

In a different line of work, a multi-scale multi-task learning framework is proposed in \cite{han2018automated} for diagnosis of Lumbar Neural Foraminal Stenosis (LNFS). Fig. \ref{fig:mri_application_model} illustrates the architecture used. The authors show that each lumbar spine image can have multiple organs captured at various scales. Furthermore, they illustrate that multi-task learning can be used such that learning from multiple related tasks can boost learning, as discussed in Sec. \ref{sec:supervised}, and we note that this strategy is seen across multiple applications domains. Feature maps are extracted at multiple scales and at each level the model tries to perform two tasks, regression of bounding boxes to locate the organs and prediction of abnormalities in the located organs. This system is validated using 200 clinical patients and it is capable of diagnosing abnormal neural foramina with 0.83 precision and 0.8 recall. 

In addition to those approaches, a Neural Memory Network (NMN) based approach is proposed in \cite{fernando2020neural}. This method utilises the recurrent structure of NMN to compare and contrast characteristics of the samples in the entire dataset using supervised learning. The memory stack stores important characteristics that separates normal and anomalous samples. Therefore, this architecture significantly deviates from rest of the approaches already described. Specifically, a ResNet-50 CNN is used to extract a $14 \times 14 \times 256$ feature from the input MRI image. This feature becomes the input to the read controller of the NMN. Utilising this as a query vector, the read controller attends to the content that is stored in the memory, comparing them to find the best possible match to the input. The output of the memory read function and the input vector are passed though an output controller which generates the memory output (i.e. a feature vector which is subsequently used to generate the final classification). As the final step, the write controller decides how to update the content in the memory slots, reflecting the information retrieved from the current input. In addition to this typical functionality of the NMN, plasticity is utilised in the NMN controllers such that they can adapt the connectivity dynamically, changing the overall behaviour of the NMN. This framework was evaluated using the dataset of \cite{cheng2015enhanced} which contains MRI images captured from 233 patients with different types of brain tumours: meningioma (708 samples), glioma (1426 samples), and pituitary (930 samples). In the 5-fold cross validation setting the model achieves 97.52\% classification accuracy. Here we would like to point out that instead of binary normal/abnormal classification, a multi-class classification was conducted where the model discriminates between different abnormal classes using the categorical cross-entropy loss. 

\subsubsection{Detecting abnormalities in Endoscopy Data}
In this section we summarise some popular deep learning architectures introduced for abnormality detection from endoscopy's. 

Considering the fact that endoscopy devices capture RGB data, CNNs pre-trained on large scale object detection benchmarks such as Image-Net \cite{deng2009imagenet} have been extensively applied. For instance, in \cite{klang2020deep} the authors apply the Xception \cite{chollet2017xception} CNN architecture pre-trained on \cite{deng2009imagenet} for the detection of ulcers in endoscopy images. The proposed system is evaluated using a dataset that consists of 49 subjects and the authors have performed both 5-fold cross validation and a leave-one-subject out evaluation. The system achieves an average of 96.05 \% accuracy in the 5-fold cross validation setting, while the performance varies between 73.7\% to 98.2\% in the leave-one-subject out evaluation. Similarly in \cite{alaskar2019application} both GoogLeNet \cite{szegedy2015going} and AlexNet \cite{krizhevsky2012imagenet} pre-trained networks have been investigated for the classification of ulcers. The models were tested on a public dataset \cite{alaskar2019application} which consists of 1875 images. Both models achieve 100\% accuracy in this dataset.  Furthermore, in \cite{fan2018computer} AlexNet \cite{krizhevsky2012imagenet} has been applied for both ulcer and erosion detection. The resultant model is capable of achieving 95.16\% and 95.34\% accuracy levels for ulcer and erosion detection when tested on 500 ulcer and 690 erosion images.

In contrast to these architectures, a two stage approach is proposed in \cite{wang2019systematic}. RetinaNet \cite{lin2017focal} has been adapted for the initial detection stage where it receives an endoscopy image and predicts the classification scores and bounding boxes for the input image. Then they extract multiple fixed size patches of size $ 160 \times 160$ from this image and pass those through a ResNet-18 \cite{he2016deep} network where the final fully connected layer produces a binary classification for the detection of ulcers. This system has been tested with 4917 ulcer frames and 5007 normal frames and the model reaches 0.9469 ROC-AUC value. 

\begin{figure}[htbp]
    \centering
    \includegraphics[width=\linewidth]{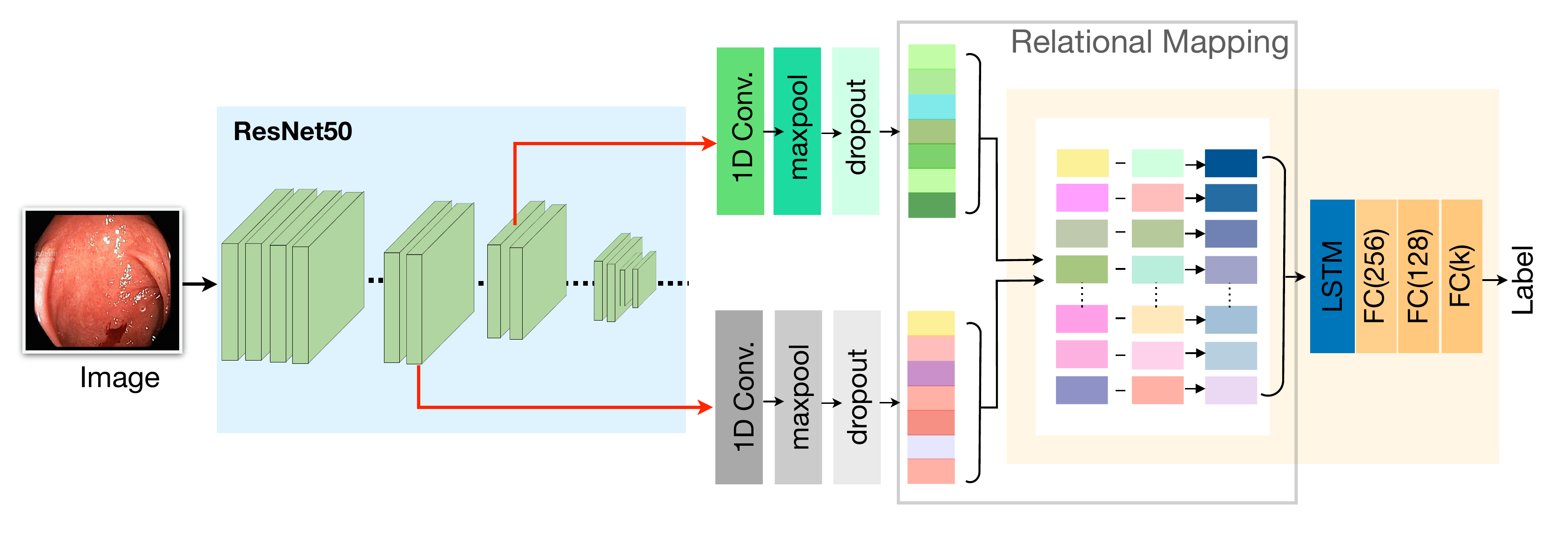}
    \caption{The architecture proposed in \cite{gammulle2020two} for abnormality detection in endoscopy data. Recreated from \cite{gammulle2020two}}
    \label{fig:endoscopy_application_model}
\end{figure}

Most recently, a two stream framework has been proposed in \cite{gammulle2020two} where the authors extract features from two levels of the ResNet-50 \cite{he2016deep} architecture, and they are combined using a relational network \cite{santoro2017simple}. Fig. \ref{fig:endoscopy_application_model} illustrates this method. Specifically, the relational network allows this approach to map all possible relationships among the features extracted at the two levels. The resultant augmented feature vector is passed through an LSTM network and classification is performed using a fully connected layer. This framework has been evaluated on two public benchmarks, Kvasir \cite{pogorelov2017kvasir} (with 8000 endoscopy images) and Nerthus \cite{pogorelov2017nerthus} (with 2,552 colonoscopy images). In the Kvasir dataset this system was able to detect 8 abnormality classes with 98.4 \% accuracy, and reaches 100 \% accuracy for classifying the cleanliness of the bowel on the Nerthus dataset. We note that this study exploits the hierarchical nature of the CNN to address the requirements in endoscopy image analysis. Top level kernels of a CNN capture local spatial features such as textures and contours in the input while the bottom level layers capture more semantic features, such as the overall representation of the image. This is because the local features are pooled together, hierarchically, when they flow through the CNN. Therefore, when extracting features from a CNN, top level layers carry spatially variant characteristics of the input while the bottom layers have spatially invariant features. The authors in \cite{gammulle2020two} leverage this characteristic of CNNs for endoscopy image analysis in which, both existence of a particular distinctive pattern as well as its location is vital for diagnosis.  

Similar to MRI image analysis, DCNNs have dominated endoscopy image abnormality detection. Furthermore, most methods utilise pre-trained feature extractors trained on large scale datasets with natural images leveraging the fact that endoscopy images are also captured with visible light. In contrast to binary supervised classification, methods such as  \cite{gammulle2020two} and \cite{pogorelov2017kvasir} use multi-class classification (often trained using categorical cross-entropy loss) such that the model can detect normal and abnormal examples while recognising individual anomalies. 

\subsubsection{Heart Sound Anomaly Detection}

In contrast to the MRI and endoscopy applications which use images, heart sound anomaly detection operates on a one-dimensional audio signal, and methods primarily use 1D CNNs and RNN architectures, however some pre-processing methods can be used to transform the audio signal to an image representation, allowing 2D CNNs to be used. An ensemble of VGG \cite{simonyan2014very} networks is proposed in \cite{wu2019applying}, where the authors first apply a Savitzky–Golay filter \cite{savitzky1964smoothing} to remove noise from the input signals. Then a series of 2D features, a spectrogram feature, a Mel Spectrogram and Mel Frequency Cepstral Coefficients (MFCCs), are extracted from the audio signal. These separate feature streams are passed through separate VGG networks and the final decision is made via majority voting. This method has also been evalauted on the PhysioNet/CinC 2016 dataset in a 10 fold cross validation setting and it reaches an accuracy of 89.81 \%.

In \cite{li2020classification} the authors leverage 497 feature values which are hand-crafted from 8 domains, including time domain features, higher-order statistics, signal energy, and frequency domain features. The extracted features are concatenated and passed through a 1D CNN with 3 convolutional layers followed by a global average pooling layer and a dense layer with a sigmoid activation to perform the normal/abnormal classification. This system is evaluated on PhysioNet/CinC 2016 dataset and achieves an accuracy value of 86.8 \%. 

In contrast to the above stated approaches, frameworks that operate on raw audio signals are proposed in \cite{yang2016classification, gao2020gated}. Specifically, in \cite{yang2016classification} the authors augment the raw audio signal from the PhysioNet/CinC 2016 dataset by performing a Discrete Fourier Transform (DFT) and adding the variance and standard deviation of each data sample to the original audio. Then the recordings are segmented into S1 and S2 heart states using the algorithm of \cite{springer2015logistic}. The segmented recordings are passed through an RNN to validate its normality. This framework achieves 80 \% accuracy on the PhysioNet/CinC 2016 challenge. A similar approach utilising GRUs has been proposed in \cite{gao2020gated}. Similar to \cite{yang2016classification} the raw audio recordings were segmented to heart states using the algorithm of \cite{springer2015logistic}. However, the authors in \cite{gao2020gated} skip the DFT based heart sound augmentation step utilised in \cite{yang2016classification}. The segmented audio is passed through a GRU network to generate the classification. The proposed framework has been validated for heart failure detection. The authors have acquired the heart failure data from patients in University-Town Hospital of Chongqing Medical University and the normal recordings were obtained from PhysioNet/CinC 2016 dataset (1286 randomly sampled normal recordings). In a 10-fold cross-validation setting the proposed model achieves an average accuracy of 98.82\%. In this paper the authors have also tested the utilisation of an LSTM and Fully Convolutional Network (FCN) instead of a GRU network, however, these models have only been able to achieve 96.29 \% and 94.65 \%, respectively. 

There has been a mixed response from researches regarding the need for heart sound segmentation prior to the abnormal heart sound detection. Heart sound segmentation has primarily been used due to the belief that features surrounding the S1 and S2 heart sound locations carry important information for detecting abnormalities. However, some argue that in the errors associated with this pre-processing step can be propagated to the abnormality detection module, and the model should be given the freedom to choose it's own informative features \cite{dissanayake2020understanding}. In \cite{dissanayake2020understanding} the authors have conducted a comparative study investigating the importance of prior segmentation of heart sounds into heart states for abnormality detection. The authors have utilised the features extracted from the state-of-the-art the sound segmentation model \cite{fernando2019heart} and trained a classifier to detect abnormalities using these features. For comparison, they also trained a separate 2D CNN model without segmentation which uses MFCC features as the inputs. The comparisons were conducted using the PhysioNet/CinC 2016 dataset and their evaluation indicates that a 2D CNN model without segmentation is capable of achieving superior results to a model that receives segmented inputs. In the 10-fold cross validation setting the unsegmeted model achieves $98.94 \pm 0.27 $ \% accuracy compared to $ 98.49 \pm 0.13$ \% for the segmented model. Utilising the SHAP model interpretations \cite{molnar2020interpretable} the authors conclude that the unsegmented model has also focused on the regions of the audio wave that correspond to S1 and S2 locations, however, this model has the capacity to learn what the informative features for the abnormality detection task are, compared to the restricted model inputs that are received by the segmented model. 

\begin{figure}[htbp]
    \centering
    \includegraphics[width=\linewidth]{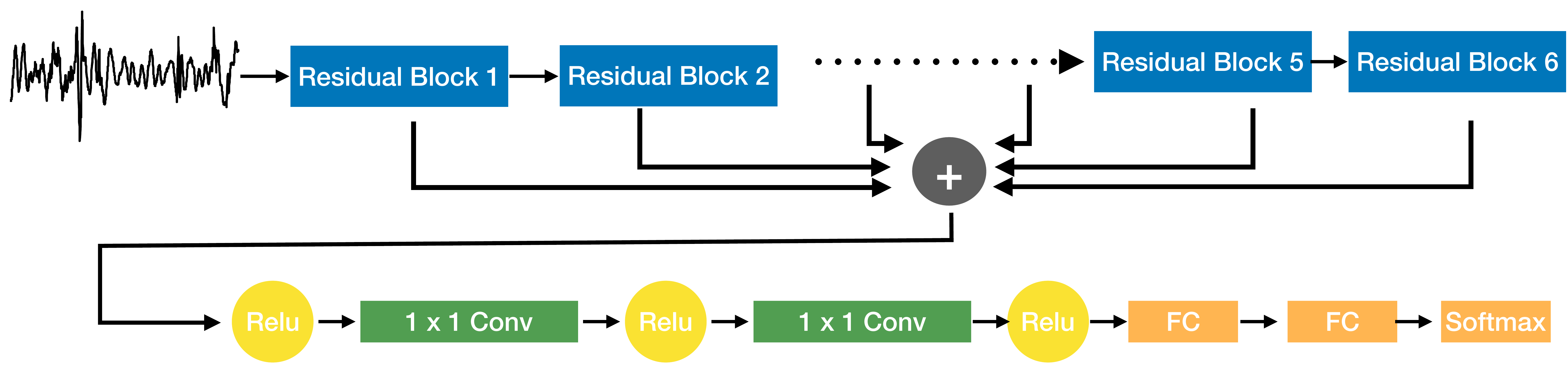}
    \caption{The architecture proposed in \cite{oh2020classification} for abnormal heart sound detection. Recreated from \cite{oh2020classification}}
    \label{fig:application_heart_sound}
\end{figure}

Finally, Oh et. al \cite{oh2020classification} proposed a deep learned model called WaveNet to classify heart sounds into five categories, namely: normal, aortic stenosis, mitral valve prolapse, mitral stenosis, and mitral regurgitation. The architecture utilised in this study is illustrated in Fig. \ref{fig:application_heart_sound}. Specifically, inspired by \cite{he2016deep} the authors proposed a residual block which is composed of 1D dilated convolutions to extract features from the raw audio signal. The architecture is composed of 6 such residual blocks and the features captured from those 6 blocks are aggregated into a single feature vector, which is subsequently passed through two 1D convolution layers and a two fully connected layers, prior to classification. This model is evaluated using an in house dataset which consists of 1000 PCG recording (200 per each category) and the model achieves an average accuracy of 97 \% in a 10-fold cross validation setting. 

As noted earlier, 1D-CNN networks and RNNs have been extensively applied for the abnormal heart sound detection. This is primarily due to the temporal nature of the signal where 1D-CNN networks can perform convolutions over the time axis and extract temporal features while the recurrent architectures can model the temporal evolution of the signal and generate better features for detecting the abnormalities. As discussed, there are only minor variations among the models and they have often utilised supervised learning to train the models. Furthermore, hand-crafted frequency domain features such as MFCCs are extensively applied within the heart sound anomaly detection domain as opposed to automatic feature learning. Finally, as observed in other application domains, supervised approaches are the most common methods for anomaly detection.

\subsubsection{Epileptic Seizure Prediction}

\begin{figure}[htbp]
  \centering
  \includegraphics[width=\linewidth]{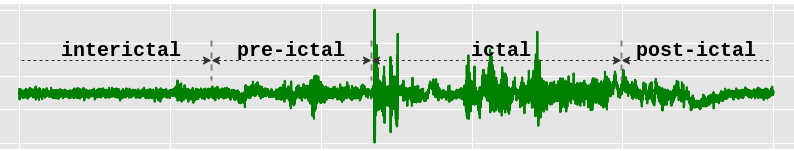}
    \caption{Variations of the EEG recording before and after a seizure.}
    \label{fig:EEG}
\end{figure}
Fig. \ref{fig:EEG} illustrates how the four brain states: interictal, pre-ictal , ictal and post-ictal; are located in an EEG. The interictal state is the normal brain state of a subject, while the brain state before a seizure event is refereed to as the pre-ictal state. The state in which the seizure occurs is denoted as ictal state, and after the seizure event, the brain shifts to the post-ictal state.

The seizure prediction problem can be viewed as an abnormality detection problem where machine learning models are trained to distinguish between the pre-ictal and interictal brain states, identifying when a particular subjects brain activity shifts from the normal interictal state to pre-ictal (abnormal state). As the pre-ictal state is the brain state before a seizure, this problem is termed seizure prediction. 

We acknowledge that epileptic seizure prediction has several distinct characteristics compared to rest of the abnormality detection application domains that we discussed above, however, numerous studies have posed this task as an abnormality detection task \cite{gadhoumi2012discriminating, tsiouris2018long, thara2019epileptic}, and hence we consider it here. 

A key challenge in designing a generalised seizure prediction framework is the vast differences in the pre-ictal duration among subjects. This can vary from minutes to hours depending on the subject \cite{Khan2018FocalNetworks}. One of the notable attempts to perform patient independent seizure prediction is the framework of \cite{truong2017generalised}, where the authors propose a 2D CNN architecture trained on Short-term Fourier transform (STFT) features extracted from raw EEG signals. This framework has been validated on both the Freiburg intracranial EEG (iEEG) \cite{winterhalder2003seizure} and CHB-MIT scalp EEG (sEEG) datasets \cite{shoeb2009application}, and achieves approximately 81 \% sensitivity in a leave-one-subject-out cross validation setting.

Despite this promising level of performance, the authors in \cite{hussein2019human, kiral2018epileptic} identified significant performance variations in \cite{truong2017generalised}. For example, the sensitivity drops to 33.3 \% for some subjects. A multi-scale CNN architecture is proposed in \cite{hussein2019human} to address this limitation. The authors re-sample the original 400Hz iEEG dataset at 100Hz and STFT features are extracted from this down sampled signal. They extract STFT as 2D images for each EEG channel, resulting in 16 STFT images per data sample. The proposed multi-scale CNN is composed of 3 convolutional streams, each with different filter sizes ($1 \times 1$, $3 \times 3$ and $5 \times 5$). The authors propose to capture features at different scales using these individual streams. These features are concatenated and passed through a fully connected layer to generate the relevant predictions. Their system is evaluated on 2016 Kaggle seizure prediction competition dataset \cite{kuhlmann2018epilepsyecosystem} and the proposed system achieves a 87.85 \% sensitivity where the lowest value per subject is only 79.65 \%.

In contrast to this approach, a fine-tuning based method is proposed in \cite{kiral2018epileptic}. The authors first train the model using a balanced dataset which consists of an equal amount of pre-ictal and interictal data. When the system is deployed, the authors propose to add a tunable processing layer which can be optimised depending on the patient requirements. This two stage framework is evaluated using the dataset proposed in \cite{cook2013prediction} and the system achieve a mean sensitivity of 69\%. 

In contrast to these CNN based approaches, recurrent neural networks are leveraged in \cite{tsiouris2018long, thara2019epileptic}. Specifically, the authors in \cite{tsiouris2018long} utilised a 2 layer LSTM network trained on hand-crafted time domain, frequency domain, graph theory based (i.e clustering coefficients, diameter, radius, local efficiency, centrality, etc.), and correlation features. The system is evaluated using the CHB-MIT sEEG dataset and reaches 99.28\% sensitivity for a 15 min pre-ictal period. Motivated by this approach, a bi-directional LSTM based architecture is given in \cite{thara2019epileptic}. Similar to \cite{tsiouris2018long}, a 2-layer LSTM is used with a bi-directional structure, however, in contrast to \cite{tsiouris2018long} it operates on the raw EEG signal. This framework has been validated using the Bonn University EEG database \cite{andrzejak2001indications} and achieves an overall 89.2 \% sensitivity score. 

Compared to heart sound anomaly detection, most existing works in seizure prediction have utilised DCNN architectures. This is mainly due to the use of hand-crafted 2D image like features which are extracted jointly by considering all EEG electrodes. Once again, supervised learning methods are most prevalent and the architectures comprise standard deep learning methods.

\begin{figure}[htbp]
    \centering
    \includegraphics[width=.65\linewidth]{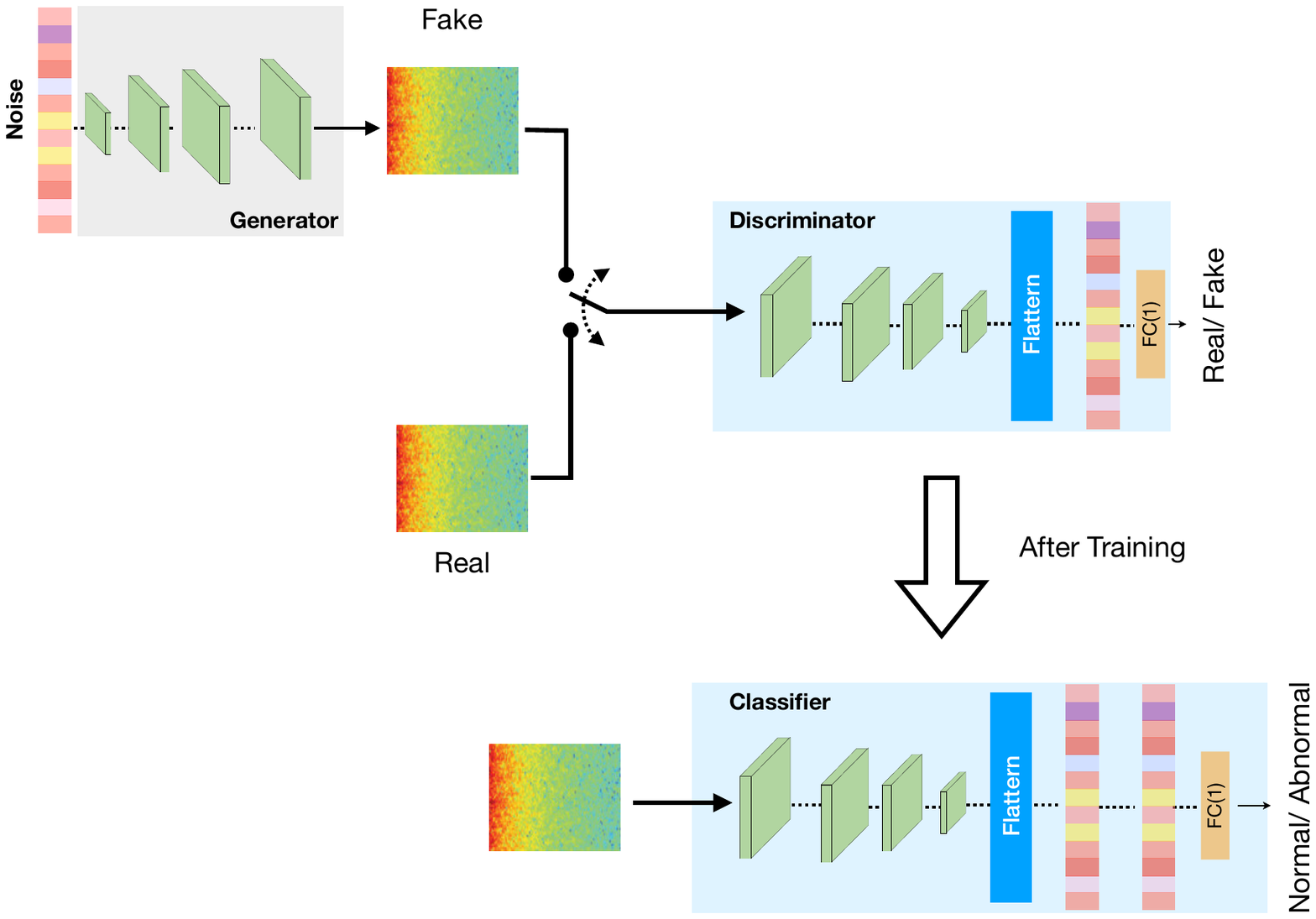}
    \caption{The architecture proposed in \cite{truong2019epileptic} for epileptic seizure prediction. Recreated from \cite{truong2019epileptic}}
    \label{fig:application_sizure_prediction}
\end{figure}

A different approach is proposed by \cite{truong2019epileptic}, who propose a GAN based method which is illustrated in Fig. \ref{fig:application_sizure_prediction}. The generator of the GAN model is capable of synthesising realistic looking STFT images using a noise vector. The generated STFTs are passed through the discriminator which performs the real fake validation. Once the generator is trained for the seizure prediction task, the authors adapt the discriminator network by adding two fully-connected layers such that it is trained to perform the normal/abnormal classification instead of real/fake classification. Therefore, the proposed system leverages the information in not only labeled EEG signals, but also the unlabeled synthesised samples in the training process. This system is validated using CHB-MIT sEEG, Freiburg iEEG, and EPILEPSIAE \cite{ihle2012epilepsiae} datasets, and achieves AUC values of 77.68\%, 75.47\% and 65.05\%, respectively.

We highlight that this approach deviates from the standard GAN model illustrated in Sec. \ref{sec:unsupervised}, as in this model a secondary training process is used where the discriminator is fine-tuned to do normal/abnormal classification using supervised learning. Hence, like the autoencoder methods discussed for MRI anomaly detection in Section \ref{sec:MRIapplications}, this is not a completely unsupervised model. Rather this architecture is semi-supervised, where both labelled and unlabelled examples are used for model training \cite{gammulle2020fine}. 
{In Tab. \ref{tab:application_summary} we provide a comprehensive summary regarding key research in these different application type, their evaluation details, results and limitations. }

\begin{table*}[htbp]
\caption{Summary of key research in different applications, their evaluation results, and limitations}
{
\resizebox{\textwidth}{!}{
\begin{tabular}{|p{0.15\textwidth}|c|p{0.3\textwidth}|p{0.3\textwidth}|p{0.3\textwidth}|p{0.3\textwidth}|p{0.4\textwidth}|}
\hline
Application Type  & Reference & Method & Task & Dataset & Results & Limitations and Research Gaps \\ \hline
\vspace{10mm}\parbox[t]{2mm}{\multirow{3}{*}{\rotatebox[origin=c]{90}{MRI based Anomaly Detection}}} &  \cite{yoo2018deep}         &  Fusion of T1-weighted MRIs and myelin water imaging of MRIs       & Diagnosis of Multiple sclerosis     &  in-house (55 relapse-remitting MS patients and 44 healthy controls)    &   $87.9 \pm 8.4$ \% accuracy      &  Simple feature concatenation is used for fusion,  Not an end-to-end learning framework, Causality or Uncertainty were not investigated.                             \\ \cline{2-7}  

                  &  \cite{lu2018multimodal}         & Fusion of MRI and FDG-PET       & Detection of Alzheimer’s Disease     & Alzheimer’s Disease Neuroimaging Initiative (ADNI) database \cite{jack2008alzheimer}        &   82.93 \% accuracy      & Simple feature concatenation is used for fusion,  Lack of Interpretability, Two stages of training is required, Causality or Uncertainty were not investigated.                               \\ \cline{2-7} 
                  
                  &  \cite{le2017automated}         & Fusion of Apparent Diffusion Coefficients of MRIs and T2-weighted MRI images       & Detection of prostate cancer lesions     & in-house (463 cancer lesions and 450 noncancerous images)       &   Sensitivity 89.85\% and Specificity 95.83\%      & Heavily reliant on pre-processing, Lack of Interpretability, Causality or Uncertainty were not investigated.                               \\ \cline{2-7} 
                  
                  &  \cite{islam2018brain}         & Decision level fusion using ensemble of classifiers       & Classification between four Alzheimer’s disease classes (non-demented, very mild, mild and moderate)     & OASIS dataset \cite{marcusoasis}       &   94 \% precision and 93 \% recall     & Higher computation cost due to the use of ensemble of classifiers, Lack of Interpretability, Causality or Uncertainty were not investigated.                               \\ \cline{2-7}
                  
                  &  \cite{zeng2018multi}         & Multi-stage training of Auto-Encoders       & Schizophrenia Diagnosis     & In-house (474 schizophrenia and 607 healthy controls)       &   85 \% accuracy     & High development time due to multiple stages of training, Heavily reliant on pre-processing,  Lack of Interpretability, Causality or Uncertainty were not investigated.                               \\ \cline{2-7}
                  
                  &  \cite{shehata2018computer}         & Multi-stage training of Auto-Encoders       & Early detection of acute renal transplant rejection     & In-house (100 subjects)      &   97\% accuracy     & 3D segmentation maps are required as inputs, Lack of Interpretability, Causality or Uncertainty were not investigated.                               \\ \cline{2-7}
                  
                   &  \cite{han2018automated}         & Multi-scale Multi-task learning algorithm       & Diagnosis of Lumbar Neural Foraminal Stenosis     & In-house (200 subjects)      &  83 \% precision and 80 \% recall    & Evaluation can be slower due to the multi-stage pipeline, Causality or Uncertainty were not investigated.                               \\ \cline{2-7}
                   
                   &  \cite{fernando2020neural}         & Neural Memory Network       & Classification of brain tumours     & The dataset of \cite{cheng2015enhanced}. Meningioma (708), glioma (1426), pituitary (930)      &  97.52\% accuracy    & Computationally expensive and data intensive due to the usage of memory network, Lack of Interpretability, Causality or Uncertainty were not investigated.                               \\ \hline
                   
\vspace{10mm}\parbox[t]{10mm}{\multirow{3}{*}{\rotatebox[origin=c]{90}{Detecting abnormalities in Endoscopy Data}}} &  \cite{klang2020deep}         &  Fine-tuning Xception       & Detection of ulcers     &  in-house (49 subjects)    &    96.05 \% accuracy      &  Cannot identify different types of ulcers, Lack of Interpretability, Causality or Uncertainty were not investigated.                        \\ \cline{2-7} 

                    &  \cite{alaskar2019application}         & Fine-tuning GoogLeNet and AlexNet       & Detection of ulcers     & The dataset of \cite{alaskar2019application}      &  100\% accuracy    & Cannot identify different types of ulcers, Lack of Interpretability, Causality or Uncertainty were not investigated.                               \\ \cline{2-7}
                    
                     &  \cite{fan2018computer}          & Fine-tuning AlexNet       & ulcer and erosion detection    & in-house (500 ulcer and 690 erosion images)      &  95.16\% and 95.34\% accuracy levels for ulcer and erosion detection    & Lack of Interpretability, Causality or Uncertainty were not investigated.                               \\ \cline{2-7}
                     
                     &  \cite{wang2019systematic}          & Two-stage approach using RetinaNet       & detection of ulcers    & in-house ( 4917 ulcer frames and 5007 normal frames)      &  0.9469 ROC-AUC    & Cannot identify different types of ulcers, Computationally expensive due to the two-stage approach, Lack of Interpretability.                              \\ \cline{2-7}
                     
                     &  \cite{gammulle2020two}          & Two stream framework using ResNet-50       & Classifying abnormalities, Classifying the cleanliness of the bowel   & Kvasir \cite{pogorelov2017kvasir} (with 8000 endoscopy images) and Nerthus \cite{pogorelov2017nerthus} (with 2,552 colonoscopy images)      &  98.4 \%  accuracy when detecting 8 abnormality classes, and 100 \% accuracy for classifying the cleanliness of the bowel on the Nerthus dataset    & Computationally expensive due to the relational network architecture, Lack of Interpretability, Causality or Uncertainty were not investigated.                              \\ \hline
                     
\vspace{10mm}\parbox[t]{10mm}{\multirow{3}{*}{\rotatebox[origin=c]{90}{Heart Sound Anomaly Detection}}} &  \cite{wu2019applying}         &  Ensemble of VGG networks      & Abnormal heart sound detection     &  PhysioNet/CinC 2016    &    89.81 \% accuracy     & Heavily reliant on pre-processing, Computationally expensive due to ensemble of models, Lack of Interpretability, Causality or Uncertainty were not investigated.                        \\ \cline{2-7} 

                    &  \cite{li2020classification}          & 1D CNN using a combination of time, statistical, energy, and frequency domain features       & Abnormal heart sound detection    & PhysioNet/CinC 2016      &  86.8 \% accuracy     &  Hand-engineered features, Lack of Interpretability, Causality or Uncertainty were not investigated.                              \\ \cline{2-7}
                    
                    &  \cite{yang2016classification}          & RNN $+$ DFT based heart sound augmentation $+$ segmentation        & Abnormal heart sound detection    & PhysioNet/CinC 2016      &  80 \% accuracy     &  Heavily reliant on pre-processing, Lack of Interpretability, Causality or Uncertainty were not investigated.                              \\ \cline{2-7}
                    
                    &  \cite{gao2020gated}          & GRU $+$ segmentation        & Heart failure detection    & University-Town Hospital of Chongqing Medical University      &  98.82\% accuracy      &  Heavily reliant on pre-processing, Lack of Interpretability, Causality or Uncertainty were not investigated.                              \\ \cline{2-7}
                    
                    &  \cite{dissanayake2020understanding}          & 2D CNN using MFCC features and without segmentation        & Abnormal heart sound detection    & PhysioNet/CinC 2016      &  $98.94 \pm 0.27 $ \% accuracy      &  Causality or Uncertainty were not investigated.                              \\ \cline{2-7}
                    
                    &  \cite{oh2020classification}          & Residual block with raw audio        & classification of heart sounds (normal, aortic stenosis, mitral valve prolapse, mitral stenosis, and mitral regurgitation)    & in-house dataset which consists of 1000 recordings.     &  97 \% accuracy      &  Causality or Uncertainty were not investigated.                              \\ \hline
                    
\vspace{10mm}\parbox[t]{10mm}{\multirow{3}{*}{\rotatebox[origin=c]{90}{Epileptic Seizure Prediction}}} &  \cite{truong2017generalised}         &  2D CNN architecture trained on Short-term Fourier transform (STFT) features      & Seizure prediction     &  Freiburg intracranial EEG (iEEG) \cite{winterhalder2003seizure} and CHB-MIT scalp EEG (sEEG) datasets \cite{shoeb2009application}    &    81 \% sensitivity     & Heavily reliant on pre-processing, Lack of Interpretability, Causality or Uncertainty were not investigated.            \\ \cline{2-7} 

                    &  \cite{hussein2019human}          & Multi-scale CNN        & Seizure prediction    & 2016 Kaggle seizure prediction competition dataset \cite{kuhlmann2018epilepsyecosystem}      &  87.85 \% sensitivity      &  Heavily reliant on pre-processing, Computationally expensive due to multi-scale architecture, Lack of Interpretability, Causality or Uncertainty were not investigated.                            \\ \cline{2-7}
                    
                    &  \cite{kiral2018epileptic}          & Tunable layer for patient based fine-tuning        & Seizure prediction    & dataset of \cite{cook2013prediction}      &  69\% sensitivity      &  Requires patient-specific data for deployment, Lack of Interpretability, Causality or Uncertainty were not investigated.                            \\ \cline{2-7}
                    
                    &  \cite{tsiouris2018long}          & 2 layer LSTM trained on time, frequency, graph theory, and correlation features        & Seizure prediction    & CHB-MIT sEEG dataset      &  99.28\% sensitivity     &   Hand-engineered features, Lack of Interpretability, Causality or Uncertainty were not investigated.                            \\ \cline{2-7}
                    
                     &  \cite{thara2019epileptic}          & bi-directional LSTM using raw signal        & Seizure prediction    & Bonn University EEG database \cite{andrzejak2001indications}      &  89.2 \% sensitivity     &   Lack of Interpretability, Causality or Uncertainty were not investigated.                           \\ \cline{2-7}
                     
                     &  \cite{truong2019epileptic}         & Synthesising STFT images using a GAN        & Seizure prediction    & CHB-MIT sEEG, Freiburg iEEG, and EPILEPSIAE \cite{ihle2012epilepsiae} datasets      &  AUC values of 77.68\%, 75.47\% and 65.05\%, respectively.     &  High development time due to multiple stages of training, Lack of Interpretability, Causality or Uncertainty were not investigated.                           \\ \hline

\end{tabular}
}}
\label{tab:application_summary}
\end{table*}

\section{Model Interpretation}
\label{sec:interpretation}
Interpretability is one of the key challenges that modern deep learning methods face. Despite their tremendous success and often astonishingly precise predictions, the application of methods to real world diagnostic tasks is hindered as we are unsure how models reached their predictions. The complexity of the deep learned models further contributes to this, as decisions are based upon hundreds of thousands of parameters, which are not human interpretable. Hence, interpretable machine learning has become an active area of research where black-box deep models are converted white-box models. 

\begin{figure}[htbp]
    \centering
    \includegraphics[width=.85 \linewidth]{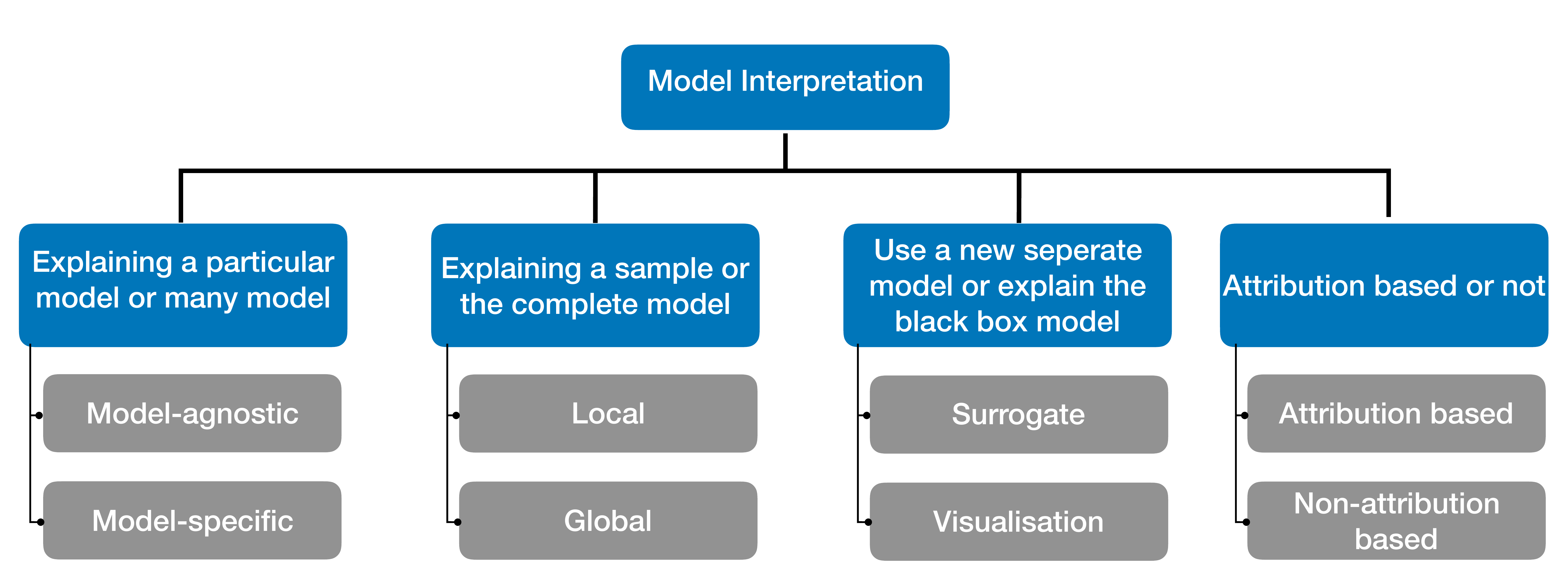}
    \caption{Taxonomy of Model Interpretation methods.}
    \label{fig:xai}
\end{figure}

Fig. \ref{fig:xai} illustrates a taxonomy of model interpretation methods, which is adopted in \cite{singh2020explainable}. Model-agnostic interpretation methods are interpretation methods that are not limited to a specific architecture. In contrast, a model-specific interpretation method seeks to explain a single model. 

Model interpretation methods can be further classified into local and global methods. Local methods try to reason regarding a particular prediction while global methods explore overall model behaviour by exploiting knowledge regarding the architecture, the training process and the data associated with training. The third class we consider is surrogate vs. visualization methods. In surrogate methods, a model with a simpler architecture (a surrogate model) is trained to mimic the behaviour of the original black box model. This is done with the intent that understanding the surrogate model's decision is simpler than the complex model. In contrast, visualisation methods use visual representations such as activation maps obtained from the black-box model to explain behaviour. Finally, as per \cite{singh2020explainable} model interpretation techniques in the medical domain can be broadly categorised into attribution based and non-attribution based methods. Attribution-based methods seek to determine the contribution of an input feature to the generated classification. Non-attribution based methods investigate generating new methods to validate model behaviour for a given problem \cite{singh2020explainable}.
 
The majority of existing literature on explainability of deep learned models in the medical domain considers attribution based methods. They leverage the model-agnostic plug and play nature of attribution based methods in their studies. The following paragraphs illustrate the most common model-agnostic interpretability methods that are used.

\textbf{Visualising Activation Maps}: This offers one of the simplest ways to understand what features lead to a certain model decision. As deep learning methods hierarchically encode features, the top layers of the model capture local features while later layers aggregate local features together to arrive at a decision. This concept is the foundation of Class Activation Maps (CAM) \cite{zhou2016learning}. 

\begin{figure}[htbp]
    \centering
    \includegraphics[width=.95\linewidth]{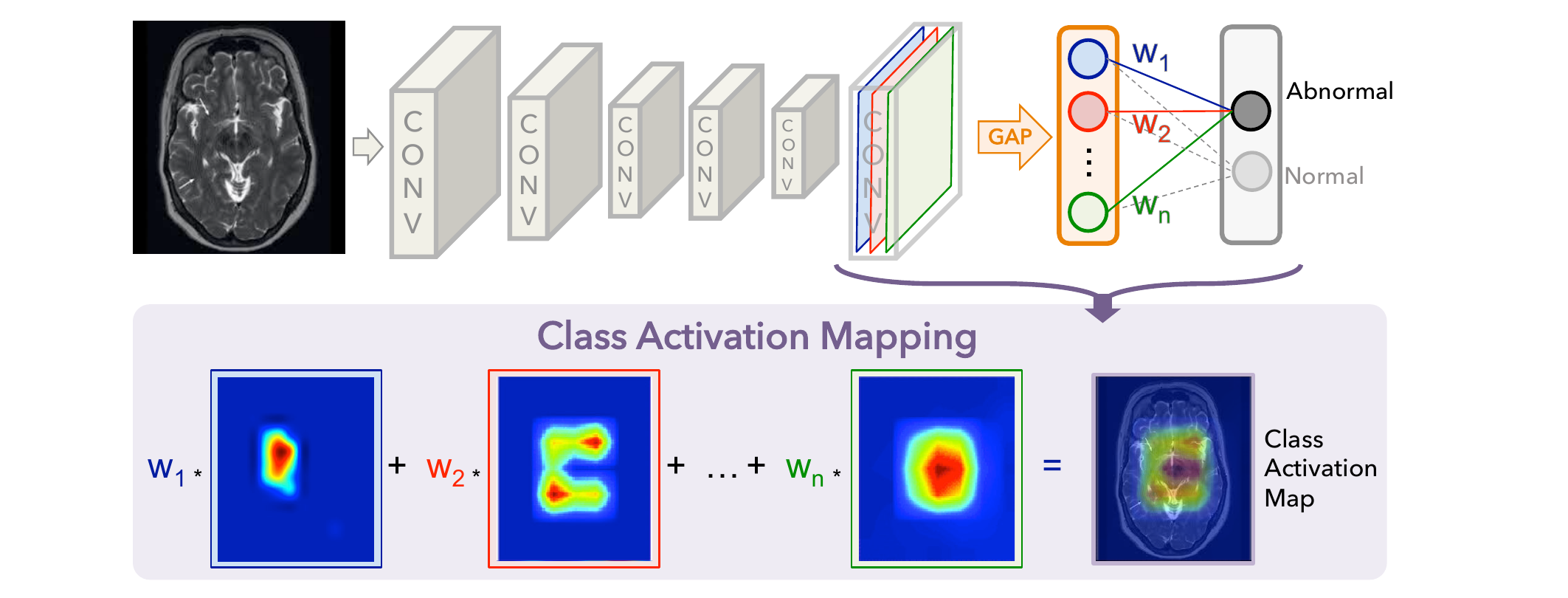}
    \caption{Illustration of the process of creating class activation maps. Recreated from \cite{zhou2016learning}}
    \label{fig:figure_class_activation_maps}
\end{figure}

We can consider kernels in a convolution layer to be a set of filters which control information flow to subsequent layers, and at the final classification layer the positive features (emphasised features from the filters) are multiplied by learned values to obtain a classification decision. Fig. \ref{fig:figure_class_activation_maps} illustrates this concept. Hence the activation maps, or feature maps extracted at the final convolution layer, are multiplied by the associated weights and they are aggregated to generate the final activation map of the predicted class. The resultant map is up-sampled such that it can be superimposed on the input image. This can reveal what regions/characteristics of the input are highly activated and pass information to the classifier. Such a technique can be applied for tasks such as CNN based MRI tumor detection to identify whether the features from the tumor region are actually contributing to the classification, or if the model is acting upon noisy features from elsewhere in the sample. 

One of the drawbacks of the CAM generation process is that the technique is constrained to architectures where global average pooling is performed over convolutional maps immediately before prediction. This is a requirement for utilising the weighted liner sum of the activations to generate the final convolution map. 

The Grad-Cam \cite{selvaraju2017grad} method addresses this shortcoming, and uses the gradient information flowing into the last convolutional layer of the network to understand how each pixel contributes to the decision. Let the $k^th$ feature map  of the final convolution layer of size $u \times v$ be denoted by $A^k$. Then the gradients of the score for class $c$, $y^c$, are computed with respect to feature maps $A^k$, and averaged across $u \times v$ such that,
\begin{equation}
\alpha^{c, k}=\frac{1}{uv}\sum_{i \in u}\sum_{j \in v}{\frac{\partial y^c}{\partial A^k_{i,j}}}
\end{equation}
Then to generate the final activation map across all $k$ feature maps a weighted combination of activation maps is computed. The resultant feature map is passed through the ReLU activation to set negetive values to zero.

\begin{equation}
L_{Grad-Cam}=\mathrm{ReLU}(\sum_{k}{ \alpha^{c, k} A^k})
\end{equation}

Grad-Cam however does not handle instances where multiple occurrences of the same object are in the input image, and in such instances it fails to properly localise the multiple instances \cite{chattopadhay2018grad}. 

\begin{figure}[htbp]
    \centering
    \includegraphics[width=.75\linewidth]{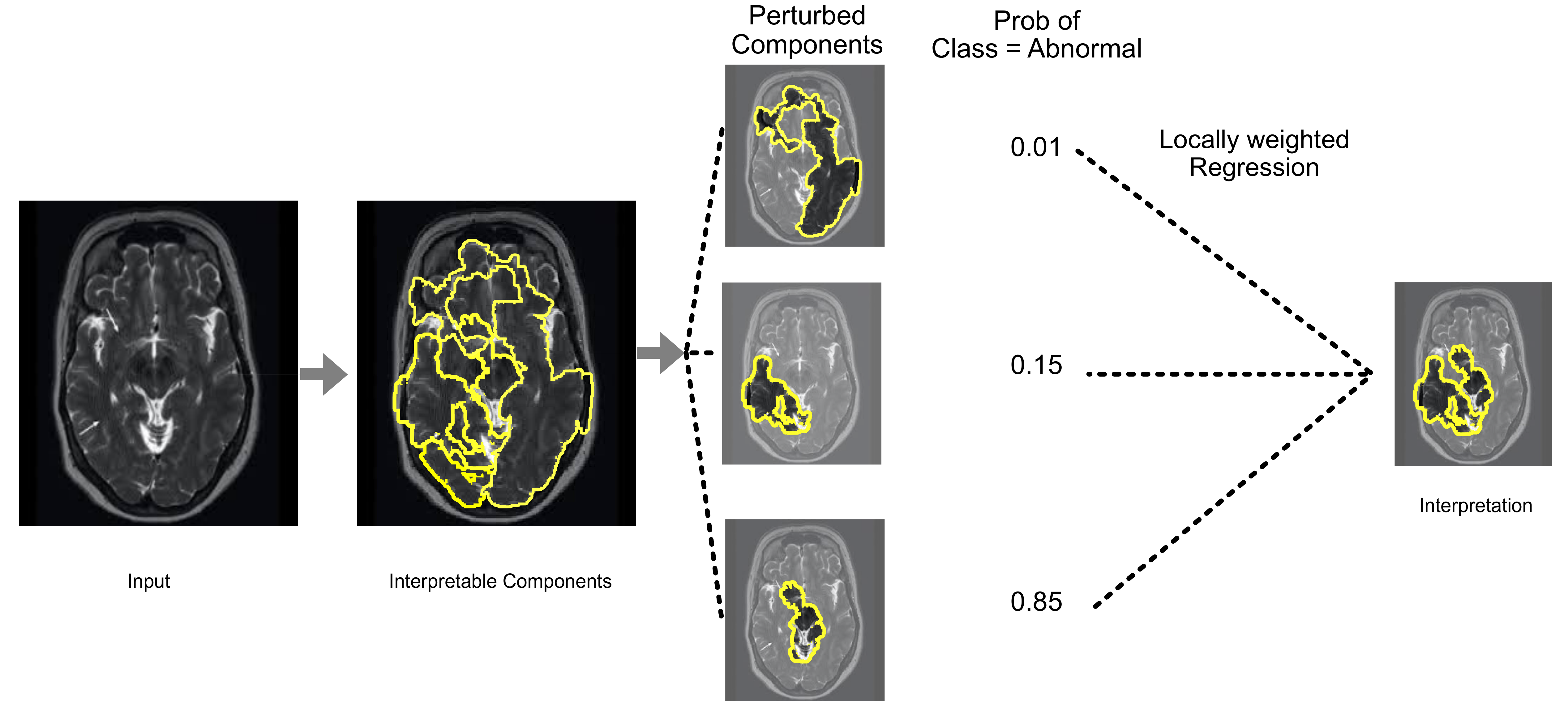}
    \caption{Illustration of Local Interpretable Model-agnostic Explanations method. Recreated from \cite{ribeiro2016should}}
    \label{fig:lime}
\end{figure}

\textbf{Local Interpretable Model-agnostic Explanations (LIME)}: As the name implies, LIME is a local interpretation method. It tries to interpret a model's behaviour when presented with different inputs by understanding how predictions change with perturbations of the input data. 
Fig. \ref{fig:lime} illustrates the concept behind LIME. First, the input is divided to a series of interpretable components where some portion of the input is masked out. Then each perturbed sample is passed through the model to get the probability of a particular class and the components of the image with the highest weights are returned as the explanation.

One of the key limitations of LIME is that when sampling the data points for interpretable components, it can sample unrealistic data points. Furthermore, as the interpretations are biased towards these data points the generated explanations can be unstable such that two components in close proximity can lead to very different explanations. 

\textbf{SHapley Additive exPlanations (SHAP)}: The inspiration for SHAP \cite{lundberg2017unified} comes from game theory. Specifically, we define individual feature values (or groups of feature values) as a `player' in the game and the contribution of each player to the game is measured by adding and removing the players. Let the input $x$ be composed of $N$ features, $\omega_i$s, where $x = [\omega_1, \omega_2, \ldots, \omega_N]$, and $M$ is the maximum number of coalitions (or feature combinations) that can be generated using $N$ features. Then the model is queried with different feature coalitions where some feature values are present and some are absent. (eg. $x_1 = [\omega_1], x_2 = [\omega_2], x_{1,2} = [\omega_1, \omega_2], x_{1,3} = [\omega_1, \omega_3], \ldots$). This allows the identification of which features contribute to a certain prediction and which do not.

\begin{figure}[htbp]
    \centering
    \includegraphics[width=\linewidth]{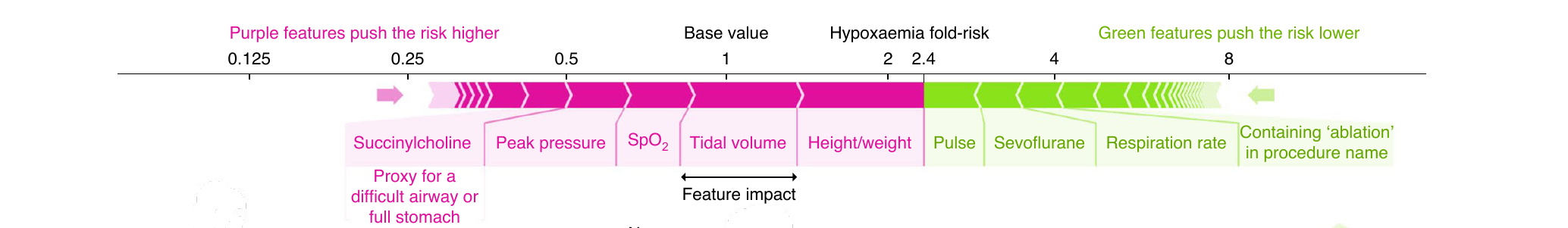}
    \caption{Illustration of SHapley Additive exPlanations which explore how different features contribute to the risk of hypoxaemia. Image taken from \cite{lundberg2018explainable}}
    \label{fig:shap}
\end{figure}
 
Fig. \ref{fig:shap} illustrates how different features such as height/weight, respiration rate and pulse affect the risk of hypoxaemia in the next five minutes. Features shown in purple increase the risk while green features reduce the risk. 

In \cite{molnar2020interpretable} the author suggests that SHAP is one of the few explanation method with a strong theoretical basis and the only method that currently exists to deliver a full interpretation. However, it is computationally expensive to calculate SAHPly values as we have to consider all possible combinations of the features. 

\textbf {Interpretation of Medical Anomaly Detection Methods:}
In addition to the above stated commonly used interpretation methods we acknowledge the methods DeepLIFT (Deep Learning Important FeaTures) \cite{shrikumar2017learning}, DeepTaylor \cite{montavon2017explaining}, Guided backpropagation (GBP) \cite{springenberg2014striving} and Integrated Gradients \cite{sundararajan2017axiomatic}, that are also proposed to explain the black-box deep learning model. 

In \cite{pereira2018automatic} the authors attempt to explain the features learned by a CNN which they proposed for automated grading of brain tumors from MRI images using GradCAM and GBP techniques. In \cite{young2019deep} 30 CNN models were trained for melanoma detection using skin images and the authors interpret the model features using SHAP and GradCAM. They illustrate that models occasionally focus on features that were irrelevant for diagnosis. 

In recent studies \cite{tsiknakis2020interpretable, chatterjee2020exploration} GradCAM, GBP, CAM, DeepLIFT and IG have been utilised to explain chest X-ray based COVID-19 detection of a deep learned model. Most recently, SHAP interpretations are used to illustrate that a heart sound anomaly detection methods can automatically learn to focus on S1 and S2 sounds without the need to provide segmented inputs \cite{dissanayake2020understanding}.

In contrast to the attribution based approaches illustrated earlier, non-attribution based methods use concepts like attention maps, and expert knowledge to interpret model decisions. However, these methods are specific to a particular problem. For instance, in \cite{zhang2017mdnet} attention is used to map the relationships between medical images and corresponding diagnostic reports. This mechanism uncovers the mapping between images and the diagnostic reports. A textual justification for a breast mass classification task is proposed in \cite{lee2019generation}. The proposed justification model receives visual features and embeddings from the classifier and generates a diagnostic sentence explaining the model classification. In \cite{zhu2019guideline} a method for generating understandable explanations utilising a set of explainable rules is presented. In this approach, a set of anatomical features are defined based on segmentation and anatomical regularities (i.e set of pre-defined rules), and the feature importance is evaluated based on perturbation. 

Considering the above discussion it is clear that the selection of the interpretation method depends on several factors:
\begin{enumerate}
    \item whether a global level model interpretation method is required, or whether it is sufficient to generate local (example level) interpretations;
    \item the end users expertise level with regards to understanding the resultant explanations; and
    \item whether the application domain has time constraints, i.e. do the interpretations need to be generated in real-time?
\end{enumerate}

Interpretable machine learning is an active area of research and the medical domain is a good test bed to evaluate the proposed methods. Better understanding regarding the black-box model decisions would not only build trust among the medical practitioners regarding machine learning algorithms, but also would help the machine learning researchers to understand the limitations of model architectures and help to design better models. However, as illustrated earlier, different interpretation approaches have different strengths and limitations, and designing optimal interpretation strategies is an open research problem for future exploration.

\section{Challenges and Open Research Questions}
\label{sec:challenges_and_future}

In this section, we outline limitations of existing deep medical anomaly detection techniques as well as various open research questions, and highlight future research directions.

\subsection{Lack of Interpretability}
As illustrated in Sec. \ref{sec:interpretation}, attribution based methods have been popular among researchers in the medical domain for deep model interpretation due to their model agnostic plug-and-play nature. However, the end-user of the given particular medical application (i.e. the clinician) should be considered when selecting one interpretation method over another. Although popular, methodologies such as GradCAM, LIME, and GBP are not specifically developed to address explainability in the medical domain, and while they are informative for machine learning practitioners, they may be of much less use for a clinicians. Therefore, more studies such as \cite{arbabshirani2018advanced, almazroa2017agreement} should be conducted using expert clinicians to rate the explanations across different application domains. Such illustrations would evaluate the applicability and the limitations of model-agnostic interpretation methods. Hybrid techniques such as Human-in-the-Loop learning techniques could be utilised to design interpretable diagnostic models where clinical experts could refine deep model decisions to mimic their own decision making process. 

Furthermore, we observe a lack of model-agnostic methods to interpret multi-modal deep methodologies. Such methods have increased complexity in that the decision depends on multiple input feature streams, requiring more sophisticated strategies to interpret behaviour.   

Finally, we present Reinforcement Learning (RL) as a possible future direction to generate explainable decisions \cite{huang2018towards, oh2019sequential}. In RL the autonomous agent's behaviour is governed by a `reward function', and the agent tries to maximise this reward. As such the agent utilises exploration to detect anomalies and improve its detection process across many iterations. The exploration process that the agent utilised to detect the anomalies could illustrate the intuition behind it's behaviour. 

\subsection{Causality and Uncertainty}
Causal identification is crucial characteristic that most existing deep medical anomaly detection methods lack. Causality is often confused with association \cite{richens2020improving, castro2020causality, altman2015association}. For instance, if $X$ and $Y$ are associated (dependent) it only implies that there is a dependency between the two factors. The association does not imply that $X$ is causing $Y$. Association can arise between variables in the presence and absence of a causal relationship. If both $X$ and $Y$ have a common cause they both can show associative relationships without causality \cite{altman2015association}. 

In medical diagnosis the doctor tries to explain the cause of the patient's symptoms, which is causal identification. However, in most existing deep learned approaches the diagnosis is purely associative. Methods try to associate the patient's symptoms with a particular disease category without trying to uncover what is actually causing these symptoms (and whether this disease is the only cause of these symptoms) \cite{richens2020improving}. As such, causality estimation is a crucial area that requires additional focus from the research community. For instance, in epilepsy prediction if the the brain regions that are actually causing the seizures can be identified then epileptologists can surgically treat that specific region to address the root cause of the patient's seizures. Existing approaches for causality estimation in deep learning studies include causal graph structures \cite{nauta2019causal}, algorithmic information theory based approaches \cite{zenil2019algorithmic, zenil2019causal}, and Causal Bayesian Networks \cite{zhang2019bayesian, pal2019bayesian}; however, these methods are seldom applied in medical abnormality detection. 

Uncertainty estimation is another characteristic that most current state-of-the-art anomaly detection algorithms lack. Such methods quantitatively estimate how a small change in input parameters affects model predictions. This can be indicative of model confidence. For instance, Bayesian Deep Learning \cite{kendall2017uncertainties} could be used to generate probabilistic scores, where the model parameters are approximated through a variational method, generating uncertainty information regarding the model weights, such that one can observe the uncertainty of the model outputs. We would like to acknowledge the work of Leibig et. al \cite{leibig2017leveraging} where they illustrate how the computed measure of uncertainty can be utilised to refer a subset of difficult cases for further inspection. Furthermore, Bayesian uncertainty estimation has been applied for estimating the quality of medical image segmentation \cite{kwon2020uncertainty, devries2018leveraging, seebock2019exploiting}, and sclerosis lesion detection \cite{nair2020exploring}. We believe further extensive investigation will allow rapid application of uncertainty estimation measure in the medical anomaly detection algorithms. 

\subsection{Lack of Generalisation}

\begin{figure}[htbp]
    \centering
    \includegraphics[width=.7\linewidth]{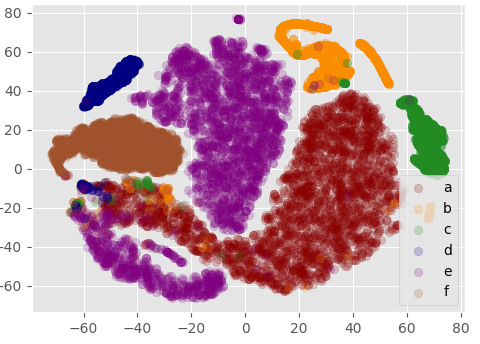}
    \caption{Visualisation of t-SNE plots for different domains in PhysioNet/CinC 2016 dataset abnormal heart sound detection challenge. Image taken from \cite{dissanayake2020domain}}
    \label{fig:Lack_of_Generalisation}
\end{figure}

Another limitation that we observe in present research is the lack of generalisation to different operating conditions. For instance, \cite{dissanayake2020domain} observed an abnormal heart sound detection model that achieves more than 99 \% accuracy drop to 52.27 \% when presented with an unseen dataset. Such performance instability significantly hinders the applicability of the deep learned models in real world life-or-death medical applications. One of the major reasons for such specificity of the models is data scarcity in the medical domain. Even though the number of datasets that are publicly available continues to increase, there are still a limited number of data samples available (compared to large scale datasets such are ImageNet). Most datasets are also highly curated, collected in controlled environments and restricted settings that do not capture the real data distribution. For instance, in Fig. \ref{fig:Lack_of_Generalisation} we visualise the t-SNE plot generated for the model in \cite{dissanayake2020domain} using 2,000 randomly chosen samples (which contain both normal and abnormal samples) from PhysioNet/CinC 2016 dataset. This dataset is composed of 6 sub datasets (denoted a-f in the figure), collected from different devices (Welch Allyn Meditron, 3M Littmann, and ABES Electronic stethoscope), different capture environments (Lab setting and hospitals), varying age groups, etc. 

As illustrated in Figure \ref{fig:Lack_of_Generalisation}, the samples of each subset are somewhat grouped together while the samples from different subsets are distributed across the embedding space. This example clearly illustrates the challenges associated with medical anomaly detection as the acquired samples may not optimally capture the population characteristics. If a diagnostic model is trained only on a particular sub-set of this dataset it would generate erroneous detections on another. Therefore, large scale datasets which capture the diverse nature of the full population are required. 

While large scale datasets akin to ImageNet are ideal, it is very expensive and sometimes not practically feasible to collect large scale annotated datasets in medical diagnostic research \cite{hu2018meta, mahajan2020meta}. Therefore, meta-learning and domain adaptation approaches could be of use when annotated examples are scarce. In particular, meta-learning, which is a sub-field of transfer learning, focuses on how a model trained for a particular task can be quickly adapted to a novel task. Hence, the learned knowledge is shared between the two tasks. Meta-learning is an emerging technique in the medical domain \cite{zhang2019metapred, mahajan2020meta, hu2018meta}, and could be extensively utilised to train large scale models using limited data samples. In contrast to meta-learning, domain adaptation focuses on how a generalised model trained for the same task can be adapted to a particular sub-domain (such as subsets a-f in Fig. \ref{fig:Lack_of_Generalisation}) \cite{dissanayake2020domain}. Such approaches can also utilised to attain generalisation in medical anomaly detection such that a model trained on a specific domain can be adapted to other domains using few labeled examples. 

\subsection{Handling Data Imbalance and Unlabelled Data}

The majority of existing public medical abnormality detection benchmarks are highly imbalanced in terms of normal and abnormal sample counts. In most scenarios it is comparatively easy to obtain normal samples compared to anomalous samples, yielding imbalanced datasets. This typically becomes an issue in supervised training as the model becomes more sensitive to the loss arising from the majority class, compared to classes with fewer examples. The most common approaches to address class imbalance in medical anomaly detection has been data re-sampling (under or over sampling) and cost sensitive training where more weight in the loss is assigned to the minority class \cite{johnson2019survey}. 

However, data augmentation strategies such using GANs have recently emerged which are capable of generating synthetic data for training, and they are favoured over traditional methods for handling data imbalance \cite{zhang2018imbalanced, mullick2019generative}. For instance, in \cite{zhou2018gan, zhang2018imbalanced, mullick2019generative} the generator is used to synthesise realistic-looking minority class samples, thereby balancing the class distribution and avoiding overfitting. Despite their superior results compared to traditional methods, generating realistic looking data samples is an open research problem \cite{mullick2019generative}. Further research is required to improve the quality of the synthesised samples and to determine effective GAN learning strategies that can better adapt to novelties in the abnormal (which is typically the minority) class.

Another interesting research direction for investigation is methods to handle unlabelled data. In most scenarios it is cheaper to obtain unlabelled data compared to labelled samples. Hence, if the training mechanism can leverage information in unlabelled samples, it could be highly beneficial. The sub-field of semi-supervised learning addresses this situation and GANs have also demonstrated tremendous success in a semi-supervised setting \cite{gammulle2020fine} where the trained discriminator is adapted to perform the normal abnormal classification task, instead of real/fake validation \cite{truong2019epileptic}. However, we observe that deep medical anomaly detection methods rarely utilise semi-supervised learning strategies. Hence, further investigation should be carried out to introduce such strategies into the medical domain. 

In addition to semi-supervised learning, self-supervised learning is another new research direction with significant potential. In contrast to semi-supervised learning, self-supervised learning considers learning from internal cues. In particular, it uses preliminary tasks such as context prediction \cite{doersch2015unsupervised}, colorization \cite{zhang2016colorful}, and design a jigsaw puzzle game \cite{noroozi2016unsupervised} to pre-train the model such that it learns about the data distribution. Most importantly, these pretext tasks do not require labelled data and the objectives are designed to generate labels automatically. Then, the learned knowledge is transferred to different downstream tasks such as image classification, object detection, and action recognition.

Recent works have investigated self-supervised learning for anomaly detection. For instance, in \cite{ali2020self} the authors investigate the objective of predicting the indices of randomly permuted video frames as the self-supervised objective. The authors show that implicitly reasoning about the relative positions of the objects and their motions, which is beneficial to detect abnormal behaviour. However, we observe that self-supervised learning has not yet emerged into the medical anomaly detection domain. We observe the potential of utilising pretext tasks such as medical image segmentation, artificially synthesising rotated images as self-supervised objectives in this filed. Therefore, further investigations can be carried out to assess the viability of such techniques. 


\section{Conclusion}
\label{sec:conclusion}
In this survey paper, we have discussed various approaches across deep learning-based medical anomaly detection. In particular, we have outlined different data capture settings across different medical applications, numerous deep learning architectures that have been motivated due to these different data types and problem specifications, and various learning strategies that have been applied. This structured analysis of deep medical anomaly detection research methodologies enabled comparing and contrasting existing state-of-the-art techniques despite their application differences. Moreover, we provided a comprehensive overview of deep model interpretation strategies, outlining the strengths and weaknesses of those interpretation mechanisms. As concluding remarks, we outlined key limitations of existing deep medical anomaly detection techniques and proposed possible future research directions for further investigation.

\bibliographystyle{IEEEtran}
\bibliography{ref}

\begin{thebibliography}{100}
\providecommand{\url}[1]{#1}
\csname url@samestyle\endcsname
\providecommand{\newblock}{\relax}
\providecommand{\bibinfo}[2]{#2}
\providecommand{\BIBentrySTDinterwordspacing}{\spaceskip=0pt\relax}
\providecommand{\BIBentryALTinterwordstretchfactor}{4}
\providecommand{\BIBentryALTinterwordspacing}{\spaceskip=\fontdimen2\font plus
\BIBentryALTinterwordstretchfactor\fontdimen3\font minus
  \fontdimen4\font\relax}
\providecommand{\BIBforeignlanguage}[2]{{%
\expandafter\ifx\csname l@#1\endcsname\relax
\typeout{** WARNING: IEEEtran.bst: No hyphenation pattern has been}%
\typeout{** loaded for the language `#1'. Using the pattern for}%
\typeout{** the default language instead.}%
\else
\language=\csname l@#1\endcsname
\fi
#2}}
\providecommand{\BIBdecl}{\relax}
\BIBdecl

\bibitem{fernando2020neural}
T.~Fernando, S.~Denman, D.~Ahmedt-Aristizabal, S.~Sridharan, K.~R. Laurens,
  P.~Johnston, and C.~Fookes, ``Neural memory plasticity for medical anomaly
  detection,'' \emph{Neural Networks}, 2020.

\bibitem{thudumu2020comprehensive}
S.~Thudumu, P.~Branch, J.~Jin, and J.~J. Singh, ``A comprehensive survey of
  anomaly detection techniques for high dimensional big data,'' \emph{Journal
  of Big Data}, vol.~7, no.~1, pp. 1--30, 2020.

\bibitem{chalapathy2019deep}
R.~Chalapathy and S.~Chawla, ``Deep learning for anomaly detection: A survey,''
  \emph{arXiv preprint arXiv:1901.03407}, 2019.

\bibitem{zhao2019robust}
Z.~Zhao, S.~Cerf, R.~Birke, B.~Robu, S.~Bouchenak, S.~B. Mokhtar, and L.~Y.
  Chen, ``Robust anomaly detection on unreliable data,'' in \emph{2019 49th
  Annual IEEE/IFIP International Conference on Dependable Systems and Networks
  (DSN)}.\hskip 1em plus 0.5em minus 0.4em\relax IEEE, 2019, pp. 630--637.

\bibitem{pogorelov2017kvasir}
K.~Pogorelov, K.~R. Randel, C.~Griwodz, S.~L. Eskeland, T.~de~Lange,
  D.~Johansen, C.~Spampinato, D.-T. Dang-Nguyen, M.~Lux, P.~T. Schmidt
  \emph{et~al.}, ``Kvasir: A multi-class image dataset for computer aided
  gastrointestinal disease detection,'' in \emph{Proceedings of the 8th ACM on
  Multimedia Systems Conference}, 2017, pp. 164--169.

\bibitem{clifford2016classification}
G.~D. Clifford, C.~Liu, B.~Moody, D.~Springer, I.~Silva, Q.~Li, and R.~G. Mark,
  ``Classification of normal/abnormal heart sound recordings: The
  physionet/computing in cardiology challenge 2016,'' in \emph{2016 Computing
  in Cardiology Conference (CinC)}.\hskip 1em plus 0.5em minus 0.4em\relax
  IEEE, 2016, pp. 609--612.

\bibitem{alahmadi2016different}
N.~Alahmadi, S.~A. Evdokimov, Y.~J. Kropotov, A.~M. M{\"u}ller, and
  L.~J{\"a}ncke, ``Different resting state eeg features in children from
  switzerland and saudi arabia,'' \emph{Frontiers in human neuroscience},
  vol.~10, p. 559, 2016.

\bibitem{gornitz2013toward}
N.~G{\"o}rnitz, M.~Kloft, K.~Rieck, and U.~Brefeld, ``Toward supervised anomaly
  detection,'' \emph{Journal of Artificial Intelligence Research}, vol.~46, pp.
  235--262, 2013.

\bibitem{beam2018big}
A.~L. Beam and I.~S. Kohane, ``Big data and machine learning in health care,''
  \emph{Jama}, vol. 319, no.~13, pp. 1317--1318, 2018.

\bibitem{hochreiter1997long}
S.~Hochreiter and J.~Schmidhuber, ``Long short-term memory,'' \emph{Neural
  computation}, vol.~9, no.~8, pp. 1735--1780, 1997.

\bibitem{chung2014empirical}
J.~Chung, C.~Gulcehre, K.~Cho, and Y.~Bengio, ``Empirical evaluation of gated
  recurrent neural networks on sequence modeling,'' \emph{arXiv preprint
  arXiv:1412.3555}, 2014.

\bibitem{rasheed2020machine}
K.~Rasheed, A.~Qayyum, J.~Qadir, S.~Sivathamboo, P.~Kwan, L.~Kuhlmann,
  T.~O'Brien, and A.~Razi, ``Machine learning for predicting epileptic seizures
  using eeg signals: A review,'' \emph{arXiv preprint arXiv:2002.01925}, 2020.

\bibitem{li2020review}
S.~Li, F.~Li, S.~Tang, and W.~Xiong, ``A review of computer-aided heart sound
  detection techniques,'' \emph{BioMed Research International}, vol. 2020,
  2020.

\bibitem{du2019review}
W.~Du, N.~Rao, D.~Liu, H.~Jiang, C.~Luo, Z.~Li, T.~Gan, and B.~Zeng, ``Review
  on the applications of deep learning in the analysis of gastrointestinal
  endoscopy images,'' \emph{IEEE Access}, vol.~7, pp. 142\,053--142\,069, 2019.

\bibitem{lundervold2019overview}
A.~S. Lundervold and A.~Lundervold, ``An overview of deep learning in medical
  imaging focusing on mri,'' \emph{Zeitschrift f{\"u}r Medizinische Physik},
  vol.~29, no.~2, pp. 102--127, 2019.

\bibitem{soffer2020deep}
S.~Soffer, E.~Klang, O.~Shimon, N.~Nachmias, R.~Eliakim, S.~Ben-Horin,
  U.~Kopylov, and Y.~Barash, ``Deep learning for wireless capsule endoscopy: a
  systematic review and meta-analysis,'' \emph{Gastrointestinal Endoscopy},
  2020.

\bibitem{ebrahimi2020review}
Z.~Ebrahimi, M.~Loni, M.~Daneshtalab, and A.~Gharehbaghi, ``A review on deep
  learning methods for ecg arrhythmia classification,'' \emph{Expert Systems
  with Applications: X}, p. 100033, 2020.

\bibitem{singh2014magnetoencephalography}
S.~P. Singh, ``Magnetoencephalography: basic principles,'' \emph{Annals of
  Indian Academy of Neurology}, vol.~17, no. Suppl 1, p. S107, 2014.

\bibitem{lardone2018mindfulness}
A.~Lardone, M.~Liparoti, P.~Sorrentino, R.~Rucco, F.~Jacini, A.~Polverino,
  R.~Minino, M.~Pesoli, F.~Baselice, A.~Sorriso \emph{et~al.}, ``Mindfulness
  meditation is related to long-lasting changes in hippocampal functional
  topology during resting state: a magnetoencephalography study,'' \emph{Neural
  plasticity}, vol. 2018, 2018.

\bibitem{jacini2018amnestic}
F.~Jacini, P.~Sorrentino, A.~Lardone, R.~Rucco, F.~Baselice, C.~Cavaliere,
  M.~Aiello, M.~Orsini, A.~Iavarone, V.~Manzo \emph{et~al.}, ``Amnestic mild
  cognitive impairment is associated with frequency-specific brain network
  alterations in temporal poles,'' \emph{Frontiers in aging neuroscience},
  vol.~10, p. 400, 2018.

\bibitem{rucco2019mutations}
R.~Rucco, M.~Liparoti, F.~Jacini, F.~Baselice, A.~Antenora, G.~De~Michele,
  C.~Criscuolo, A.~Vettoliere, L.~Mandolesi, G.~Sorrentino \emph{et~al.},
  ``Mutations in the spast gene causing hereditary spastic paraplegia are
  related to global topological alterations in brain functional networks,''
  \emph{Neurological Sciences}, vol.~40, no.~5, pp. 979--984, 2019.

\bibitem{hasasneh2018deep}
A.~Hasasneh, N.~Kampel, P.~Sripad, N.~J. Shah, and J.~Dammers, ``Deep learning
  approach for automatic classification of ocular and cardiac artifacts in meg
  data,'' \emph{Journal of Engineering}, vol. 2018, 2018.

\bibitem{lopez2020detection}
M.~Lopez-Martin, A.~Nevado, and B.~Carro, ``Detection of early stages of
  alzheimer’s disease based on meg activity with a randomized convolutional
  neural network,'' \emph{Artificial Intelligence in Medicine}, vol. 107, p.
  101924, 2020.

\bibitem{aoe2019automatic}
J.~Aoe, R.~Fukuma, T.~Yanagisawa, T.~Harada, M.~Tanaka, M.~Kobayashi, Y.~Inoue,
  S.~Yamamoto, Y.~Ohnishi, and H.~Kishima, ``Automatic diagnosis of
  neurological diseases using meg signals with a deep neural network,''
  \emph{Scientific reports}, vol.~9, no.~1, pp. 1--9, 2019.

\bibitem{ballard1987modular}
D.~H. Ballard, ``Modular learning in neural networks.'' in \emph{AAAI}, 1987,
  pp. 279--284.

\bibitem{charte2018practical}
D.~Charte, F.~Charte, S.~Garc{\'\i}a, M.~J. del Jesus, and F.~Herrera, ``A
  practical tutorial on autoencoders for nonlinear feature fusion: Taxonomy,
  models, software and guidelines,'' \emph{Information Fusion}, vol.~44, pp.
  78--96, 2018.

\bibitem{vincent2008extracting}
P.~Vincent, H.~Larochelle, Y.~Bengio, and P.-A. Manzagol, ``Extracting and
  composing robust features with denoising autoencoders,'' in \emph{Proceedings
  of the 25th international conference on Machine learning}, 2008, pp.
  1096--1103.

\bibitem{cowton2018combined}
J.~Cowton, I.~Kyriazakis, T.~Pl{\"o}tz, and J.~Bacardit, ``A combined deep
  learning gru-autoencoder for the early detection of respiratory disease in
  pigs using multiple environmental sensors,'' \emph{Sensors}, vol.~18, no.~8,
  p. 2521, 2018.

\bibitem{wang2016research}
K.~Wang, Y.~Zhao, Q.~Xiong, M.~Fan, G.~Sun, L.~Ma, and T.~Liu, ``Research on
  healthy anomaly detection model based on deep learning from multiple
  time-series physiological signals,'' \emph{Scientific Programming}, vol.
  2016, 2016.

\bibitem{sato2018primitive}
D.~Sato, S.~Hanaoka, Y.~Nomura, T.~Takenaga, S.~Miki, T.~Yoshikawa, N.~Hayashi,
  and O.~Abe, ``A primitive study on unsupervised anomaly detection with an
  autoencoder in emergency head ct volumes,'' in \emph{Medical Imaging 2018:
  Computer-Aided Diagnosis}, vol. 10575.\hskip 1em plus 0.5em minus 0.4em\relax
  International Society for Optics and Photonics, 2018, p. 105751P.

\bibitem{lu2018anomaly}
Y.~Lu and P.~Xu, ``Anomaly detection for skin disease images using variational
  autoencoder,'' \emph{arXiv preprint arXiv:1807.01349}, 2018.

\bibitem{zimmerer2018context}
D.~Zimmerer, S.~A. Kohl, J.~Petersen, F.~Isensee, and K.~H. Maier-Hein,
  ``Context-encoding variational autoencoder for unsupervised anomaly
  detection,'' \emph{arXiv preprint arXiv:1812.05941}, 2018.

\bibitem{uzunova2019unsupervised}
H.~Uzunova, S.~Schultz, H.~Handels, and J.~Ehrhardt, ``Unsupervised pathology
  detection in medical images using conditional variational autoencoders,''
  \emph{International journal of computer assisted radiology and surgery},
  vol.~14, no.~3, pp. 451--461, 2019.

\bibitem{saxena2020generative}
D.~Saxena and J.~Cao, ``Generative adversarial networks (gans): Challenges,
  solutions, and future directions,'' \emph{arXiv preprint arXiv:2005.00065},
  2020.

\bibitem{goodfellow2014generative}
I.~Goodfellow, J.~Pouget-Abadie, M.~Mirza, B.~Xu, D.~Warde-Farley, S.~Ozair,
  A.~Courville, and Y.~Bengio, ``Generative adversarial nets,'' in
  \emph{Advances in neural information processing systems}, 2014, pp.
  2672--2680.

\bibitem{fernando2018task}
T.~Fernando, S.~Denman, S.~Sridharan, and C.~Fookes, ``Task specific visual
  saliency prediction with memory augmented conditional generative adversarial
  networks,'' in \emph{2018 IEEE Winter Conference on Applications of Computer
  Vision (WACV)}.\hskip 1em plus 0.5em minus 0.4em\relax IEEE, 2018, pp.
  1539--1548.

\bibitem{fernando2018gd}
------, ``Gd-gan: Generative adversarial networks for trajectory prediction and
  group detection in crowds,'' in \emph{Asian Conference on Computer
  Vision}.\hskip 1em plus 0.5em minus 0.4em\relax Springer, 2018, pp. 314--330.

\bibitem{schlegl2019f}
T.~Schlegl, P.~Seeb{\"o}ck, S.~M. Waldstein, G.~Langs, and U.~Schmidt-Erfurth,
  ``f-anogan: Fast unsupervised anomaly detection with generative adversarial
  networks,'' \emph{Medical image analysis}, vol.~54, pp. 30--44, 2019.

\bibitem{zhang2018generalized}
Z.~Zhang and M.~Sabuncu, ``Generalized cross entropy loss for training deep
  neural networks with noisy labels,'' in \emph{Advances in neural information
  processing systems}, 2018, pp. 8778--8788.

\bibitem{schmidt2018artificial}
U.~Schmidt-Erfurth, A.~Sadeghipour, B.~S. Gerendas, S.~M. Waldstein, and
  H.~Bogunovi{\'c}, ``Artificial intelligence in retina,'' \emph{Progress in
  retinal and eye research}, vol.~67, pp. 1--29, 2018.

\bibitem{esteva2017dermatologist}
A.~Esteva, B.~Kuprel, R.~A. Novoa, J.~Ko, S.~M. Swetter, H.~M. Blau, and
  S.~Thrun, ``Dermatologist-level classification of skin cancer with deep
  neural networks,'' \emph{nature}, vol. 542, no. 7639, pp. 115--118, 2017.

\bibitem{turner2017deep}
J.~Turner, A.~Page, T.~Mohsenin, and T.~Oates, ``Deep belief networks used on
  high resolution multichannel electroencephalography data for seizure
  detection,'' \emph{arXiv preprint arXiv:1708.08430}, 2017.

\bibitem{jawanpuria2015efficient}
P.~K. Jawanpuria, M.~Lapin, M.~Hein, and B.~Schiele, ``Efficient output kernel
  learning for multiple tasks,'' in \emph{Advances in neural information
  processing systems}, 2015, pp. 1189--1197.

\bibitem{kisilev2016medical}
P.~Kisilev, E.~Sason, E.~Barkan, and S.~Hashoul, ``Medical image description
  using multi-task-loss cnn,'' in \emph{Deep Learning and Data Labeling for
  Medical Applications}.\hskip 1em plus 0.5em minus 0.4em\relax Springer, 2016,
  pp. 121--129.

\bibitem{williams1995gradient}
R.~Williams, ``Gradient-based learning algorithm for recurrent networks,''
  \emph{Back-propagation: theory, architectures and applications}, 1995.

\bibitem{hochreiter1998vanishing}
S.~Hochreiter, ``The vanishing gradient problem during learning recurrent
  neural nets and problem solutions,'' \emph{International Journal of
  Uncertainty, Fuzziness and Knowledge-Based Systems}, vol.~6, no.~02, pp.
  107--116, 1998.

\bibitem{bengio1994learning}
Y.~Bengio, P.~Simard, and P.~Frasconi, ``Learning long-term dependencies with
  gradient descent is difficult,'' \emph{IEEE transactions on neural networks},
  vol.~5, no.~2, pp. 157--166, 1994.

\bibitem{cho2014learning}
K.~Cho, B.~Van~Merri{\"e}nboer, C.~Gulcehre, D.~Bahdanau, F.~Bougares,
  H.~Schwenk, and Y.~Bengio, ``Learning phrase representations using rnn
  encoder-decoder for statistical machine translation,'' \emph{arXiv preprint
  arXiv:1406.1078}, 2014.

\bibitem{fernando2018tree}
T.~Fernando, S.~Denman, A.~McFadyen, S.~Sridharan, and C.~Fookes, ``Tree memory
  networks for modelling long-term temporal dependencies,''
  \emph{Neurocomputing}, vol. 304, pp. 64--81, 2018.

\bibitem{fernando2018learning}
T.~Fernando, S.~Denman, S.~Sridharan, and C.~Fookes, ``Learning temporal
  strategic relationships using generative adversarial imitation learning,'' in
  \emph{Proceedings of the 17th International Conference on Autonomous Agents
  and MultiAgent Systems}.\hskip 1em plus 0.5em minus 0.4em\relax International
  Foundation for Autonomous Agents and Multiagent Systems, 2018, pp. 113--121.

\bibitem{fernando2019memory}
------, ``Memory augmented deep generative models for forecasting the next shot
  location in tennis,'' \emph{IEEE Transactions on Knowledge and Data
  Engineering}, 2019.

\bibitem{gammulle2019forecasting}
H.~Gammulle, S.~Denman, S.~Sridharan, and C.~Fookes, ``Forecasting future
  action sequences with neural memory networks,'' \emph{British Machine Vision
  Conference (BMVC)}, 2019.

\bibitem{ma2019taxonomy}
Y.~Ma and J.~C. Principe, ``A taxonomy for neural memory networks,'' \emph{IEEE
  transactions on neural networks and learning systems}, 2019.

\bibitem{munkhdalai2017neural}
T.~Munkhdalai and H.~Yu, ``Neural semantic encoders,'' in \emph{Proceedings of
  the conference. Association for Computational Linguistics. Meeting},
  vol.~1.\hskip 1em plus 0.5em minus 0.4em\relax NIH Public Access, 2017, p.
  397.

\bibitem{jagannatha2016bidirectional}
A.~N. Jagannatha and H.~Yu, ``Bidirectional rnn for medical event detection in
  electronic health records,'' in \emph{Proceedings of the conference.
  Association for Computational Linguistics. North American Chapter. Meeting},
  vol. 2016.\hskip 1em plus 0.5em minus 0.4em\relax NIH Public Access, 2016, p.
  473.

\bibitem{yang2018toward}
H.~Yang and H.~Gao, ``Toward sustainable virtualized healthcare: extracting
  medical entities from chinese online health consultations using deep neural
  networks,'' \emph{Sustainability}, vol.~10, no.~9, p. 3292, 2018.

\bibitem{latif2018phonocardiographic}
S.~Latif, M.~Usman, R.~Rana, and J.~Qadir, ``Phonocardiographic sensing using
  deep learning for abnormal heartbeat detection,'' \emph{IEEE Sensors
  Journal}, vol.~18, no.~22, pp. 9393--9400, 2018.

\bibitem{ahmedt2019neural}
D.~Ahmedt-Aristizabal, T.~Fernando, S.~Denman, L.~Petersson, M.~J. Aburn, and
  C.~Fookes, ``Neural memory networks for robust classification of seizure
  type,'' \emph{International Conferences of the IEEE Engineering in Medicine
  and Biology Society}, 2020.

\bibitem{yoo2018deep}
Y.~Yoo, L.~Y. Tang, T.~Brosch, D.~K. Li, S.~Kolind, I.~Vavasour, A.~Rauscher,
  A.~L. MacKay, A.~Traboulsee, and R.~C. Tam, ``Deep learning of joint myelin
  and t1w mri features in normal-appearing brain tissue to distinguish between
  multiple sclerosis patients and healthy controls,'' \emph{NeuroImage:
  Clinical}, vol.~17, pp. 169--178, 2018.

\bibitem{hinton2009deep}
G.~E. Hinton, ``Deep belief networks,'' \emph{Scholarpedia}, vol.~4, no.~5, p.
  5947, 2009.

\bibitem{breiman2001random}
L.~Breiman, ``Random forests,'' \emph{Machine learning}, vol.~45, no.~1, pp.
  5--32, 2001.

\bibitem{lu2018multimodal}
D.~Lu, K.~Popuri, G.~W. Ding, R.~Balachandar, and M.~F. Beg, ``Multimodal and
  multiscale deep neural networks for the early diagnosis of alzheimer’s
  disease using structural mr and fdg-pet images,'' \emph{Scientific reports},
  vol.~8, no.~1, pp. 1--13, 2018.

\bibitem{jack2008alzheimer}
C.~R. Jack~Jr, M.~A. Bernstein, N.~C. Fox, P.~Thompson, G.~Alexander,
  D.~Harvey, B.~Borowski, P.~J. Britson, J.~L.~Whitwell, C.~Ward \emph{et~al.},
  ``The alzheimer's disease neuroimaging initiative (adni): Mri methods,''
  \emph{Journal of Magnetic Resonance Imaging: An Official Journal of the
  International Society for Magnetic Resonance in Medicine}, vol.~27, no.~4,
  pp. 685--691, 2008.

\bibitem{le2017automated}
M.~H. Le, J.~Chen, L.~Wang, Z.~Wang, W.~Liu, K.-T.~T. Cheng, and X.~Yang,
  ``Automated diagnosis of prostate cancer in multi-parametric mri based on
  multimodal convolutional neural networks,'' \emph{Physics in Medicine \&
  Biology}, vol.~62, no.~16, p. 6497, 2017.

\bibitem{islam2018brain}
J.~Islam and Y.~Zhang, ``Brain mri analysis for alzheimer’s disease diagnosis
  using an ensemble system of deep convolutional neural networks,'' \emph{Brain
  informatics}, vol.~5, no.~2, p.~2, 2018.

\bibitem{marcusoasis}
D.~Marcus, T.~Wang \emph{et~al.}, ``Oasis: Cross-sectional mri data in young,
  middle aged, nondemented, and demented older adults,'' \emph{Journal of
  Cognitive Neuroscience}.

\bibitem{shehata2018computer}
M.~Shehata, F.~Khalifa, A.~Soliman, M.~Ghazal, F.~Taher, M.~Abou El-Ghar, A.~C.
  Dwyer, G.~Gimel’farb, R.~S. Keynton, and A.~El-Baz, ``Computer-aided
  diagnostic system for early detection of acute renal transplant rejection
  using diffusion-weighted mri,'' \emph{IEEE Transactions on Biomedical
  Engineering}, vol.~66, no.~2, pp. 539--552, 2018.

\bibitem{zeng2018multi}
L.-L. Zeng, H.~Wang, P.~Hu, B.~Yang, W.~Pu, H.~Shen, X.~Chen, Z.~Liu, H.~Yin,
  Q.~Tan \emph{et~al.}, ``Multi-site diagnostic classification of schizophrenia
  using discriminant deep learning with functional connectivity mri,''
  \emph{EBioMedicine}, vol.~30, pp. 74--85, 2018.

\bibitem{han2018automated}
Z.~Han, B.~Wei, S.~Leung, I.~B. Nachum, D.~Laidley, and S.~Li, ``Automated
  pathogenesis-based diagnosis of lumbar neural foraminal stenosis via deep
  multiscale multitask learning,'' \emph{Neuroinformatics}, vol.~16, no. 3-4,
  pp. 325--337, 2018.

\bibitem{cheng2015enhanced}
J.~Cheng, W.~Huang, S.~Cao, R.~Yang, W.~Yang, Z.~Yun, Z.~Wang, and Q.~Feng,
  ``Enhanced performance of brain tumor classification via tumor region
  augmentation and partition,'' \emph{PloS one}, vol.~10, no.~10, p. e0140381,
  2015.

\bibitem{deng2009imagenet}
J.~Deng, W.~Dong, R.~Socher, L.-J. Li, K.~Li, and L.~Fei-Fei, ``Imagenet: A
  very deep convolutional networks for large-scale image recognition
  hierarchical image database,'' in \emph{2009 IEEE conference on computer
  vision and pattern recognition}.\hskip 1em plus 0.5em minus 0.4em\relax Ieee,
  2009, pp. 248--255.

\bibitem{klang2020deep}
E.~Klang, Y.~Barash, R.~Y. Margalit, S.~Soffer, O.~Shimon, A.~Albshesh,
  S.~Ben-Horin, M.~M. Amitai, R.~Eliakim, and U.~Kopylov, ``Deep learning
  algorithms for automated detection of crohn’s disease ulcers by video
  capsule endoscopy,'' \emph{Gastrointestinal Endoscopy}, vol.~91, no.~3, pp.
  606--613, 2020.

\bibitem{chollet2017xception}
F.~Chollet, ``Xception: Deep learning with depthwise separable convolutions,''
  in \emph{Proceedings of the IEEE conference on computer vision and pattern
  recognition}, 2017, pp. 1251--1258.

\bibitem{alaskar2019application}
H.~Alaskar, A.~Hussain, N.~Al-Aseem, P.~Liatsis, and D.~Al-Jumeily,
  ``Application of convolutional neural networks for automated ulcer detection
  in wireless capsule endoscopy images,'' \emph{Sensors}, vol.~19, no.~6, p.
  1265, 2019.

\bibitem{szegedy2015going}
C.~Szegedy, W.~Liu, Y.~Jia, P.~Sermanet, S.~Reed, D.~Anguelov, D.~Erhan,
  V.~Vanhoucke, and A.~Rabinovich, ``Going deeper with convolutions,'' in
  \emph{Proceedings of the IEEE conference on computer vision and pattern
  recognition}, 2015, pp. 1--9.

\bibitem{krizhevsky2012imagenet}
A.~Krizhevsky, I.~Sutskever, and G.~E. Hinton, ``Imagenet classification with
  deep convolutional neural networks,'' in \emph{Advances in neural information
  processing systems}, 2012, pp. 1097--1105.

\bibitem{fan2018computer}
S.~Fan, L.~Xu, Y.~Fan, K.~Wei, and L.~Li, ``Computer-aided detection of small
  intestinal ulcer and erosion in wireless capsule endoscopy images,''
  \emph{Physics in Medicine \& Biology}, vol.~63, no.~16, p. 165001, 2018.

\bibitem{wang2019systematic}
S.~Wang, Y.~Xing, L.~Zhang, H.~Gao, and H.~Zhang, ``A systematic evaluation and
  optimization of automatic detection of ulcers in wireless capsule endoscopy
  on a large dataset using deep convolutional neural networks,'' \emph{Physics
  in Medicine \& Biology}, vol.~64, no.~23, p. 235014, 2019.

\bibitem{lin2017focal}
T.-Y. Lin, P.~Goyal, R.~Girshick, K.~He, and P.~Doll{\'a}r, ``Focal loss for
  dense object detection,'' in \emph{Proceedings of the IEEE international
  conference on computer vision}, 2017, pp. 2980--2988.

\bibitem{he2016deep}
K.~He, X.~Zhang, S.~Ren, and J.~Sun, ``Deep residual learning for image
  recognition,'' in \emph{Proceedings of the IEEE conference on computer vision
  and pattern recognition}, 2016, pp. 770--778.

\bibitem{gammulle2020two}
H.~Gammulle, S.~Denman, S.~Sridharan, and C.~Fookes, ``Two-stream deep feature
  modelling for automated video endoscopy data analysis,'' \emph{International
  Conference on Medical Image Computing and Computer Assisted Intervention},
  2020.

\bibitem{santoro2017simple}
A.~Santoro, D.~Raposo, D.~G. Barrett, M.~Malinowski, R.~Pascanu, P.~Battaglia,
  and T.~Lillicrap, ``A simple neural network module for relational
  reasoning,'' in \emph{Advances in neural information processing systems},
  2017, pp. 4967--4976.

\bibitem{pogorelov2017nerthus}
K.~Pogorelov, K.~R. Randel, T.~de~Lange, S.~L. Eskeland, C.~Griwodz,
  D.~Johansen, C.~Spampinato, M.~Taschwer, M.~Lux, P.~T. Schmidt \emph{et~al.},
  ``Nerthus: A bowel preparation quality video dataset,'' in \emph{Proceedings
  of the 8th ACM on Multimedia Systems Conference}, 2017, pp. 170--174.

\bibitem{simonyan2014very}
K.~Simonyan and A.~Zisserman, ``Very deep convolutional networks for
  large-scale image recognition,'' \emph{arXiv preprint arXiv:1409.1556}, 2014.

\bibitem{wu2019applying}
J.~M.-T. Wu, M.-H. Tsai, Y.~Z. Huang, S.~H. Islam, M.~M. Hassan, A.~Alelaiwi,
  and G.~Fortino, ``Applying an ensemble convolutional neural network with
  savitzky--golay filter to construct a phonocardiogram prediction model,''
  \emph{Applied Soft Computing}, vol.~78, pp. 29--40, 2019.

\bibitem{savitzky1964smoothing}
A.~Savitzky and M.~J. Golay, ``Smoothing and differentiation of data by
  simplified least squares procedures.'' \emph{Analytical chemistry}, vol.~36,
  no.~8, pp. 1627--1639, 1964.

\bibitem{li2020classification}
F.~Li, H.~Tang, S.~Shang, K.~Mathiak, and F.~Cong, ``Classification of heart
  sounds using convolutional neural network,'' \emph{Applied Sciences},
  vol.~10, no.~11, p. 3956, 2020.

\bibitem{yang2016classification}
T.-c.~I. Yang and H.~Hsieh, ``Classification of acoustic physiological signals
  based on deep learning neural networks with augmented features,'' in
  \emph{2016 Computing in Cardiology Conference (CinC)}.\hskip 1em plus 0.5em
  minus 0.4em\relax IEEE, 2016, pp. 569--572.

\bibitem{gao2020gated}
S.~Gao, Y.~Zheng, and X.~Guo, ``Gated recurrent unit-based heart sound analysis
  for heart failure screening,'' \emph{BioMedical Engineering OnLine}, vol.~19,
  no.~1, p.~3, 2020.

\bibitem{springer2015logistic}
D.~B. Springer, L.~Tarassenko, and G.~D. Clifford, ``Logistic
  regression-hsmm-based heart sound segmentation,'' \emph{IEEE Transactions on
  Biomedical Engineering}, vol.~63, no.~4, pp. 822--832, 2015.

\bibitem{dissanayake2020understanding}
T.~Dissanayake, T.~Fernando, S.~Denman, S.~Sridharan, H.~Ghaemmaghami, and
  C.~Fookes, ``A robust interpretable deep learning classifier for heart
  anomaly detection without segmentation,'' \emph{IEEE Journal of Biomedical
  and Health Informatics}, 2020.

\bibitem{fernando2019heart}
T.~Fernando, H.~Ghaemmaghami, S.~Denman, S.~Sridharan, N.~Hussain, and
  C.~Fookes, ``Heart sound segmentation using bidirectional lstms with
  attention,'' \emph{IEEE Journal of Biomedical and Health Informatics},
  vol.~24, no.~6, pp. 1601--1609, 2019.

\bibitem{molnar2020interpretable}
C.~Molnar, \emph{Interpretable Machine Learning}.\hskip 1em plus 0.5em minus
  0.4em\relax Lulu. com, 2020.

\bibitem{oh2020classification}
S.~L. Oh, V.~Jahmunah, C.~P. Ooi, R.-S. Tan, E.~J. Ciaccio, T.~Yamakawa,
  M.~Tanabe, M.~Kobayashi, and U.~R. Acharya, ``Classification of heart sound
  signals using a novel deep wavenet model,'' \emph{Computer Methods and
  Programs in Biomedicine}, p. 105604, 2020.

\bibitem{gadhoumi2012discriminating}
K.~Gadhoumi, J.-M. Lina, and J.~Gotman, ``Discriminating preictal and
  interictal states in patients with temporal lobe epilepsy using wavelet
  analysis of intracerebral eeg,'' \emph{Clinical neurophysiology}, vol. 123,
  no.~10, pp. 1906--1916, 2012.

\bibitem{tsiouris2018long}
K.~M. Tsiouris, V.~C. Pezoulas, M.~Zervakis, S.~Konitsiotis, D.~D. Koutsouris,
  and D.~I. Fotiadis, ``A long short-term memory deep learning network for the
  prediction of epileptic seizures using eeg signals,'' \emph{Computers in
  biology and medicine}, vol.~99, pp. 24--37, 2018.

\bibitem{thara2019epileptic}
D.~Thara, B.~PremaSudha, and F.~Xiong, ``Epileptic seizure detection and
  prediction using stacked bidirectional long short term memory,''
  \emph{Pattern Recognition Letters}, vol. 128, pp. 529--535, 2019.

\bibitem{Khan2018FocalNetworks}
H.~Khan, L.~Marcuse, M.~Fields, K.~Swann, and B.~Yener, ``{Focal onset seizure
  prediction using convolutional networks},'' \emph{IEEE Transactions on
  Biomedical Engineering}, vol.~65, no.~9, pp. 2109--2118, 9 2018.

\bibitem{truong2017generalised}
N.~D. Truong, A.~D. Nguyen, L.~Kuhlmann, M.~R. Bonyadi, J.~Yang, and
  O.~Kavehei, ``A generalised seizure prediction with convolutional neural
  networks for intracranial and scalp electroencephalogram data analysis,''
  \emph{arXiv preprint arXiv:1707.01976}, 2017.

\bibitem{winterhalder2003seizure}
M.~Winterhalder, T.~Maiwald, H.~Voss, R.~Aschenbrenner-Scheibe, J.~Timmer, and
  A.~Schulze-Bonhage, ``The seizure prediction characteristic: a general
  framework to assess and compare seizure prediction methods,'' \emph{Epilepsy
  \& Behavior}, vol.~4, no.~3, pp. 318--325, 2003.

\bibitem{shoeb2009application}
A.~H. Shoeb, ``Application of machine learning to epileptic seizure onset
  detection and treatment,'' Ph.D. dissertation, Massachusetts Institute of
  Technology, 2009.

\bibitem{hussein2019human}
R.~Hussein, M.~O. Ahmed, R.~Ward, Z.~J. Wang, L.~Kuhlmann, and Y.~Guo, ``Human
  intracranial eeg quantitative analysis and automatic feature learning for
  epileptic seizure prediction,'' \emph{arXiv preprint arXiv:1904.03603}, 2019.

\bibitem{kiral2018epileptic}
I.~Kiral-Kornek, S.~Roy, E.~Nurse, B.~Mashford, P.~Karoly, T.~Carroll,
  D.~Payne, S.~Saha, S.~Baldassano, T.~O'Brien \emph{et~al.}, ``Epileptic
  seizure prediction using big data and deep learning: toward a mobile
  system,'' \emph{EBioMedicine}, vol.~27, pp. 103--111, 2018.

\bibitem{kuhlmann2018epilepsyecosystem}
L.~Kuhlmann, P.~Karoly, D.~R. Freestone, B.~H. Brinkmann, A.~Temko,
  A.~Barachant, F.~Li, G.~Titericz~Jr, B.~W. Lang, D.~Lavery \emph{et~al.},
  ``Epilepsyecosystem. org: crowd-sourcing reproducible seizure prediction with
  long-term human intracranial eeg,'' \emph{Brain}, vol. 141, no.~9, pp.
  2619--2630, 2018.

\bibitem{cook2013prediction}
M.~J. Cook, T.~J. O'Brien, S.~F. Berkovic, M.~Murphy, A.~Morokoff, G.~Fabinyi,
  W.~D'Souza, R.~Yerra, J.~Archer, L.~Litewka \emph{et~al.}, ``Prediction of
  seizure likelihood with a long-term, implanted seizure advisory system in
  patients with drug-resistant epilepsy: a first-in-man study,'' \emph{The
  Lancet Neurology}, vol.~12, no.~6, pp. 563--571, 2013.

\bibitem{andrzejak2001indications}
R.~G. Andrzejak, K.~Lehnertz, F.~Mormann, C.~Rieke, P.~David, and C.~E. Elger,
  ``Indications of nonlinear deterministic and finite-dimensional structures in
  time series of brain electrical activity: Dependence on recording region and
  brain state,'' \emph{Physical Review E}, vol.~64, no.~6, p. 061907, 2001.

\bibitem{truong2019epileptic}
N.~D. Truong, L.~Kuhlmann, M.~R. Bonyadi, D.~Querlioz, L.~Zhou, and O.~Kavehei,
  ``Epileptic seizure forecasting with generative adversarial networks,''
  \emph{IEEE Access}, vol.~7, pp. 143\,999--144\,009, 2019.

\bibitem{ihle2012epilepsiae}
M.~Ihle, H.~Feldwisch-Drentrup, C.~A. Teixeira, A.~Witon, B.~Schelter,
  J.~Timmer, and A.~Schulze-Bonhage, ``Epilepsiae--a european epilepsy
  database,'' \emph{Computer methods and programs in biomedicine}, vol. 106,
  no.~3, pp. 127--138, 2012.

\bibitem{gammulle2020fine}
H.~Gammulle, S.~Denman, S.~Sridharan, and C.~Fookes, ``Fine-grained action
  segmentation using the semi-supervised action gan,'' \emph{Pattern
  Recognition}, vol.~98, p. 107039, 2020.

\bibitem{singh2020explainable}
A.~Singh, S.~Sengupta, and V.~Lakshminarayanan, ``Explainable deep learning
  models in medical image analysis,'' \emph{arXiv preprint arXiv:2005.13799},
  2020.

\bibitem{zhou2016learning}
B.~Zhou, A.~Khosla, A.~Lapedriza, A.~Oliva, and A.~Torralba, ``Learning deep
  features for discriminative localization,'' in \emph{Proceedings of the IEEE
  conference on computer vision and pattern recognition}, 2016, pp. 2921--2929.

\bibitem{selvaraju2017grad}
R.~R. Selvaraju, M.~Cogswell, A.~Das, R.~Vedantam, D.~Parikh, and D.~Batra,
  ``Grad-cam: Visual explanations from deep networks via gradient-based
  localization,'' in \emph{Proceedings of the IEEE international conference on
  computer vision}, 2017, pp. 618--626.

\bibitem{chattopadhay2018grad}
A.~Chattopadhay, A.~Sarkar, P.~Howlader, and V.~N. Balasubramanian,
  ``Grad-cam++: Generalized gradient-based visual explanations for deep
  convolutional networks,'' in \emph{2018 IEEE Winter Conference on
  Applications of Computer Vision (WACV)}.\hskip 1em plus 0.5em minus
  0.4em\relax IEEE, 2018, pp. 839--847.

\bibitem{ribeiro2016should}
M.~T. Ribeiro, S.~Singh, and C.~Guestrin, ``" why should i trust you?"
  explaining the predictions of any classifier,'' in \emph{Proceedings of the
  22nd ACM SIGKDD international conference on knowledge discovery and data
  mining}, 2016, pp. 1135--1144.

\bibitem{lundberg2017unified}
S.~M. Lundberg and S.-I. Lee, ``A unified approach to interpreting model
  predictions,'' in \emph{Advances in neural information processing systems},
  2017, pp. 4765--4774.

\bibitem{lundberg2018explainable}
S.~M. Lundberg, B.~Nair, M.~S. Vavilala, M.~Horibe, M.~J. Eisses, T.~Adams,
  D.~E. Liston, D.~K.-W. Low, S.-F. Newman, J.~Kim \emph{et~al.}, ``Explainable
  machine-learning predictions for the prevention of hypoxaemia during
  surgery,'' \emph{Nature biomedical engineering}, vol.~2, no.~10, pp.
  749--760, 2018.

\bibitem{shrikumar2017learning}
A.~Shrikumar, P.~Greenside, and A.~Kundaje, ``Learning important features
  through propagating activation differences,'' \emph{arXiv preprint
  arXiv:1704.02685}, 2017.

\bibitem{montavon2017explaining}
G.~Montavon, S.~Lapuschkin, A.~Binder, W.~Samek, and K.-R. M{\"u}ller,
  ``Explaining nonlinear classification decisions with deep taylor
  decomposition,'' \emph{Pattern Recognition}, vol.~65, pp. 211--222, 2017.

\bibitem{springenberg2014striving}
J.~T. Springenberg, A.~Dosovitskiy, T.~Brox, and M.~Riedmiller, ``Striving for
  simplicity: The all convolutional net,'' \emph{arXiv preprint
  arXiv:1412.6806}, 2014.

\bibitem{sundararajan2017axiomatic}
M.~Sundararajan, A.~Taly, and Q.~Yan, ``Axiomatic attribution for deep
  networks,'' \emph{arXiv preprint arXiv:1703.01365}, 2017.

\bibitem{pereira2018automatic}
S.~Pereira, R.~Meier, V.~Alves, M.~Reyes, and C.~A. Silva, ``Automatic brain
  tumor grading from mri data using convolutional neural networks and quality
  assessment,'' in \emph{Understanding and interpreting machine learning in
  medical image computing applications}.\hskip 1em plus 0.5em minus 0.4em\relax
  Springer, 2018, pp. 106--114.

\bibitem{young2019deep}
K.~Young, G.~Booth, B.~Simpson, R.~Dutton, and S.~Shrapnel, ``Deep neural
  network or dermatologist?'' in \emph{Interpretability of Machine Intelligence
  in Medical Image Computing and Multimodal Learning for Clinical Decision
  Support}.\hskip 1em plus 0.5em minus 0.4em\relax Springer, 2019, pp. 48--55.

\bibitem{tsiknakis2020interpretable}
N.~Tsiknakis, E.~Trivizakis, E.~E. Vassalou, G.~Z. Papadakis, D.~A. Spandidos,
  A.~Tsatsakis, J.~S{\'a}nchez-Garc{\'\i}a, R.~L{\'o}pez-Gonz{\'a}lez,
  N.~Papanikolaou, A.~H. Karantanas \emph{et~al.}, ``Interpretable artificial
  intelligence framework for covid-19 screening on chest x-rays,''
  \emph{Experimental and Therapeutic Medicine}, vol.~20, no.~2, pp. 727--735,
  2020.

\bibitem{chatterjee2020exploration}
S.~Chatterjee, F.~Saad, C.~Sarasaen, S.~Ghosh, R.~Khatun, P.~Radeva, G.~Rose,
  S.~Stober, O.~Speck, and A.~N{\"u}rnberger, ``Exploration of interpretability
  techniques for deep covid-19 classification using chest x-ray images,''
  \emph{arXiv preprint arXiv:2006.02570}, 2020.

\bibitem{zhang2017mdnet}
Z.~Zhang, Y.~Xie, F.~Xing, M.~McGough, and L.~Yang, ``Mdnet: A semantically and
  visually interpretable medical image diagnosis network,'' in
  \emph{Proceedings of the IEEE conference on computer vision and pattern
  recognition}, 2017, pp. 6428--6436.

\bibitem{lee2019generation}
H.~Lee, S.~T. Kim, and Y.~M. Ro, ``Generation of multimodal justification using
  visual word constraint model for explainable computer-aided diagnosis,'' in
  \emph{Interpretability of Machine Intelligence in Medical Image Computing and
  Multimodal Learning for Clinical Decision Support}.\hskip 1em plus 0.5em
  minus 0.4em\relax Springer, 2019, pp. 21--29.

\bibitem{zhu2019guideline}
P.~Zhu and M.~Ogino, ``Guideline-based additive explanation for computer-aided
  diagnosis of lung nodules,'' in \emph{Interpretability of Machine
  Intelligence in Medical Image Computing and Multimodal Learning for Clinical
  Decision Support}.\hskip 1em plus 0.5em minus 0.4em\relax Springer, 2019, pp.
  39--47.

\bibitem{arbabshirani2018advanced}
M.~R. Arbabshirani, B.~K. Fornwalt, G.~J. Mongelluzzo, J.~D. Suever, B.~D.
  Geise, A.~A. Patel, and G.~J. Moore, ``Advanced machine learning in action:
  identification of intracranial hemorrhage on computed tomography scans of the
  head with clinical workflow integration,'' \emph{NPJ digital medicine},
  vol.~1, no.~1, pp. 1--7, 2018.

\bibitem{almazroa2017agreement}
A.~Almazroa, S.~Alodhayb, E.~Osman, E.~Ramadan, M.~Hummadi, M.~Dlaim,
  M.~Alkatee, K.~Raahemifar, and V.~Lakshminarayanan, ``Agreement among
  ophthalmologists in marking the optic disc and optic cup in fundus images,''
  \emph{International ophthalmology}, vol.~37, no.~3, pp. 701--717, 2017.

\bibitem{huang2018towards}
C.~Huang, Y.~Wu, Y.~Zuo, K.~Pei, and G.~Min, ``Towards experienced anomaly
  detector through reinforcement learning,'' in \emph{AAAI}, 2018.

\bibitem{oh2019sequential}
M.-h. Oh and G.~Iyengar, ``Sequential anomaly detection using inverse
  reinforcement learning,'' in \emph{Proceedings of the 25th ACM SIGKDD
  International Conference on Knowledge Discovery \& Data Mining}, 2019, pp.
  1480--1490.

\bibitem{richens2020improving}
J.~G. Richens, C.~M. Lee, and S.~Johri, ``Improving the accuracy of medical
  diagnosis with causal machine learning,'' \emph{Nature communications},
  vol.~11, no.~1, pp. 1--9, 2020.

\bibitem{castro2020causality}
D.~C. Castro, I.~Walker, and B.~Glocker, ``Causality matters in medical
  imaging,'' \emph{Nature Communications}, vol.~11, no.~1, pp. 1--10, 2020.

\bibitem{altman2015association}
N.~Altman and M.~Krzywinski, ``Association, correlation and causation,''
  \emph{Nature Methods}, 2015.

\bibitem{nauta2019causal}
M.~Nauta, D.~Bucur, and C.~Seifert, ``Causal discovery with attention-based
  convolutional neural networks,'' \emph{Machine Learning and Knowledge
  Extraction}, vol.~1, no.~1, pp. 312--340, 2019.

\bibitem{zenil2019algorithmic}
H.~Zenil, N.~A. Kiani, F.~Marabita, Y.~Deng, S.~Elias, A.~Schmidt, G.~Ball, and
  J.~Tegn{\'e}r, ``An algorithmic information calculus for causal discovery and
  reprogramming systems,'' \emph{iScience}, vol.~19, pp. 1160--1172, 2019.

\bibitem{zenil2019causal}
H.~Zenil, N.~A. Kiani, A.~A. Zea, and J.~Tegn{\'e}r, ``Causal deconvolution by
  algorithmic generative models,'' \emph{Nature Machine Intelligence}, vol.~1,
  no.~1, pp. 58--66, 2019.

\bibitem{zhang2019bayesian}
Y.~Zhang, S.~Pal, M.~Coates, and D.~Ustebay, ``Bayesian graph convolutional
  neural networks for semi-supervised classification,'' in \emph{Proceedings of
  the AAAI Conference on Artificial Intelligence}, vol.~33, 2019, pp.
  5829--5836.

\bibitem{pal2019bayesian}
S.~Pal, F.~Regol, and M.~Coates, ``Bayesian graph convolutional neural networks
  using node copying,'' \emph{arXiv preprint arXiv:1911.04965}, 2019.

\bibitem{kendall2017uncertainties}
A.~Kendall and Y.~Gal, ``What uncertainties do we need in bayesian deep
  learning for computer vision?'' in \emph{Advances in neural information
  processing systems}, 2017, pp. 5574--5584.

\bibitem{leibig2017leveraging}
C.~Leibig, V.~Allken, M.~S. Ayhan, P.~Berens, and S.~Wahl, ``Leveraging
  uncertainty information from deep neural networks for disease detection,''
  \emph{Scientific reports}, vol.~7, no.~1, pp. 1--14, 2017.

\bibitem{kwon2020uncertainty}
Y.~Kwon, J.-H. Won, B.~J. Kim, and M.~C. Paik, ``Uncertainty quantification
  using bayesian neural networks in classification: Application to biomedical
  image segmentation,'' \emph{Computational Statistics \& Data Analysis}, vol.
  142, p. 106816, 2020.

\bibitem{devries2018leveraging}
T.~DeVries and G.~W. Taylor, ``Leveraging uncertainty estimates for predicting
  segmentation quality,'' \emph{arXiv preprint arXiv:1807.00502}, 2018.

\bibitem{seebock2019exploiting}
P.~Seeb{\"o}ck, J.~I. Orlando, T.~Schlegl, S.~M. Waldstein, H.~Bogunovi{\'c},
  S.~Klimscha, G.~Langs, and U.~Schmidt-Erfurth, ``Exploiting epistemic
  uncertainty of anatomy segmentation for anomaly detection in retinal oct,''
  \emph{IEEE transactions on medical imaging}, vol.~39, no.~1, pp. 87--98,
  2019.

\bibitem{nair2020exploring}
T.~Nair, D.~Precup, D.~L. Arnold, and T.~Arbel, ``Exploring uncertainty
  measures in deep networks for multiple sclerosis lesion detection and
  segmentation,'' \emph{Medical image analysis}, vol.~59, p. 101557, 2020.

\bibitem{dissanayake2020domain}
T.~Dissanayake, T.~Fernando, S.~Denman, H.~Ghaemmaghami, S.~Sridharan, and
  C.~Fookes, ``Domain generalization in biosignal classification,'' \emph{arXiv
  preprint arXiv:2011.06207}, 2020.

\bibitem{hu2018meta}
S.~Hu, J.~Tomczak, and M.~Welling, ``Meta-learning for medical image
  classification,'' \emph{Conference on Medical Imaging with Deep Learning
  (MIDL 2018)}, 2018.

\bibitem{mahajan2020meta}
K.~Mahajan, M.~Sharma, and L.~Vig, ``Meta-dermdiagnosis: Few-shot skin disease
  identification using meta-learning,'' in \emph{Proceedings of the IEEE/CVF
  Conference on Computer Vision and Pattern Recognition Workshops}, 2020, pp.
  730--731.

\bibitem{zhang2019metapred}
X.~S. Zhang, F.~Tang, H.~H. Dodge, J.~Zhou, and F.~Wang, ``Metapred:
  Meta-learning for clinical risk prediction with limited patient electronic
  health records,'' in \emph{Proceedings of the 25th ACM SIGKDD International
  Conference on Knowledge Discovery \& Data Mining}, 2019, pp. 2487--2495.

\bibitem{johnson2019survey}
J.~M. Johnson and T.~M. Khoshgoftaar, ``Survey on deep learning with class
  imbalance,'' \emph{Journal of Big Data}, vol.~6, no.~1, p.~27, 2019.

\bibitem{zhang2018imbalanced}
L.~Zhang, H.~Yang, and Z.~Jiang, ``Imbalanced biomedical data classification
  using self-adaptive multilayer elm combined with dynamic gan,''
  \emph{Biomedical engineering online}, vol.~17, no.~1, p. 181, 2018.

\bibitem{mullick2019generative}
S.~S. Mullick, S.~Datta, and S.~Das, ``Generative adversarial minority
  oversampling,'' in \emph{Proceedings of the IEEE International Conference on
  Computer Vision}, 2019, pp. 1695--1704.

\bibitem{zhou2018gan}
T.~Zhou, W.~Liu, C.~Zhou, and L.~Chen, ``Gan-based semi-supervised for
  imbalanced data classification,'' in \emph{2018 4th International Conference
  on Information Management (ICIM)}.\hskip 1em plus 0.5em minus 0.4em\relax
  IEEE, 2018, pp. 17--21.

\bibitem{doersch2015unsupervised}
C.~Doersch, A.~Gupta, and A.~A. Efros, ``Unsupervised visual representation
  learning by context prediction,'' in \emph{Proceedings of the IEEE
  international conference on computer vision}, 2015, pp. 1422--1430.

\bibitem{zhang2016colorful}
R.~Zhang, P.~Isola, and A.~A. Efros, ``Colorful image colorization,'' in
  \emph{European conference on computer vision}.\hskip 1em plus 0.5em minus
  0.4em\relax Springer, 2016, pp. 649--666.

\bibitem{noroozi2016unsupervised}
M.~Noroozi and P.~Favaro, ``Unsupervised learning of visual representations by
  solving jigsaw puzzles,'' in \emph{European Conference on Computer
  Vision}.\hskip 1em plus 0.5em minus 0.4em\relax Springer, 2016, pp. 69--84.

\bibitem{ali2020self}
R.~Ali, M.~U.~K. Khan, and C.~M. Kyung, ``Self-supervised representation
  learning for visual anomaly detection,'' \emph{arXiv preprint
  arXiv:2006.09654}, 2020.

\bibitem{bhandari2019chest}
E.~D. Bhandari and M.~S. Badmera, ``Chest abnormality detection from x-ray
  using deep learning,'' \emph{Chest}, vol.~6, no.~11, 2019.

\bibitem{tataru2017deep}
C.~Tataru, D.~Yi, A.~Shenoyas, and A.~Ma, ``Deep learning for abnormality
  detection in chest x-ray images,'' in \emph{IEEE Conference on Deep
  Learning}, 2017.

\bibitem{tang2020automated}
Y.-X. Tang, Y.-B. Tang, Y.~Peng, K.~Yan, M.~Bagheri, B.~A. Redd, C.~J. Brandon,
  Z.~Lu, M.~Han, J.~Xiao \emph{et~al.}, ``Automated abnormality classification
  of chest radiographs using deep convolutional neural networks,'' \emph{NPJ
  Digital Medicine}, vol.~3, no.~1, pp. 1--8, 2020.

\bibitem{wang2018chestnet}
H.~Wang and Y.~Xia, ``Chestnet: A deep neural network for classification of
  thoracic diseases on chest radiography,'' \emph{arXiv preprint
  arXiv:1807.03058}, 2018.

\bibitem{pham2020interpreting}
H.~H. Pham, T.~T. Le, D.~T. Ngo, D.~Q. Tran, and H.~Q. Nguyen, ``Interpreting
  chest x-rays via cnns that exploit hierarchical disease dependencies and
  uncertainty labels,'' \emph{arXiv preprint arXiv:2005.12734}, 2020.

\bibitem{zhang2020viral}
J.~Zhang, Y.~Xie, Z.~Liao, G.~Pang, J.~Verjans, W.~Li, Z.~Sun, J.~He, Y.~Li,
  C.~Shen \emph{et~al.}, ``Viral pneumonia screening on chest x-ray images
  using confidence-aware anomaly detection,'' \emph{arXiv: 2003.12338}, 2020.

\bibitem{basu2020deep}
S.~Basu and S.~Mitra, ``Deep learning for screening covid-19 using chest x-ray
  images,'' \emph{arXiv preprint arXiv:2004.10507}, 2020.

\bibitem{shah2020diagnosis}
V.~Shah, R.~Keniya, A.~Shridharani, M.~Punjabi, J.~Shah, and N.~Mehendale,
  ``Diagnosis of covid-19 using ct scan images and deep learning techniques,''
  \emph{medRxiv}, 2020.

\bibitem{mishra2020identifying}
A.~K. Mishra, S.~K. Das, P.~Roy, and S.~Bandyopadhyay, ``Identifying covid19
  from chest ct images: A deep convolutional neural networks based approach,''
  \emph{Journal of Healthcare Engineering}, vol. 2020, 2020.

\bibitem{perumal2020detection}
V.~Perumal, V.~Narayanan, and S.~J.~S. Rajasekar, ``Detection of covid-19 using
  cxr and ct images using transfer learning and haralick features,''
  \emph{Applied Intelligence}, pp. 1--18, 2020.

\bibitem{sharma2020drawing}
S.~Sharma, ``Drawing insights from covid-19-infected patients using ct scan
  images and machine learning techniques: a study on 200 patients,''
  \emph{Environmental Science and Pollution Research}, pp. 1--9, 2020.

\bibitem{harmon2020artificial}
S.~A. Harmon, T.~H. Sanford, S.~Xu, E.~B. Turkbey, H.~Roth, Z.~Xu, D.~Yang,
  A.~Myronenko, V.~Anderson, A.~Amalou \emph{et~al.}, ``Artificial intelligence
  for the detection of covid-19 pneumonia on chest ct using multinational
  datasets,'' \emph{Nature communications}, vol.~11, no.~1, pp. 1--7, 2020.

\bibitem{makde2017deep}
V.~Makde, J.~Bhavsar, S.~Jain, and P.~Sharma, ``Deep neural network based
  classification of tumourous and non-tumorous medical images,'' in
  \emph{International Conference on Information and Communication Technology
  for Intelligent Systems}.\hskip 1em plus 0.5em minus 0.4em\relax Springer,
  2017, pp. 199--206.

\bibitem{huang2017lung}
X.~Huang, J.~Shan, and V.~Vaidya, ``Lung nodule detection in ct using 3d
  convolutional neural networks,'' in \emph{2017 IEEE 14th International
  Symposium on Biomedical Imaging (ISBI 2017)}.\hskip 1em plus 0.5em minus
  0.4em\relax IEEE, 2017, pp. 379--383.

\bibitem{jakimovski2019using}
G.~Jakimovski and D.~Davcev, ``Using double convolution neural network for lung
  cancer stage detection,'' \emph{Applied Sciences}, vol.~9, no.~3, p. 427,
  2019.

\bibitem{afshar2018brain}
P.~Afshar, A.~Mohammadi, and K.~N. Plataniotis, ``Brain tumor type
  classification via capsule networks,'' in \emph{2018 25th IEEE International
  Conference on Image Processing (ICIP)}.\hskip 1em plus 0.5em minus
  0.4em\relax IEEE, 2018, pp. 3129--3133.

\bibitem{abiwinanda2019brain}
N.~Abiwinanda, M.~Hanif, S.~T. Hesaputra, A.~Handayani, and T.~R. Mengko,
  ``Brain tumor classification using convolutional neural network,'' in
  \emph{World Congress on Medical Physics and Biomedical Engineering
  2018}.\hskip 1em plus 0.5em minus 0.4em\relax Springer, 2019, pp. 183--189.

\bibitem{ari2018deep}
A.~Ari and D.~Hanbay, ``Deep learning based brain tumor classification and
  detection system,'' \emph{Turkish Journal of Electrical Engineering \&
  Computer Sciences}, vol.~26, no.~5, pp. 2275--2286, 2018.

\bibitem{zhou2018holistic}
Y.~Zhou, Z.~Li, H.~Zhu, C.~Chen, M.~Gao, K.~Xu, and J.~Xu, ``Holistic brain
  tumor screening and classification based on densenet and recurrent neural
  network,'' in \emph{International MICCAI Brainlesion Workshop}.\hskip 1em
  plus 0.5em minus 0.4em\relax Springer, 2018, pp. 208--217.

\bibitem{xu2015deep}
Y.~Xu, Z.~Jia, Y.~Ai, F.~Zhang, M.~Lai, I.~Eric, and C.~Chang, ``Deep
  convolutional activation features for large scale brain tumor histopathology
  image classification and segmentation,'' in \emph{2015 IEEE international
  conference on acoustics, speech and signal processing (ICASSP)}.\hskip 1em
  plus 0.5em minus 0.4em\relax IEEE, 2015, pp. 947--951.

\bibitem{deng2017automated}
C.-L. DENG, H.-Y. JIANG, and H.-M. LI, ``Automated high uptake regions
  recognition and lymphoma detection based on fully convolutional networks on
  chest and abdomen pet image,'' \emph{DEStech Transactions on Biology and
  Health}, no. icmsb, 2017.

\bibitem{kawauchi2020convolutional}
K.~Kawauchi, S.~Furuya, K.~Hirata, C.~Katoh, O.~Manabe, K.~Kobayashi,
  S.~Watanabe, and T.~Shiga, ``A convolutional neural network-based system to
  classify patients using fdg pet/ct examinations,'' \emph{BMC cancer},
  vol.~20, no.~1, pp. 1--10, 2020.

\bibitem{teramoto2016automated}
A.~Teramoto, H.~Fujita, O.~Yamamuro, and T.~Tamaki, ``Automated detection of
  pulmonary nodules in pet/ct images: Ensemble false-positive reduction using a
  convolutional neural network technique,'' \emph{Medical physics}, vol.~43,
  no. 6Part1, pp. 2821--2827, 2016.

\bibitem{xu2018automated}
L.~Xu, G.~Tetteh, J.~Lipkova, Y.~Zhao, H.~Li, P.~Christ, M.~Piraud, A.~Buck,
  K.~Shi, and B.~H. Menze, ``Automated whole-body bone lesion detection for
  multiple myeloma on 68ga-pentixafor pet/ct imaging using deep learning
  methods,'' \emph{Contrast media \& molecular imaging}, vol. 2018, 2018.

\bibitem{jamieson2012breast}
A.~R. Jamieson, K.~Drukker, and M.~L. Giger, ``Breast image feature learning
  with adaptive deconvolutional networks,'' in \emph{Medical Imaging 2012:
  Computer-Aided Diagnosis}, vol. 8315.\hskip 1em plus 0.5em minus 0.4em\relax
  International Society for Optics and Photonics, 2012, p. 831506.

\bibitem{liu2015tumor}
X.~Liu, J.~Shi, and Q.~Zhang, ``Tumor classification by deep polynomial network
  and multiple kernel learning on small ultrasound image dataset,'' in
  \emph{International Workshop on Machine Learning in Medical Imaging}.\hskip
  1em plus 0.5em minus 0.4em\relax Springer, 2015, pp. 313--320.

\bibitem{shi2016stacked}
J.~Shi, S.~Zhou, X.~Liu, Q.~Zhang, M.~Lu, and T.~Wang, ``Stacked deep
  polynomial network based representation learning for tumor classification
  with small ultrasound image dataset,'' \emph{Neurocomputing}, vol. 194, pp.
  87--94, 2016.

\bibitem{han2017deep}
S.~Han, H.-K. Kang, J.-Y. Jeong, M.-H. Park, W.~Kim, W.-C. Bang, and Y.-K.
  Seong, ``A deep learning framework for supporting the classification of
  breast lesions in ultrasound images,'' \emph{Physics in Medicine \& Biology},
  vol.~62, no.~19, p. 7714, 2017.

\bibitem{antropova2017deep}
N.~Antropova, B.~Q. Huynh, and M.~L. Giger, ``A deep feature fusion methodology
  for breast cancer diagnosis demonstrated on three imaging modality
  datasets,'' \emph{Medical physics}, vol.~44, no.~10, pp. 5162--5171, 2017.

\bibitem{liu2017feature}
T.~Liu, S.~Xie, Y.~Zhang, J.~Yu, L.~Niu, and W.~Sun, ``Feature selection and
  thyroid nodule classification using transfer learning,'' in \emph{2017 IEEE
  14th International Symposium on Biomedical Imaging (ISBI 2017)}.\hskip 1em
  plus 0.5em minus 0.4em\relax IEEE, 2017, pp. 1096--1099.

\bibitem{liu2017classification}
T.~Liu, S.~Xie, J.~Yu, L.~Niu, and W.~Sun, ``Classification of thyroid nodules
  in ultrasound images using deep model based transfer learning and hybrid
  features,'' in \emph{2017 IEEE International Conference on Acoustics, Speech
  and Signal Processing (ICASSP)}.\hskip 1em plus 0.5em minus 0.4em\relax IEEE,
  2017, pp. 919--923.

\bibitem{petscharnig2017inception}
S.~Petscharnig, K.~Sch{\"o}ffmann, and M.~Lux, ``An inception-like cnn
  architecture for gi disease and anatomical landmark classification.'' in
  \emph{MediaEval}, 2017.

\bibitem{borgli2019automatic}
R.~J. Borgli, H.~K. Stensland, M.~A. Riegler, and P.~Halvorsen, ``Automatic
  hyperparameter optimization for transfer learning on medical image datasets
  using bayesian optimization,'' in \emph{2019 13th International Symposium on
  Medical Information and Communication Technology (ISMICT)}.\hskip 1em plus
  0.5em minus 0.4em\relax IEEE, 2019, pp. 1--6.

\bibitem{agrawal2019evaluating}
T.~Agrawal, R.~Gupta, and S.~Narayanan, ``On evaluating cnn representations for
  low resource medical image classification,'' in \emph{ICASSP 2019-2019 IEEE
  International Conference on Acoustics, Speech and Signal Processing
  (ICASSP)}.\hskip 1em plus 0.5em minus 0.4em\relax IEEE, 2019, pp. 1363--1367.

\bibitem{ahmedt2018hierarchical}
D.~Ahmedt-Aristizabal, C.~Fookes, S.~Denman, K.~Nguyen, T.~Fernando,
  S.~Sridharan, and S.~Dionisio, ``A hierarchical multimodal system for motion
  analysis in patients with epilepsy,'' \emph{Epilepsy \& Behavior}, vol.~87,
  pp. 46--58, 2018.

\bibitem{ahmedt2018deep}
D.~Ahmedt-Aristizabal, K.~Nguyen, S.~Denman, S.~Sridharan, S.~Dionisio, and
  C.~Fookes, ``Deep motion analysis for epileptic seizure classification,'' in
  \emph{2018 40th Annual International Conference of the IEEE Engineering in
  Medicine and Biology Society (EMBC)}.\hskip 1em plus 0.5em minus 0.4em\relax
  IEEE, 2018, pp. 3578--3581.

\bibitem{ahmedt2019motion}
D.~Ahmedt-Aristizabal, M.~S. Sarfraz, S.~Denman, K.~Nguyen, C.~Fookes,
  S.~Dionisio, and R.~Stiefelhagen, ``Motion signatures for the analysis of
  seizure evolution in epilepsy,'' in \emph{2019 41st Annual International
  Conference of the IEEE Engineering in Medicine and Biology Society
  (EMBC)}.\hskip 1em plus 0.5em minus 0.4em\relax IEEE, 2019, pp. 2099--2105.

\bibitem{ahmedt2019aberrant}
D.~Ahmedt-Aristizabal, C.~Fookes, S.~Denman, K.~Nguyen, S.~Sridharan, and
  S.~Dionisio, ``Aberrant epileptic seizure identification: A computer vision
  perspective,'' \emph{Seizure}, vol.~65, pp. 65--71, 2019.

\bibitem{gurovich2019identifying}
Y.~Gurovich, Y.~Hanani, O.~Bar, G.~Nadav, N.~Fleischer, D.~Gelbman,
  L.~Basel-Salmon, P.~M. Krawitz, S.~B. Kamphausen, M.~Zenker \emph{et~al.},
  ``Identifying facial phenotypes of genetic disorders using deep learning,''
  \emph{Nature medicine}, vol.~25, no.~1, pp. 60--64, 2019.

\bibitem{marbach2019discovery}
F.~Marbach, C.~F. Rustad, A.~Riess, D.uki{\'c}, T.-C. Hsieh, I.~Jobani,
  T.~Prescott, A.~Bevot, F.~Erger, G.~Houge \emph{et~al.}, ``The discovery of a
  lemd2-associated nuclear envelopathy with early progeroid appearance suggests
  advanced applications for ai-driven facial phenotyping,'' \emph{The American
  Journal of Human Genetics}, vol. 104, no.~4, pp. 749--757, 2019.

\bibitem{yang2019development}
J.~Yang, K.~Zhang, H.~Fan, Z.~Huang, Y.~Xiang, J.~Yang, L.~He, L.~Zhang,
  Y.~Yang, R.~Li \emph{et~al.}, ``Development and validation of deep learning
  algorithms for scoliosis screening using back images,'' \emph{Communications
  biology}, vol.~2, no.~1, pp. 1--8, 2019.

\bibitem{wu2019studies}
Z.~Wu, S.~Zhao, Y.~Peng, X.~He, X.~Zhao, K.~Huang, X.~Wu, W.~Fan, F.~Li,
  M.~Chen \emph{et~al.}, ``Studies on different cnn algorithms for face skin
  disease classification based on clinical images,'' \emph{IEEE Access},
  vol.~7, pp. 66\,505--66\,511, 2019.

\bibitem{subasi2019practical}
A.~Subasi, \emph{Practical guide for biomedical signals analysis using machine
  learning techniques: A MATLAB based approach}.\hskip 1em plus 0.5em minus
  0.4em\relax Academic Press, 2019.

\bibitem{liu2018multiple}
W.~Liu, Q.~Huang, S.~Chang, H.~Wang, and J.~He, ``Multiple-feature-branch
  convolutional neural network for myocardial infarction diagnosis using
  electrocardiogram,'' \emph{Biomedical Signal Processing and Control},
  vol.~45, pp. 22--32, 2018.

\bibitem{takalo2018inter}
J.~Takalo-Mattila, J.~Kiljander, and J.-P. Soininen, ``Inter-patient ecg
  classification using deep convolutional neural networks,'' in \emph{2018 21st
  Euromicro Conference on Digital System Design (DSD)}.\hskip 1em plus 0.5em
  minus 0.4em\relax IEEE, 2018, pp. 421--425.

\bibitem{plesinger2017automatic}
F.~Plesinger, P.~Nejedly, I.~Viscor, J.~Halamek, and P.~Jurak, ``Automatic
  detection of atrial fibrillation and other arrhythmias in holter ecg
  recordings using rhythm features and neural networks,'' in \emph{2017
  Computing in Cardiology (CinC)}.\hskip 1em plus 0.5em minus 0.4em\relax IEEE,
  2017, pp. 1--4.

\bibitem{jun2018ecg}
T.~J. Jun, H.~M. Nguyen, D.~Kang, D.~Kim, D.~Kim, and Y.-H. Kim, ``Ecg
  arrhythmia classification using a 2-d convolutional neural network,''
  \emph{arXiv preprint arXiv:1804.06812}, 2018.

\bibitem{acharya2019deep}
U.~R. Acharya, H.~Fujita, S.~L. Oh, Y.~Hagiwara, J.~H. Tan, M.~Adam, and
  R.~San~Tan, ``Deep convolutional neural network for the automated diagnosis
  of congestive heart failure using ecg signals,'' \emph{Applied Intelligence},
  vol.~49, no.~1, pp. 16--27, 2019.

\bibitem{singh2018classification}
S.~Singh, S.~K. Pandey, U.~Pawar, and R.~R. Janghel, ``Classification of ecg
  arrhythmia using recurrent neural networks,'' \emph{Procedia computer
  science}, vol. 132, pp. 1290--1297, 2018.

\bibitem{schwab2017beat}
P.~Schwab, G.~C. Scebba, J.~Zhang, M.~Delai, and W.~Karlen, ``Beat by beat:
  Classifying cardiac arrhythmias with recurrent neural networks,'' in
  \emph{2017 Computing in Cardiology (CinC)}.\hskip 1em plus 0.5em minus
  0.4em\relax IEEE, 2017, pp. 1--4.

\bibitem{sujadevi2017real}
V.~Sujadevi, K.~Soman, and R.~Vinayakumar, ``Real-time detection of atrial
  fibrillation from short time single lead ecg traces using recurrent neural
  networks,'' in \emph{The International Symposium on Intelligent Systems
  Technologies and Applications}.\hskip 1em plus 0.5em minus 0.4em\relax
  Springer, 2017, pp. 212--221.

\bibitem{faust2018automated}
O.~Faust, A.~Shenfield, M.~Kareem, T.~R. San, H.~Fujita, and U.~R. Acharya,
  ``Automated detection of atrial fibrillation using long short-term memory
  network with rr interval signals,'' \emph{Computers in biology and medicine},
  vol. 102, pp. 327--335, 2018.

\bibitem{liu2018classification}
M.~Liu and Y.~Kim, ``Classification of heart diseases based on ecg signals
  using long short-term memory,'' in \emph{2018 40th Annual International
  Conference of the IEEE Engineering in Medicine and Biology Society
  (EMBC)}.\hskip 1em plus 0.5em minus 0.4em\relax IEEE, 2018, pp. 2707--2710.

\bibitem{yildirim2018novel}
{\"O}.~Yildirim, ``A novel wavelet sequence based on deep bidirectional lstm
  network model for ecg signal classification,'' \emph{Computers in biology and
  medicine}, vol.~96, pp. 189--202, 2018.

\bibitem{zhang2017patient}
C.~Zhang, G.~Wang, J.~Zhao, P.~Gao, J.~Lin, and H.~Yang, ``Patient-specific ecg
  classification based on recurrent neural networks and clustering technique,''
  in \emph{2017 13th IASTED International Conference on Biomedical Engineering
  (BioMed)}.\hskip 1em plus 0.5em minus 0.4em\relax IEEE, 2017, pp. 63--67.

\bibitem{chu2017individual}
L.~Chu, R.~Qiu, H.~Liu, Z.~Ling, T.~Zhang, and J.~Wang, ``Individual
  recognition in schizophrenia using deep learning methods with random forest
  and voting classifiers: Insights from resting state eeg streams,''
  \emph{arXiv preprint arXiv:1707.03467}, 2017.

\bibitem{ahmedt2020identification}
D.~Ahmedt~Aristizabal, T.~Fernando, S.~Denman, J.~E. Robinson, S.~Sridharan,
  P.~J. Johnston, K.~R. Laurens, and C.~Fookes, ``Identification of children at
  risk of schizophrenia via deep learning and eeg responses,'' \emph{IEEE
  Journal of Biomedical and Health Informatics}, pp. 1--7, 2020.

\bibitem{acharya2018deep}
U.~R. Acharya, S.~L. Oh, Y.~Hagiwara, J.~H. Tan, and H.~Adeli, ``Deep
  convolutional neural network for the automated detection and diagnosis of
  seizure using eeg signals,'' \emph{Computers in biology and medicine}, vol.
  100, pp. 270--278, 2018.

\bibitem{o2018investigating}
A.~O’Shea, G.~Lightbody, G.~Boylan, and A.~Temko, ``Investigating the impact
  of cnn depth on neonatal seizure detection performance,'' in \emph{2018 40th
  Annual International Conference of the IEEE Engineering in Medicine and
  Biology Society (EMBC)}.\hskip 1em plus 0.5em minus 0.4em\relax IEEE, 2018,
  pp. 5862--5865.

\bibitem{thodoroff2016learning}
P.~Thodoroff, J.~Pineau, and A.~Lim, ``Learning robust features using deep
  learning for automatic seizure detection,'' in \emph{Machine learning for
  healthcare conference}, 2016, pp. 178--190.

\bibitem{ullah2018automated}
I.~Ullah, M.~Hussain, H.~Aboalsamh \emph{et~al.}, ``An automated system for
  epilepsy detection using eeg brain signals based on deep learning approach,''
  \emph{Expert Systems with Applications}, vol. 107, pp. 61--71, 2018.

\bibitem{o2017neonatal}
A.~O'Shea, G.~Lightbody, G.~Boylan, and A.~Temko, ``Neonatal seizure detection
  using convolutional neural networks,'' in \emph{2017 IEEE 27th International
  Workshop on Machine Learning for Signal Processing (MLSP)}.\hskip 1em plus
  0.5em minus 0.4em\relax IEEE, 2017, pp. 1--6.

\bibitem{talathi2017deep}
S.~S. Talathi, ``Deep recurrent neural networks for seizure detection and early
  seizure detection systems,'' \emph{arXiv preprint arXiv:1706.03283}, 2017.

\bibitem{hussein2018epileptic}
R.~Hussein, H.~Palangi, R.~Ward, and Z.~J. Wang, ``Epileptic seizure detection:
  A deep learning approach,'' \emph{arXiv preprint arXiv:1803.09848}, 2018.

\bibitem{naderi2010analysis}
M.~A. Naderi and H.~Mahdavi-Nasab, ``Analysis and classification of eeg signals
  using spectral analysis and recurrent neural networks,'' in \emph{2010 17th
  Iranian Conference of Biomedical Engineering (ICBME)}.\hskip 1em plus 0.5em
  minus 0.4em\relax IEEE, 2010, pp. 1--4.

\bibitem{ruffini2019deep}
G.~Ruffini, D.~Iba{\~n}ez, M.~Castellano, L.~Dubreuil-Vall, A.~Soria-Frisch,
  R.~Postuma, J.-F. Gagnon, and J.~Montplaisir, ``Deep learning with eeg
  spectrograms in rapid eye movement behavior disorder,'' \emph{Frontiers in
  neurology}, vol.~10, 2019.

\bibitem{khowailed2019neural}
I.~A. Khowailed and A.~Abotabl, ``Neural muscle activation detection: A deep
  learning approach using surface electromyography,'' \emph{Journal of
  biomechanics}, vol.~95, p. 109322, 2019.

\bibitem{olsson2019extraction}
A.~E. Olsson, P.~Sager, E.~Andersson, A.~Bj{\"o}rkman, N.~Male{\v{s}}evi{\'c},
  and C.~Antfolk, ``Extraction of multi-labelled movement information from the
  raw hd-semg image with time-domain depth,'' \emph{Scientific reports},
  vol.~9, no.~1, pp. 1--10, 2019.

\bibitem{dantas2019deep}
H.~Dantas, D.~J. Warren, S.~M. Wendelken, T.~S. Davis, G.~A. Clark, and V.~J.
  Mathews, ``Deep learning movement intent decoders trained with dataset
  aggregation for prosthetic limb control,'' \emph{IEEE Transactions on
  Biomedical Engineering}, vol.~66, no.~11, pp. 3192--3203, 2019.

\bibitem{Zhang2019AbnormalSegmentation}
W.~Zhang, J.~Han, and S.~Deng, ``{Abnormal heart sound detection using temporal
  quasi-periodic features and long short-term memory without segmentation},''
  \emph{Biomedical Signal Processing and Control}, vol.~53, 8 2019.

\bibitem{Maknickas2017RecognitionCoefficients}
V.~Maknickas and A.~Maknickas, ``{Recognition of normal-abnormal
  phonocardiographic signals using deep convolutional neural networks and
  mel-frequency spectral coefficients},'' \emph{Physiological Measurement},
  vol.~38, no.~8, pp. 1671--1684, 7 2017.

\bibitem{potes2016ensemble}
C.~Potes, S.~Parvaneh, A.~Rahman, and B.~Conroy, ``Ensemble of feature-based
  and deep learning-based classifiers for detection of abnormal heart sounds,''
  in \emph{2016 Computing in Cardiology Conference (CinC)}.\hskip 1em plus
  0.5em minus 0.4em\relax IEEE, 2016, pp. 621--624.

\bibitem{sannino2019artificial}
G.~Sannino, N.~Bouguila, G.~De~Pietro, and A.~Celesti, ``Artificial
  intelligence for mobile health data analysis and processing,'' \emph{Mobile
  Information Systems}, vol. 2019, 2019.

\bibitem{potluri2019deep}
S.~Potluri, S.~Ravuri, C.~Diedrich, and L.~Schega, ``Deep learning based gait
  abnormality detection using wearable sensor system,'' in \emph{2019 41st
  Annual International Conference of the IEEE Engineering in Medicine and
  Biology Society (EMBC)}.\hskip 1em plus 0.5em minus 0.4em\relax IEEE, 2019,
  pp. 3613--3619.

\bibitem{mauldin2018smartfall}
T.~R. Mauldin, M.~E. Canby, V.~Metsis, A.~H. Ngu, and C.~C. Rivera,
  ``Smartfall: A smartwatch-based fall detection system using deep learning,''
  \emph{Sensors}, vol.~18, no.~10, p. 3363, 2018.

\end{thebibliography}

\pagebreak

\appendix[Deep Learning for Medical Anomaly Detection - A Survey: Supplementary Material]

\section{Types of Data}
\subsection{Biomedical Imagining}
\textbf{X-ray radiography}: X-rays have shorter wave lengths than visible light and can pass through most tissue types in the human body. However, the calcium contained in bones is denser and scatters the x-rays. The film that sits on the opposite side of the x-ray source is a negative image such that areas that are exposed to more light appear darker. Therefore, as more x-rays penetrate tissues such as lungs and mussels, these areas are darkened on the film and the bones appear as brighter regions. X-ray imaging is typically used for various diagnostic purposes, including detecting bone fractures, dental problems, pneumonia, and certain types of tumor.

Considering application of deep learning models for abnormality detection using X-ray images, in \cite{bhandari2019chest} the authors utilised a CNN architecture to detect abnormalities in chest x-ray images, while in \cite{tataru2017deep, tang2020automated} the authors investigate the utilisation of different off-the-shelf pre-trained CNN architecture such as VGG-16 and GoogLeNet to classify X-ray images. In \cite{wang2018chestnet} the authors purpose an architecture called ChestNet where a two branch framework is proposed for diagnosis. A classification branch detects abnormalities, and an attention branch is introduced which tries to uncover the correlation between class labels and the locations of pathological abnormalities. In a different line of work, the authors of \cite{pham2020interpreting} exploit the ability of a multi-label classification framework to capture the hierarchical dependencies among different abnormality classes. Instead of performing binary normal/abnormal classification, they formulate the task as multi-class classification where the model predicts the abnormality class of the input x-ray. Most recently, deep learning methods have been introduced for screening COVID-19 using chest x-rays \cite{zhang2020viral, basu2020deep}

\textbf{Computed Tomography scan (CT):}
In CT imaging, cross sectional images of the body are generated using a narrow beam of x-rays that are emitted while the patient is quickly rotated. CT imaging collects a number of cross sectional slices which are stacked together to generate a 3 dimensional representation, which is more informative than a conventional X-ray image. Fig. \ref{fig:CT_machine} illustrates an example of the CT scan procedure of the abdomen.
\begin{figure}[htbp]
  \centering
  \includegraphics[width=\linewidth]{figures/data_types/wikimedia/CT.jpg}
    \caption{Illustration of the Computed Tomography scan (CT) procedure. Image  \href{https://www.cancer.gov/publications/dictionaries/cancer-terms/def/computed-tomography-scan}{source}}
    \label{fig:CT_machine}
\end{figure}
CT scans are a popular diagnostic tool when identifying disease or injury within various regions of the body. Applications include detecting tumors or lesions in the abdomen, and localising head injuries, tumors, and clots. They are also used for diagnosing complex bone fractures and bone tumors. Fig. \ref{fig:CT_normal_abnormal} shows an examples of chest CT scans of healthy and diseased patients. This example is taken from \href{https://www.kaggle.com/plameneduardo/sarscov2-ctscan-dataset}{SARS-CoV-2 CT scan dataset} where the diseased subject has COVID-19 symptoms.

\begin{figure}[htbp]
  \centering
  \includegraphics[width=.95\linewidth]{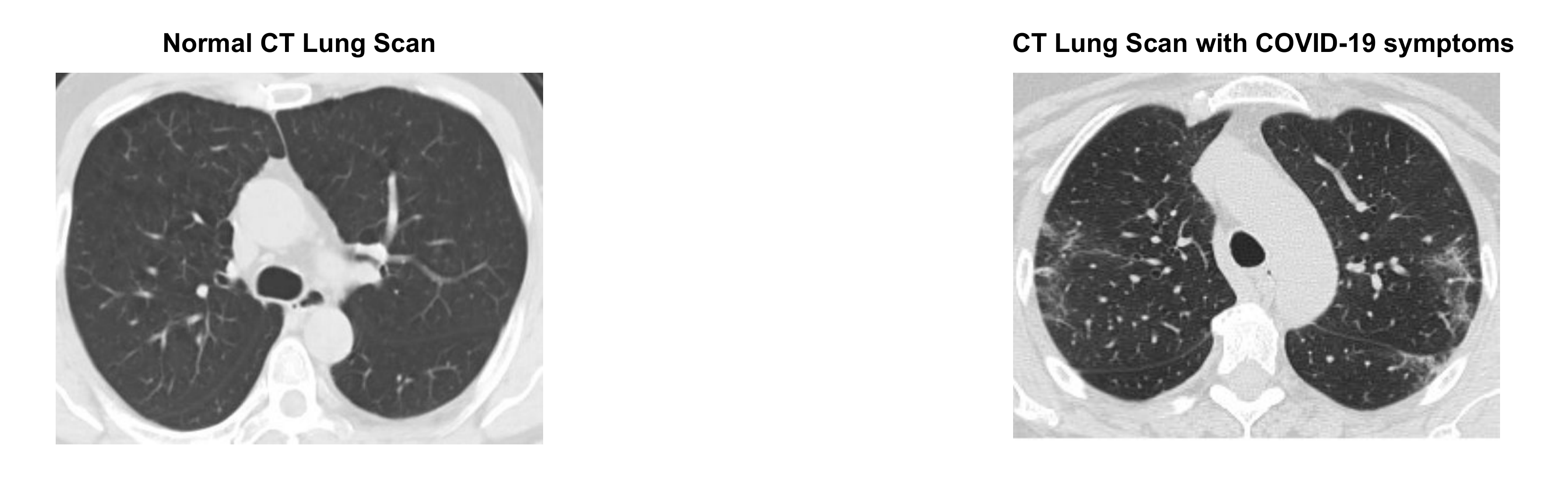}
    \caption{Lung CT scans of healthy and diseased subjects taken from the  \href{https://www.kaggle.com/plameneduardo/sarscov2-ctscan-dataset}{SARS-CoV-2 CT scan dataset}}
\label{fig:CT_normal_abnormal}
\end{figure}

There exist a multitude of works that utilise deep learning for diagnosing symptoms in CT scans. For instance, in \cite{shah2020diagnosis, mishra2020identifying, perumal2020detection, sharma2020drawing, harmon2020artificial} authors utilise chest CT scans to detect COVID-19 pneumonia symptoms. Makde et. al \cite{makde2017deep} proposed an algorithm to detect kidney tumors in CT images while \cite{huang2017lung, jakimovski2019using} detect lung cancers using CT scans.

\begin{figure}[htbp]
  \centering
  \subfloat[][]{\includegraphics[width=.5\linewidth]{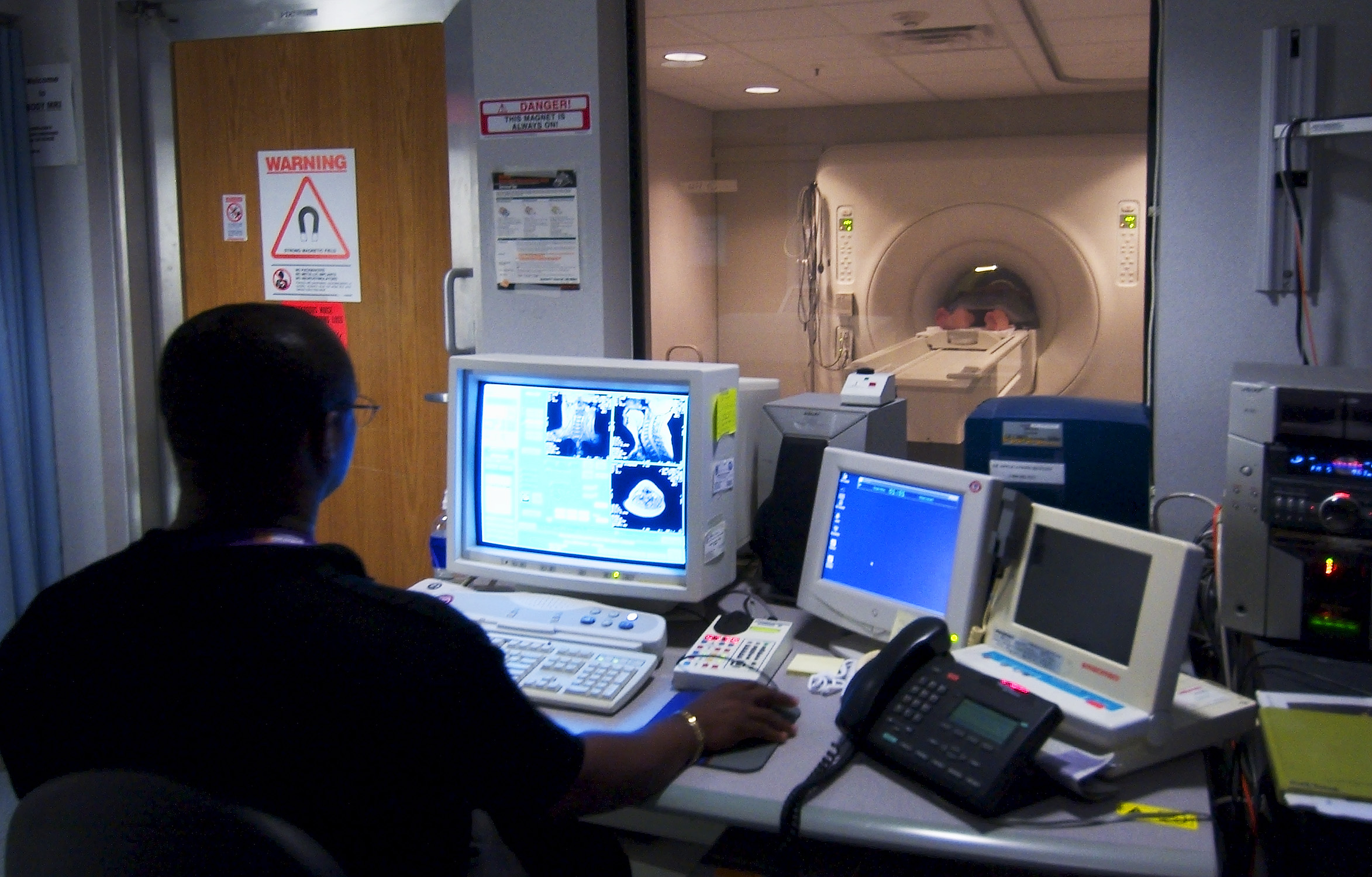}}
  \subfloat[][]{\includegraphics[width=.32\linewidth]{figures/data_types/MRI.jpg}}
    \caption{(a)An example of the Magnetic Resonance Imaging (MRI) system. Image   \href{https://commons.wikimedia.org/wiki/Main_Page}{source}. (b) An MRI image with a brain tumor taken from \href{https://www.kaggle.com/navoneel/brain-mri-images-for-brain-tumor-detection}{Kaggle Brain MRI Images for Brain Tumor Detection dataset.}
}
    \label{fig:MRI_machine}
\end{figure}

\textbf{Magnetic Resonance Imaging (MRI):}
As the name implies MRI employs a magnetic field for imagining by forcing protons in the body to align with the applied field. Fig. \ref{fig:MRI_machine} (a) provides an illustration of the MRI procedure.

Specifically, the protons in the human body spin and create a small magnetic field. When a strong magnetic field such as from the MRI machine is introduced, the protons align with that field. Then a radio frequency pulse is introduced which disrupts the alignment. When the radio frequency pulse is turned off the protons discharge energy and try to re-align with the magnetic field. The energy released varies for different tissue types, allowing the MRI scan to segregate different regions. Therefore, MRIs are typically used to image non-bony or soft tissue regions of the human body. Comparison studies have shown that the brain, spinal cord, nerves and muscles are better captured by MRIs than CT scans. Therefore, MRI is the modality of choice for tasks such as brain tumor detection and identifying tissue damage.

A multitude of research has focused on applying deep learning techniques to detect abnormalities in MRIs. For instance, in \cite{afshar2018brain, abiwinanda2019brain, ari2018deep} a Convolutional Neural Network (CNN) is utilised to detect abnormalities, while in \cite{zhou2018holistic} a hybrid model using a combination of CNNs and a Recurrent Neural Network (RNN) is employed. In \cite{xu2015deep} the authors carried out an investigation to see how CNN activations are transferred from a pre-training natural image classification task to the MRI domain for a brain tumor detection task. 

\begin{figure}[htbp]
  \centering
  \includegraphics[width=\linewidth]{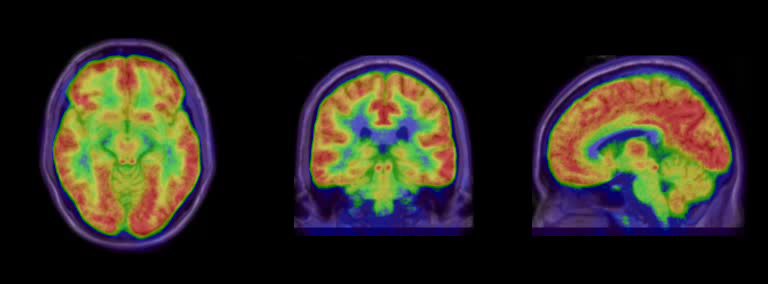}
    \caption{An example PET scan. Image taken from \href{https://www.kaggle.com/c/pet-radiomics-challenges}{PET radiomics challenges}}
\label{fig:PET}
\end{figure}

\textbf{Positron Emission Tomography (PET)}: PET works by injecting, swallowing or inhaling a radioactive material (tracer), and this material is collected in regions with higher levels of chemical activity which usually correspond to areas of disease. The emitted energy from the radioactive material is detected by a ring of detector placed within the PET scanner. PET scans are valuable for revealing or evaluating several conditions including different cancers types (Esophageal, Melanoma, Thyroid, Cervical, etc), heart conditions such as decreased blood flow to the heart, and brain disorders. Fig. \ref{fig:PET} illustrates a PET brain image. 

Regarding deep learning based anomaly detection using PET data, the work of Deng et. al \cite{deng2017automated} purposes a framework to detect lymphomas. A CNN based system is proposed in \cite{kawauchi2020convolutional} which can classify whole-body fluorodeoxyglucose PET scans. A PET-CT hybrid system is proposed in \cite{teramoto2016automated} where the authors employ a CNN to extract features from the PET and CT images, and an ensemble of rule based and SVM classifiers are employed to get the final classification. An automated whole-body bone lesion detection framework is proposed in \cite{xu2018automated} which also uses a combination of CT and PET scans, and in this work a complete deep learning pipeline in contrast to the two stage approach of \cite{teramoto2016automated}.

\textbf{Ultrasound}: The main component in an ultrasound imaging device is a transducer which converts electrical energy to sound waves. This is done using an array of piezoelectric crystals in the transducer, which vibrate when an electric signal is applied and generate high frequency sound waves. These ultrasound waves travel through the body and they are reflected back by different tissues with different characteristics at different depths. The piezoelectric crystals detect the reflected ultrasound waves, and generate back the electric signal. The returned electric signal is converted to an image via computer. 

Ultrasound is a non-invasive method for imaging internal organs. One of the primary applications of ultrasound is to monitor the growth and development of the fetus during pregnancy. They are also used as tool to detect abnormalities in the heart, blood vessels, breasts, abdominal organs, etc. One key characteristic of ultrasound imaging is it's ability to be displayed in either 2D, 3D, or 4D (where the 3D image is visualised together with motion). 

\begin{figure}[htbp]
  \centering
  \subfloat[][]{\includegraphics[width=.35\linewidth]{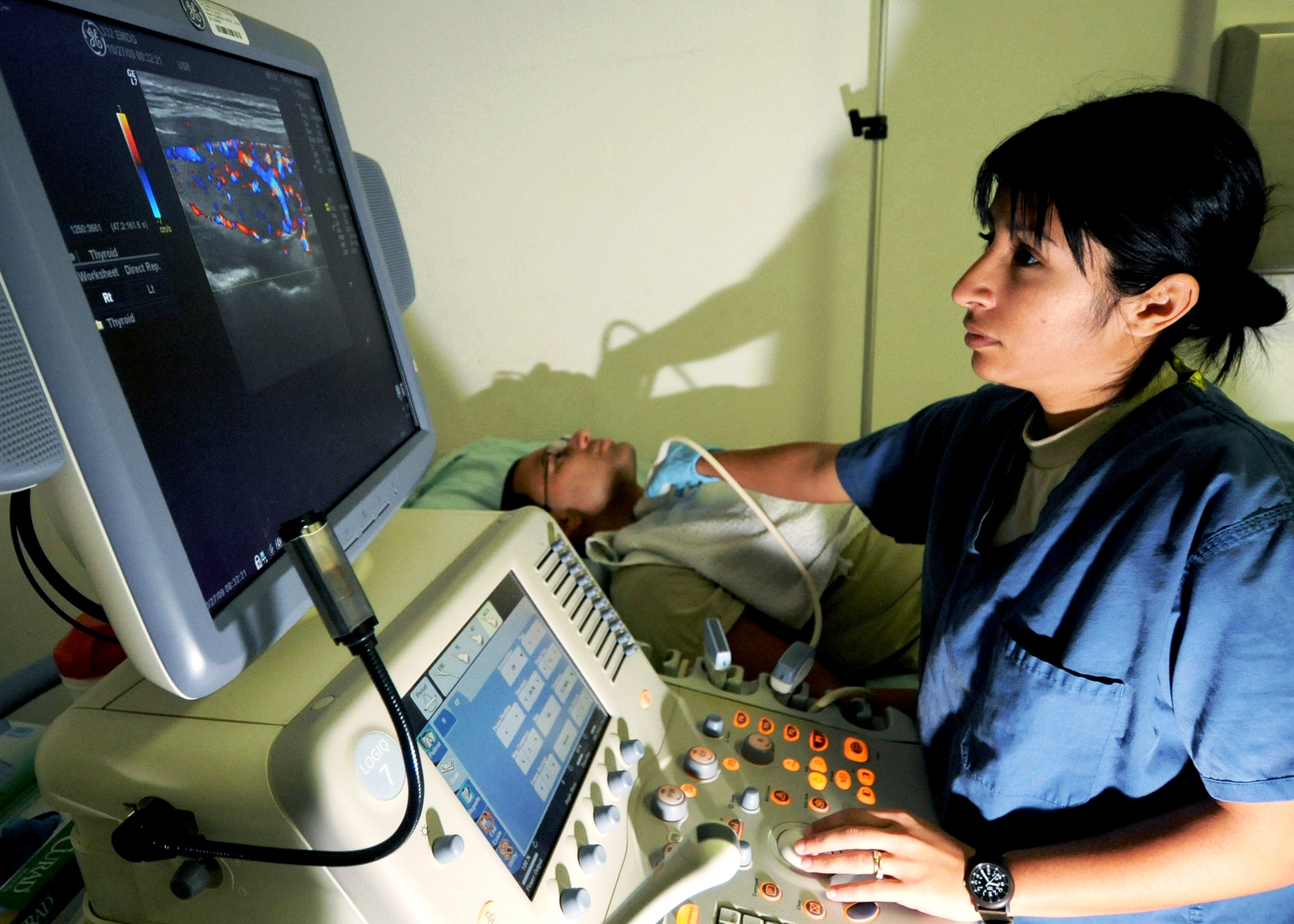}}
  \subfloat[][]{\includegraphics[width=.345\linewidth]{figures/data_types/10_1.jpg}}
    \caption{(a)Illustration of an ultrasound scan. Image   \href{https://commons.wikimedia.org/wiki/Main_Page}{source}. (b) An ultrasound image of the neck which is taken from \href{https://www.kaggle.com/navoneel/brain-mri-images-for-brain-tumor-detection}{Kaggle Ultrasound Nerve Segmentation dataset.}
}
    \label{fig:ultrasound}
\end{figure}

One of the first studies that applied deep learning to ultrasound images was conducted by Jamieson et. al in \cite{jamieson2012breast}, where they proposed an unsupervised deep learning model to classify breast tumors in ultrasound images. Subsequently, supervised deep learning techniques \cite{liu2015tumor, shi2016stacked, han2017deep} have been extensively applied for abnormality detection in ultrasound images. Furthermore, strategies to fuse features extracted from pre-trained deep learning models and hand crafted features are proposed in \cite{antropova2017deep, liu2017feature, liu2017classification}.

\textbf{Medical Optical Imaging}
Optical Imagining is similar to X-ray and PET imaging techniques which were discussed earlier, however, in contrast to those techniques which use radiation to create images, optical imaging uses visible light. This is a key advantage of medical optical imaging as it reduces the exposure of the patient to harmful radiation. Furthermore, optical imaging is particularly useful when imaging soft-tissue as they have different absorption characteristics when exposed to light. Optical imaging is often used as a complimentary modality to other techniques such as MRI or X-rays, and can be designed to capture higher resolution images. 

\begin{figure}[htbp]
  \centering
  \subfloat[][]{\includegraphics[width=.35\linewidth]{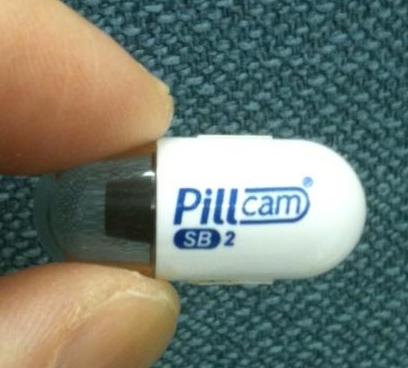}}
  \subfloat[][]{\includegraphics[width=.325\linewidth]{figures/data_types/nurths_dataset_image.png}}
    \caption{(a)An image of a capsule endoscopy device. Image   \href{https://commons.wikimedia.org/wiki/Main_Page}{source}. (b) An endoscopy image of the gastrointestinal tract which is taken from \href{https://datasets.simula.no/nerthus/}{The Nerthus endoscopy dataset.}
}
    \label{fig:optical}
\end{figure}

The endoscopy is one of the most common optical imaging techniques where a physician inserts an imaging device and light source inside the patient and a series of images are taken for later diagnosis. For example, when diagnosing abnormalities in the digestive system, the endoscope is inserted through the patient's mouth and the captured images can be used to diagnose conditions such bowel diseases, gastrointestinal bleeding and cancer. Fig. \ref{fig:optical} (a) shows an image of a capsule endoscopy device, where a small wireless camera is placed within a capsule like component. The patient swallows this capsule and the endoscopy device takes a video of the patient's gastrointestinal tract. 

 Deep learning techniques have predominately been applied to extract discriminative features from pre-trained CNNs in computer aided video endoscopy analysis   \cite{pogorelov2017kvasir,petscharnig2017inception, borgli2019automatic}. The authors of these works leverage the transferability of the learned features among different visual datasets, and try to adapt feature extraction models trained on large data corpora such as ImageNet \cite{deng2009imagenet} to extract features from small scale endoscopy datasets. In  \cite{agrawal2019evaluating}, the authors tested multiple existing pre-trained CNN network features to better detect abnormalities.
 
 Apart from endoscopy image analysis, optical images and videos have been use for other diagnostic purposes. For instance, in \cite{ahmedt2018hierarchical, ahmedt2018deep, ahmedt2019motion, ahmedt2019aberrant} the authors use deep learning to obtain semiological features to classify different types of epilepsy. Specifically, they extract facial features together with upper-body, hand movement features, and pose features using deep learning techniques for classification. Facial features extracted from deep learning techniques have also been extensively analysed in \cite{gurovich2019identifying, marbach2019discovery} for identification of genetic disorders from facial images. In addition, works such as \cite{yang2019development} have tested the viability of optical images for scoliosis screening, and in \cite{wu2019studies} for skin disease recognition.
 
 \textbf{Summary}
 
 In summary, most biomedical imaging techniques utilise similar types of imagining sensors together with different energy sources, such as X-ray or visible light for imaging, producing images with 2 or more dimensions. One interesting characteristic of biomedical imaging is its ability to create multi-dimensional spatial representations ranging from 2D (such as X-ray), 3D (such as CT) and 4D (such as ultrasound).This has motivated most of the abnormality detection techniques in biomedical imaging to use CNN based architectures to model this data. Please refer to Sec. 2.2 and Sec. 2.3 of the main document where we discuss such approaches in detail. 
 
 Tab. \ref{tab:list_of_datasets_imagin} summarises a list of datasets for biomedical imagining which are publicly available and have been extensively applied in machine learning research.
 
 \begin{table*}[htbp]
\caption{List of publicly available datasets for biomedical imaging}
\centering
\begin{adjustbox}{max width=\textwidth} 
\begin{tabular}{|c|c|p{0.3\textwidth}|p{0.5\textwidth}|c|}
\hline
Data Type              & Task                               & Dataset Name                                             & Description                                                                                                                                                                                                                                                             & References                                                              \\ \hline
\multirow{3}{*}{X-ray} & \multirow{2}{*}{Chest radiography} & CheXpert                                                 & The dataset consists of 224,316 chest radiographs of 65,240 patients labeled for the presence of 14 common chest radiographic observations.                                                                                                                             & \href{https://stanfordmlgroup.github.io/competitions/chexpert/}{link}                \\ \cline{3-5} 
                      &                                    & ChestX-ray8                                              & Comprises 108,948 frontal view X-ray images of 32,717 unique patients with the text mined for eight disease labels                                                                                                                                                    & \href{https://github.com/TRKuan/cxr8}{link}                                          \\ \cline{2-5} 
                      & Bone radiography                   & MURA (musculoskeletal radiographs)                       & Contains 9,045 normal and 5,818 abnormal musculoskeletal radiographic studies of the upper extremities including the shoulder, humerus, elbow, forearm, wrist, hand, and finger.                                                                                          & \href{https://stanfordmlgroup.github.io/competitions/mura/}{link}                   \\ \hline
\multirow{3}{*}{MRI} & \multirow{2}{*}{Assessment of Alzheimer's Disease} & ADNI                                                 & The dataset consists of 819 subjects (229  Normal, 398 with mild cognitive impairment, and 192 with Alzheimer disease) .                                                                                                                             & \href{http://adni.loni.usc.edu/data-samples/access-data/}{link}                \\ \cline{2-5}
&      Autism  Disorders Identification                            & Autism Brain Imaging Data Exchange (ABIDE)                                              & 539 individuals suffering from Autism spectrum disorders and 573 healthy control subjects.                                                                                                                                                    & \href{http://preprocessed-connectomes-project.org/abide/}{link}                                          \\ \hline

\multirow{3}{*}{Ultrasound} & \multirow{2}{*}{Cardiac functionality assessment} & EchoNet-Dynamic                                                 & includes 10,030 labeled echocardiogram videos and human expert annotations (measurements, tracings, and calculations)  .                                                                                                                             & \href{https://echonet.github.io/dynamic/index.html#access}{link}                \\ \hline
\multirow{3}{*}{CT} & \multirow{2}{*}{Lung cancer Detection} & Lung Image Database Consortium (LIDC)                                                 & includes data from 1018 Participants which comprises 244,527 images.                                                                                                                             & \href{https://wiki.cancerimagingarchive.net/display/Public/LIDC-IDRI#}{link}                 \\ \cline{3-5} 
                      &                                    & RIDER Lung CT                                              & 32 patients with  lung cancer, each of whom underwent two CT scans of the chest within 15 minutes time lag                                                                                                                                                & \href{https://wiki.cancerimagingarchive.net/display/Public/RIDER+Lung+CT}{link}  \\ \hline

\end{tabular}
\end{adjustbox}
\label{tab:list_of_datasets_imagin}
\end{table*}
 
\subsection{Electrical Biomedical Signals}
\begin{figure}[htbp]
  \centering
  \includegraphics[width=\linewidth]{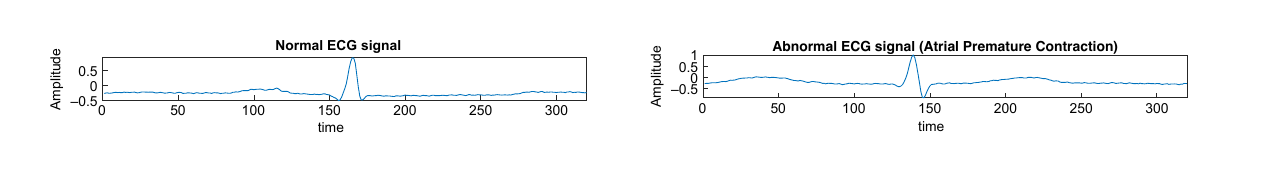}
    \caption{Illustration of normal and abnormal ECG signals. Recreated from \cite{subasi2019practical}}
    \label{fig:ECG}
\end{figure}

\textbf{Electrocardiogram (ECG)}: ECG is a tool to visualise electricity flowing through the heart which creates the heart beat, and starts at the top of the heart and travels to the bottom. At rest, heart cells are negatively charged compared to the outside environment and when they become depolarized they become positively charged. The difference in polarization is captured by the ECG. There are two types of information which can be extracted by analysing the ECG \cite{ebrahimi2020review}. First, by measuring time intervals on an ECG one can screen for irregular electrical activities. Second, the strength of the electrical activity provides an indication of the regions of the heart that are over worked or stressed. 
Fig. \ref{fig:ECG} provides samples of normal and abnormal (Atrial Premature Contraction) ECG signals. ECG is used as a preliminary tool to screen heart arrhythmia and other cardiovascular diseases such as blocked or narrowed arteries. 

Many works have exploited deep learning for ECG analysis. For instance, CNN-based ECG arrhythmia detection is investigated in \cite{liu2018multiple, takalo2018inter, plesinger2017automatic, jun2018ecg, acharya2019deep}. When applying 2D CNN architectures the authors have considered the temporal representation of the 1D signal as an image and passed the 2D representation through CNN kernals for feature extraction. Another area of investigation has been the use of Recurrent Neural Networks (RNNs) to model the temporal evolution of the ECG signal. For instance, in \cite{singh2018classification, schwab2017beat, sujadevi2017real, faust2018automated, liu2018classification, yildirim2018novel} the authors have investigated variants of RNNs such as LSTMs and GRUs while in \cite{zhang2017patient} the authors employ RNNs to model features and then use clustering techniques to identify abnormalities. 

\textbf{Electroencephalogram (EEG)}: The EEG detects electrical activity in the brain, which uses electrical impulses to communicate. To capture the electrical activity, small metal discs (electrodes) are placed on the scalp. The electrical signals captured by these electrodes are amplified in to better visualise brain activity. 

There exist different standards for placing the electrodes on the scalp. Fig. \ref{fig:EEG} (a) shows one popular standard, termed the 10-20 system. The brain regions are divided into  pre-frontal (Fp), frontal (F), temporal (T), parietal (P), occipital (O), and central (C). The number associated with the electrode reflects whether it is placed on the left or right side of the brain. If it is an even-number (2,4,6,8) the electrode is place on the right side of the head, whereas odd numbers (1,3,5,7) are are on the left. Two reference electrodes labelled `A1' and `A2' are placed behind the outer ear.

\begin{figure}[htbp]
  \centering
  \subfloat[][]{\includegraphics[width=.35\linewidth]{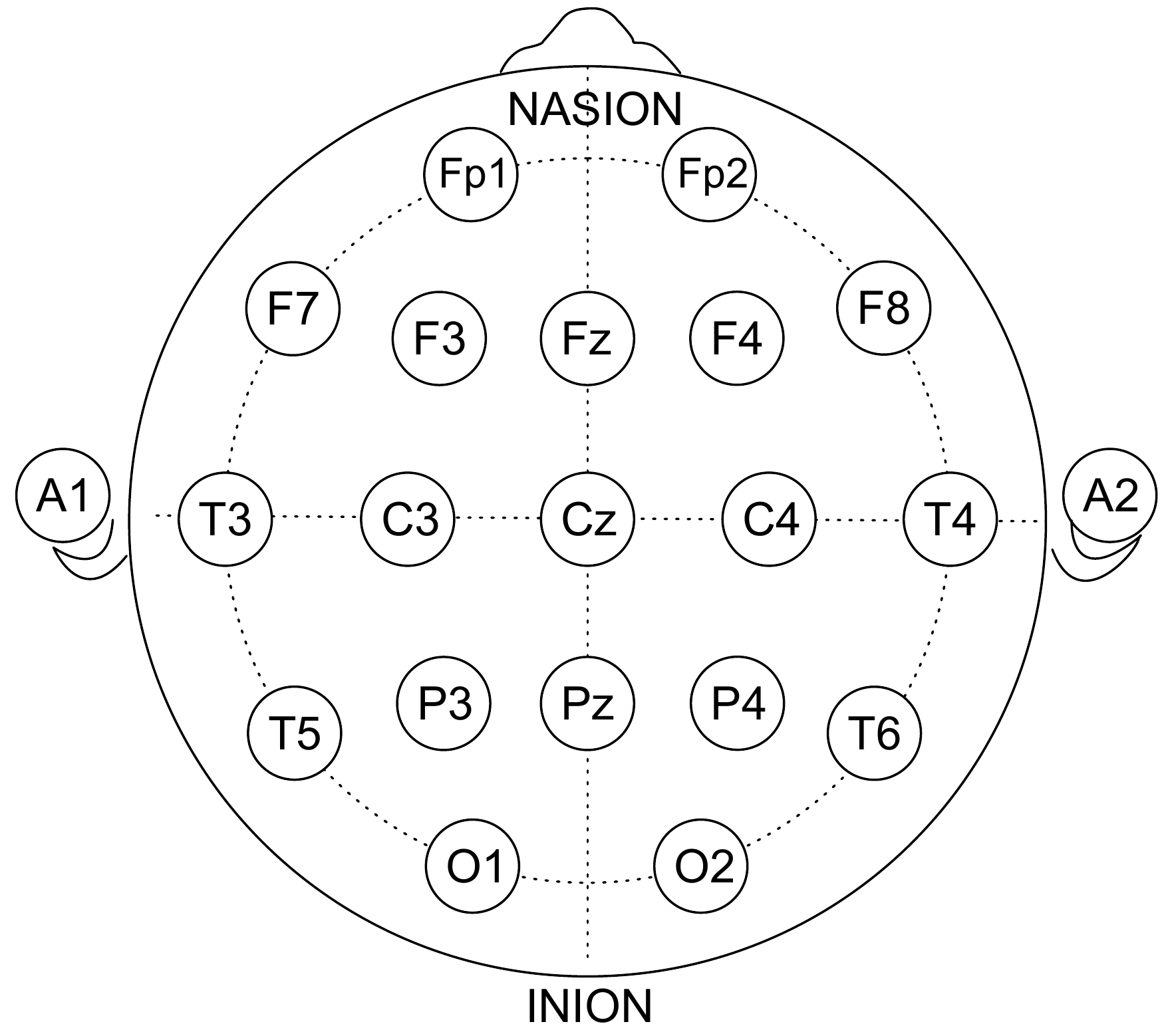}}
  \subfloat[][]{\includegraphics[width=.65\linewidth]{figures/data_types/eeg.png}}
    \caption{(a)Illustration of the 10-20 EEG electrode placement system. (b) Variations of the EEG recording before and after a seizure.}
    \label{fig:EEG}
\end{figure}

EEGs are a prominent tool for observing the cognitive process of a subject. They are often used to study sleep patterns, psychological disorders, brain damage from head injury, and epilepsy. Fig. \ref{fig:EEG} (b) shows how clearly the change in brain activity is captured by the EEG when a patient suffers an epileptic seizure. 

Similar to ECG based analysis, CNNs and RNNs have been a popular choice for EEG analysis. For instance, a CNN based feature extractor is applied in \cite{chu2017individual} for EEG based Schizophrenia detection while in \cite{ahmedt2020identification} the authors use a hybrid of a CNN and LSTM. CNNs are also widely used to analyse EEGs for Epilepsy Detection \cite{acharya2018deep, o2018investigating, thodoroff2016learning, ullah2018automated, o2017neonatal}; though RNNs are also commonly used as in \cite{talathi2017deep, hussein2018epileptic, naderi2010analysis}. When considering deep learning based studies on sleep abnormality detection using EEGs, the authors in \cite{ruffini2019deep} have assessed CNN and RNN architectures together with spectrogram features for diagnosing sleep disorders. 

\textbf{Magnetoencephalography (MEG)}: s a functional neuroimaging mechanism which captures the magnetic fields produced by the brain’s electrical activity, hence, we group MEG with other electrical biomedical signals. Fig. 9 illustrates this process. During synaptic transmission ionic currents flow in the dendrites of neurons and produce a magnetic field. MEG is readily used for the recognition of perceptual and cognitive brain processes, as well as for localising abnormal regions in the brain.

\begin{figure}[htbp]
  \centering
  \includegraphics[width=\linewidth]{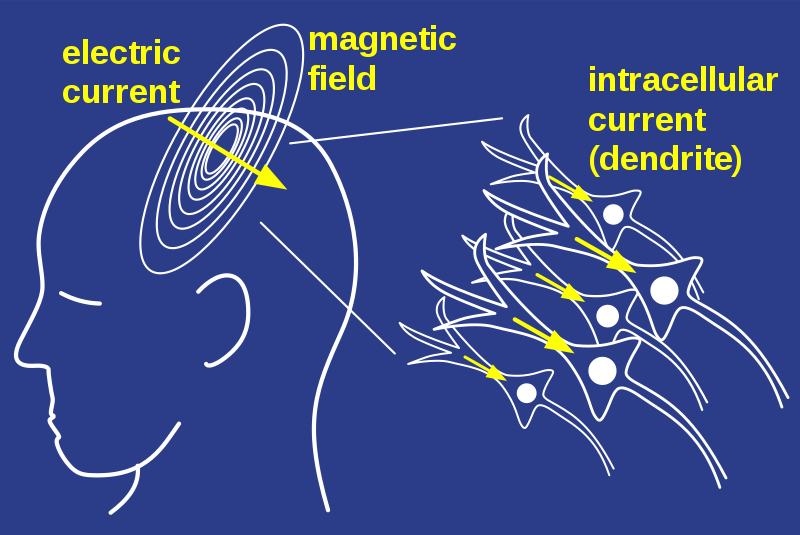}
    \caption{Illustration of brain's magnetic field. Image \href{https://commons.wikimedia.org/wiki/Main_Page}{source}}
    \label{fig:MEG}
\end{figure}

Several studies \cite{lardone2018mindfulness, jacini2018amnestic, rucco2019mutations, hasasneh2018deep, lopez2020detection, aoe2019automatic} have investigated the utility of MEG signals for the detection of anomalous brain activities and conditions. Specifically, in \cite{hasasneh2018deep} the authors propose a spatio-temporal neural network structure to identify ocular and cardiac activities. 1D and 2D CNN architectures for early detection of Alzheimer’s disease based on MEG signals is proposed in \cite{lopez2020detection}. The authors hand-craft 870 features to extract from the MEG signal, and apply a 1D convolution followed by two fully connected layers prior to generating a binary classification (i.e. healthy or Mild Cognitive Impairment). For the 2D CNN architecture, the authors first reshape the 870 features to a matrix, which constitutes the input to the 2D convolution layer. In a different line of work, a deep learning architecture which operates on raw MEG signals is proposed in \cite{aoe2019automatic}. Specifically, authors propose a method to distinguish between patients with epilepsy, spinal cord injuries and healthy subjects. The input to the model is a 160-channel MEG signal which is processed using a series of convolution and pooling layers. Prior to classification the authors concatenate the resultant feature vector from this network together with a Fourier transformation of the input MEG signal which is then passed through a series of fully-connected layers to generate the 3 class classification.

\begin{figure}[htbp]
  \centering
  \includegraphics[width=\linewidth]{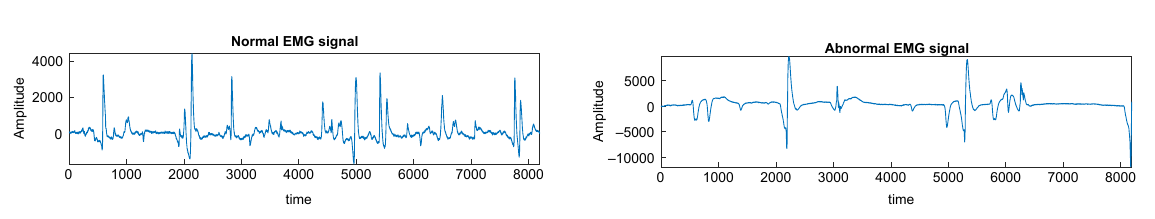}
    \caption{Illustration of normal and abnormal EMG signals. Recreated from \cite{subasi2019practical}}
    \label{fig:EMG}
\end{figure}

\textbf{Electromyography (EMG)}: EMG is the process of recording electric potential generated by muscle cells to diagnose the health of muscles and motor neurons. Depending on the study either needle or surface electrodes are placed to measure electrical activity. Fig. \ref{fig:EMG} illustrates an example of normal and abnormal EMG signals. The needle electrode can directly measure the electrical activity of the a muscle while surface electrodes are used to measure the electrical signal that is traveling between two or more points. EMGs are frequently used to diagnose muscle disorders such as muscular dystrophy and polymyositis, disorders that affect the motor neurons (eg. amyotrophic lateral sclerosis or polio), nerves and muscles (eg. myasthenia gravis). 

Considering the existing literature we observe that EMG based abnormality detection in not widely studied. However, there does exist deep learning based research conducted using EMGs to detect muscle activation \cite{khowailed2019neural}, to extract information on muscle movement \cite{olsson2019extraction} and to predict movement intent \cite{dantas2019deep}

\textbf{Summary}

Similar to biomedical imaging sensors we observe that electrical biomedical sensors also have similar designs and signal characteristics among different sensor types. However, in contrast to biomedical imaging, the electrical biomedical signals carry temporal information. Hence, it is vital to model the temporal evolution of the signal and to identify how different characteristics of the signal change over time. Therefore, temporal modelling has become vital when detecting abnormalities in electrical biomedical signals, and several different such architectures have emerged. Please refer to Sec. 2.2 of the main document for detail comparison between such architectures. 

Tab. \ref{tab:list_of_datasets_electric} summarises datasets with electrical biomedical signals which are publicly available and extensively used in machine learning research.

\begin{table*}[htbp]
\caption{List of publicly available datasets with electrical biomedical signals.}
\centering
\begin{adjustbox}{max width=\textwidth} 
\begin{tabular}{|c|c|p{0.3\textwidth}|p{0.5\textwidth}|c|c|c|}
\hline
Data Type              & Task                               & Dataset Name                                             & Description   & {No Channels} & {No Subjects}                                                                                                                                                                                                                                                          & References                                                              \\ \hline

\multirow{5}{*}{ECG}   & \multirow{3}{*}{Cardiac Abnormality Detection}          & PhysioNet-CinC 2020                                          &  Identify clinical diagnoses & {12} & {43,135} &\href{https://physionet.org/content/challenge-2020/1.0.0/}{link} \\ \cline{3-7} 
                      &                                    & MIT-BIH Atrial Fibrillation Database
                                & Subjects with atrial fibrillation.                                                                                                                              & {2} &  {25 (records)}     & \href{https://physionet.org/content/afdb/1.0.0/}{link}                    \\ \cline{3-7} 
                                
                    &                                    & {MIT-BIH Arrhythmia Database}
                                & {Evaluation of arrhythmia detectors}                                                                                                                              & {2} &   {48}    & {\href{https://www.physionet.org/content/mitdb/1.0.0/}{link}}                    \\ \cline{3-7} 
                                
                     &                                    & {BIDMC}
                                & {Congestive heart failure detection}                                                                                                                              & {2} &   {15}    & {\href{https://physionet.org/content/chfdb/1.0.0/}{link}}                    \\ \cline{3-7} 
                                
                    &                                    & {Fantasia}
                                & {Congestive heart failure detection}                                                                                                                              & {2} &   {40}    & {\href{https://physionet.org/content/fantasia/1.0.0/}{link} }                   \\ \cline{3-7} 
                                
                    &                                    & {INCART}
                                & {Arrhythmia detection}                                                                                                                              & {12} &   {32}    & {\href{https://physionet.org/content/incartdb/1.0.0/}{link} }                   \\ \cline{3-7} 
                    &                                    & {CCDD}
                                & {Cardiovascular disease diagnosis}                                                                                                                              & {12} &   {1500 (records)}    & {\href{https://ieeexplore.ieee.org/document/5521712}{link}}                    \\ \cline{3-7} 
                                
                    &                                    & {CSE}
                                & {Cardiovascular disease diagnosis}                                                                                                                              & {12} &   {1000 (records)}    & {\href{https://archive.physionet.org/physiobank/other.shtml}{link}      }              \\ \hline

\multirow{5}{*}{EEG}   & \multirow{3}{*}{Epilepsy}          & Bonn University                                          & Epilepsy detection & {1} & {25} & \href{http://epileptologie-bonn.de/cms/upload/workgroup/lehnertz/eegdata.html}{link} \\ \cline{3-7} 
                      &                                    & TUH Abnormal EEG database                                &   Brain disorders diagnosis                                                                                                                               & {24-36} &  {10874}  & \href{https://www.isip.piconepress.com/projects/tuh\_eeg/}{link}                    \\ \cline{3-7} 
                      &                                    & Freiburg Hospital                                        &   Epilepsy detection                                & {128} &  {200}   & \href{https://epilepsy.uni-freiburg.de/database}{link}                             \\ \cline{3-7} 
                      
                       &                                    & {EPILEPSIAE}                                   &   {Epilepsy detection}                                & {122} &  {30}   & {\href{http://epilepsy-database.eu/}{link}}                             \\ \cline{3-7}
                       
                       &                                    & {MSSM}                                        &   {Epilepsy detection}                                & {22} &  {28}   & {\href{https://icahn.mssm.edu/research/portal?tab=Labs}{link}}                             \\ \cline{3-7}
                       
                       &                                    & {CHB-MIT}                                        &   {Epilepsy detection}                                 & {23} &  {22}   & {\href{https://icahn.mssm.edu/research/portal?tab=Labs}{link}}                             \\ \cline{2-7}
                       
                      & \multirow{2}{*}{Sleep}             & Sleep EDF database                                       & Sleep disorder diagnosis                                                                                                                                                                             & {2} & {197 (records)}   & \href{https://physionet.org/content/sleep-edfx/}{link}                             \\ \cline{3-7} 
                      &                                    & Cyclic Alternating Pattern (CAP) of EEG Activity dataset & Sleep disorder diagnosis                                                                                                                                                                  & {3} & {108}  & \href{https://physionet.org/content/capslpdb/1.0.0/}{link}                       \\ \hline
                      
{MEG}   & {Brain Functions}         & {Cam-CAN }                                         & {Age-related segregation of brain functions} & {306} & {674} & {\href{https://camcan-archive.mrc-cbu.cam.ac.uk/dataaccess/}{link}} \\ \hline

\end{tabular}
\end{adjustbox}
\label{tab:list_of_datasets_electric}
\end{table*}

\subsection{Miscellaneous data types}
\begin{figure}[htbp]
  \centering
  \subfloat[][]{\includegraphics[width=.55\linewidth]{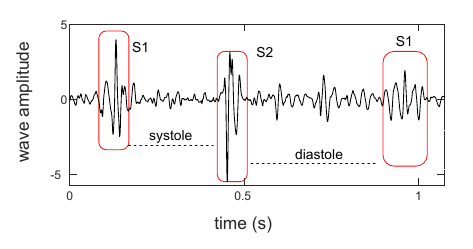}}
  \subfloat[][]{\includegraphics[width=.35\linewidth]{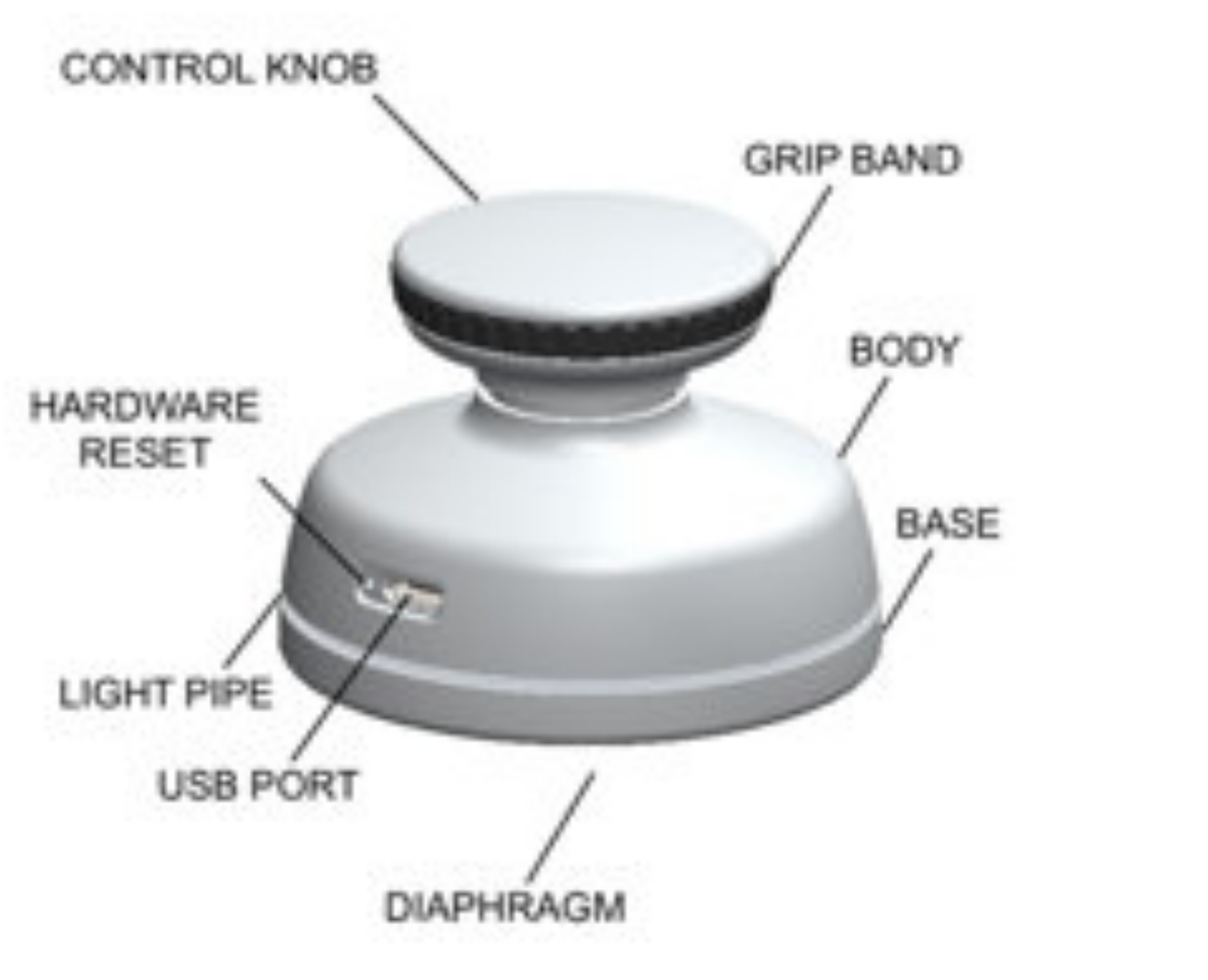}}
    \caption{(a)Illustration of heart states and systole / diastole intervals. (b) Illustration of \href{https://www.stethee.com/} { Stethee\textregistered{}}{a wireless electronic stethoscope)}}
    \label{fig:PCG}
\end{figure}

\textbf{Phonocardiography (PCG)}: PCG is the recording of the sounds observed during cardiac auscultation. This is a fundamental step of a physical examination in clinical practice. It has proven to be one of the most cost effective
methods for screening for numerous heart abnormalities, including arrhythmia, valve disease and heart failure. In each heart cycle the variation in blood pressure produced by the closure of the mitral and tricuspid valves and their vibrations causes the first heart sound (S1). Then the second heart sound (S2) is generated by the closure of the aortic and pulmonary valves and their vibrations. The window between S1 and S2 is termed the systole interval, and the diastole interval is from S2 to the beginning of S1 in the next heart cycle. Fig. \ref{fig:PCG} (a) shows how the first and second heart sounds appear in a typical PCG recording. 

The availability of electronic stethoscopes such as Littmann, Cardionics E-Scope and Stethee which digitise the analogue audio recordings, and the public availability of large scale datasets such as 2016 PhysioNet/CinC Challenge \cite{clifford2016classification} has encouraged many deep learning practitioners to apply deep learning techniques to detect abnormal heart sounds. For instance, in \cite{Zhang2019AbnormalSegmentation} the authors propose the use of temporal quasi-periodic features together with an LSTM network to detect abnormal heart sounds. In contrast \cite{Maknickas2017RecognitionCoefficients,dissanayake2020understanding} employed a convolutional neural network structure and extracted  Mel-frequency cepstral coefficients (MFCCs) from the heart sound recordings. In \cite{potes2016ensemble} the authors propose using a hand crafted set of features such as PCG Amplitudes, and Power spectral and Frequency-based features together with a CNN. 

\textbf{Wearable Medical Devices}: Wearable sensing devices are an emerging technology to monitor the wearer's health and detect abnormalities. Modern wearable devices such as smartwatches, smart clothing, smart footwear, and fitness trackers have the ability to measure various parameters, including, heart rate, body temperature, muscle activity, and blood/tissue oxygenation \cite{sannino2019artificial}. 

There exists a small number of deep learning approaches for anomaly detection using such wearable devices. In \cite{potluri2019deep} the authors detect gait abnormalities using data captured by an array of insole sensors. These sensors measure the pressure distribution during walking and standing and analyse the collected data using an LSTM based architecture. In another line of work \cite{mauldin2018smartfall}, accelerometer data collected from smartwatches is leveraged to design a fall detection system. The authors proposed to run their RNN anomaly detection framework on a smart phone which has been paired with the smart watch to generate alarms when a fall is detected. 

\textbf{Summary}
Despite the vast differences in sensory types that we discussed, ranging from auditory sensors, motion senors to temperature sensors, all of the above devices capture temporal signals similar to electrical biomedical sensors. Therefore, in the existing literature similar model architectures are utilised to capture temporal information within these biomedical signals. In Sec. 2.3 of the main document we discuss such algorithms in detail. 
 
Tab. \ref{tab:list_of_datasets_mis} summarises a list of audio and wearable device datasets that are publicly available for machine learning research.

\begin{table*}[htbp]
\caption{List of publicly available audio, vital sign and wearable device datasets.}
\centering
\begin{adjustbox}{max width=\textwidth} 
\begin{tabular}{|c|c|p{0.2\textwidth}|p{0.45\textwidth}|c|c|c|}
\hline
Data Type              & Task                               & Dataset Name                                             & Description                                                                                                                                                                                                                            & {No recordings}  & {Other Modalities}                                & References                                                              \\ \hline

\multirow{5}{*}{PCG}   & \multirow{3}{*}{Heart sound Anomaly Detection}          & PhysioNet-CinC 2016                                          &  Identifying cardiovascular diseases  & {3,126} & {ECG for subset (a)} & \href{https://physionet.org/content/challenge-2016/}{link} \\ \cline{3-7} 
                      &                                    & Fetal Heart sound database
                                &  constructed using recordings made from pregnant women                                                                                                                                    & {109} &  {PCG from mother and the fetal} & \href{https://physionet.org/content/sufhsdb/}{link}                    \\ \cline{3-7} 
                                
                     &                                    & {PASCAL}
                                &  {Identifying cardiovascular diseases}                                                                                                                                   & {461} &  {NO} & {\href{http://www.peterjbentley.com/heartchallenge/}{link}}                    \\ \cline{3-7} 
                                
                     &                                    & {Murmur database} 
                                &  {Identifying cardiovascular diseases}                                                                                                                                   &  {1000} &  {NO} & {\href{https://github.com/yaseen21khan/Classification-of-Heart-Sound-Signal-Using-Multiple-Features-}{link}}                    \\ \hline

\multirow{5}{*}{{Respiratory Sounds}}   & \multirow{3}{*}{{Classification of Respiratory Diseases}}          & {Respiratory Sound Database}                                          & {Detection of respiratory cycles and abnormalties} & {920} & {NO} &  {\href{https://www.kaggle.com/vbookshelf/respiratory-sound-database}{link}} \\ \cline{3-7} 

                                 &   & {R.A.L.E}   & {Detection of respiratory cycles and abnormalties} & {50} & {NO} &  {\href{http://www.rale.ca/LungSounds.htm}{link}} \\ \cline{3-7}  
                                 
                                  &  & {HF Lung V1}   & {Detection of respiratory cycles and abnormalties} & {9765} & {NO} &  {\href{https://gitlab.com/techsupportHF/HF_Lung_V1/-/tree/master/}{link}} \\ \hline 

\multirow{5}{*}{{Vital Sign}}   & \multirow{3}{*}{{Patient Monitoring}}          

& {The University of Queensland Vital Signs Dataset}
                                           & {Anesthesia Patient Monitoring}  & {32} & {EEG, ECG, pulse oximeter, capnograph, arterial blood pressure} &  {\href{https://outbox.eait.uq.edu.au/uqdliu3/uqvitalsignsdataset/browse.html}{link}} \\ \cline{3-7} 
& & {MIMIC-III}                                           & {Mortality and length-of-stay predictions} & {34,472} & {ECG, respiration waveform, arterial blood pressure} & { \href{https://mimic.physionet.org/}{link}} \\ \cline{3-7} 
& & {MIMIC II: Waveform Database}                                           & {Abnormality detection and blood pressure predictions} & {23,180} & {ECG, respiration waveform, arterial blood pressure, plethysmograph} &  {\href{https://archive.physionet.org/mimic2/mimic2_waveform_overview.shtml}{link}} \\ \hline 

\multirow{5}{*}{Wearable Devices} 
& {Stress and Affect Detection} & WESAD                                          &  affective states and emotion identification & {15} & {skin temperature, ECG, blood volume pulse, respiration, accelerometers} &  \href{https://archive.ics.uci.edu/ml/datasets/WESAD+\%28Wearable+Stress+and+Affect+Detection\%29}{link} \\ \cline{2-7}

& \multirow{3}{*}{{Fall Detection}} & {MobiFall and MobiAct}  & {Fall detection in active daily living scenarios} & {3200} & {accelerometer, gyroscope} &  {\href{https://bmi.hmu.gr/the-mobifall-and-mobiact-datasets-2/}{link}} \\ \cline{3-7} 
& & {UR Fall Detection Dataset}  & {Fall detection in active daily living scenarios} & {70} & {accelerometer, RGB images, depth images}  &  {\href{http://fenix.univ.rzeszow.pl/~mkepski/ds/uf.html}{link}} \\ \cline{3-7}
& & {TST V2}  & {Fall detection in active daily living scenarios} & {264} & {accelerometer, RGB images, depth images}  &  {\href{https://ieee-dataport.org/documents/tst-fall-detection-dataset-v2}{link}} \\ \cline{3-7}
& & {UniMiB SHAR}
  & {Motion abnormalities and fall detection} & {11,771}  & {accelerometer} &  {\href{http://www.sal.disco.unimib.it/technologies/unimib-shar/}{link}} \\ \hline

\end{tabular}
\end{adjustbox}
\label{tab:list_of_datasets_mis}
\end{table*}

\end{document}